\documentclass[letterpaper, 10 pt, conference]{ieeeconf}  

\makeatletter
\newcommand*\titleheader[1]{\gdef\@titleheader{#1}}
\AtBeginDocument{%
  \let\st@red@title\@title
  \def\@title{%
    \vskip-3.0em
    \bgroup\normalfont\small\centering\@titleheader\par\egroup
    \vskip1.5em\st@red@title}
}
\makeatother

\titleheader{Please cite as: N. Stathoulopoulos, B. Lindqvist, A. Koval, A. -A. Agha-Mohammadi and G. Nikolakopoulos, ``FRAME: A Modular Framework for Autonomous Map Merging: Advancements in the Field," in IEEE Transactions on Field Robotics, vol. 1, pp. 1-26, 2024, doi: 10.1109/TFR.2024.3419439.}

\IEEEoverridecommandlockouts                              

\usepackage{graphicx}
\usepackage{setspace}
\usepackage{amsmath,amssymb,amsfonts}
\usepackage[linesnumbered,lined,ruled,commentsnumbered]{algorithm2e}\usepackage{graphicx}
\usepackage[breaklinks=true,colorlinks,citecolor=blue]{hyperref}
\usepackage{textcomp}
\usepackage{xcolor}
\usepackage{graphics} 
\usepackage{epsfig} 
\usepackage{times} 
\usepackage{mathtools}
\usepackage{subcaption}
\usepackage{booktabs}
\usepackage{adjustbox}
\usepackage{caption}
\usepackage{stfloats}
\usepackage{tabularray}
\usepackage{units}

\def\BibTeX{{\rm B\kern-.05em{\sc i\kern-.025em b}\kern-.08em
    T\kern-.1667em\lower.7ex\hbox{E}\kern-.125emX}}

\newcommand{\black}[1]{\textcolor{black}{#1}}

\title{\LARGE \bf
FRAME: A Modular Framework for Autonomous Map Merging: Advancements in the Field$^*$
}

\author{Nikolaos Stathoulopoulos$^{\dagger \, 1}$, Bj\"{o}rn Lindqvist$^1$, Anton Koval$^1$, Ali-akbar Agha-mohammadi$^2$ \\
and George Nikolakopoulos$^1$
\thanks{$^*$This article is an extended version of our paper FRAME~\cite{stathoulopoulos2023frame}, published in the 2023 IEEE International Conference on Robotics and Automation}
\thanks{$^{\dagger}$Corresponding author's e-mail: \texttt{niksta@ltu.se}}
\thanks{$^{1}$The authors are with the Department of Computer, Electrical and Space Engineering, Robotics and AI Group, Lule\r{a} University of Technology,}%
\thanks{$^{2}$The author is with Jet Propulsion Laboratory, California Institute of Technology Pasadena, CA, 911 09.}%
}

\begin{document}

\maketitle
\thispagestyle{empty}
\pagestyle{empty}

\begin{abstract}
In this article, a novel approach for merging 3D point cloud maps in the context of egocentric multi-robot exploration is presented. Unlike traditional methods, the proposed approach leverages state-of-the-art place recognition and learned descriptors to efficiently detect overlap between maps, eliminating the need for the time-consuming global feature extraction and feature matching process. The estimated overlapping regions are used to calculate a homogeneous rigid transform, which serves as an initial condition for the GICP point cloud registration algorithm to refine the alignment between the maps. The advantages of this approach include faster processing time, improved accuracy, and increased robustness in challenging environments. Furthermore, the effectiveness of the proposed framework is successfully demonstrated through multiple field missions of robot exploration in a variety of different underground environments.
\end{abstract}


\section{Introduction}

Three-dimensional point cloud maps are a powerful tool for representing and understanding the geometry of real-world environments. These maps consist of a set of 3D points that correspond to the surfaces and features in an environment, and they can be created using a variety of methods, such as laser scanning and structured light sensing. In recent years, great emphasis has been given to researching and developing techniques for locating and reconstructing unknown environments autonomously, with numerous real-world applications, such as mine exploration~\cite{dark_mine}, planetary exploration~\cite{planetary}, search and rescue missions~\cite{compra}, industrial inspection~\cite{ind_insp} and so on. In order to accommodate for greater and more complex environments, researchers have turned to Multi-Robot Systems (MRS) to explore and map environments more efficiently and reliably~\cite{nebula}. These systems allow any number of agents to cooperate and explore an environment, leveraging task parallelization to improve time efficiency and providing redundancy and resiliency to improve reliability.

It is more than evident that all these applications share the need for an autonomous map merging procedure, especially when multi-robot systems are deployed in the field. Three-dimensional point cloud map merging is a crucial technology in the field of robotics and automation, as it involves combining multiple 3D point cloud maps into a single comprehensive map to provide a complete picture of the environment. This technique is particularly useful in multi-robot coordination, where several robots can work together to explore and map a large area. Typically, each robot generates a local map, within its local frame, that acts as a source of information for localization, collision avoidance, navigation, and path planning, and can later on be shared and fused into a global map. By combining the maps generated by multiple robots, a more comprehensive and accurate picture of the environment can be created, enabling the robots to work together more effectively and avoid collisions or redundant exploration of already-visited areas. Additionally, map merging can help to reduce the overall exploration time and increase the coverage area, as multiple robots can simultaneously explore different parts of the environment. Overall, map merging is a critical technology that enhances the capabilities of multi-robot exploration and enables more efficient and effective exploration of large areas.

One of the main use cases for map merging is in subterranean mining and construction, which will be the main evaluation use case in this manuscript as well. Mining companies are already using advanced 3D point cloud mapping technologies to generate accurate maps of their vast operational areas. But, mining tunnels are constantly changing as the tunnels are excavated further, and new areas need to constantly be integrated in the global mine map. The development of autonomous routine mapping and inspection missions by robotic platforms with automatic map merging and map integration could greatly reduce costs and increase efficiency of surveying and updating the maps of these massive mining areas. Similarly, map merging and comparison technologies~\cite{stathoulopoulos2023irregular} can enable the detection of tunnels drifting over time which can be an indicator of high rock stress or possible collapse, indicating the need for maintenance. In general, the resulting comprehensive map serves as a valuable asset for planning and decision-making, as it provides a detailed and accurate representation of the mine. Overall, map merging is a critical technology for the mining industry that enables better monitoring, maintenance, and decision-making, leading to improved safety and operational efficiency. 

\begin{figure*}[t!] 
    \includegraphics[width=\textwidth]{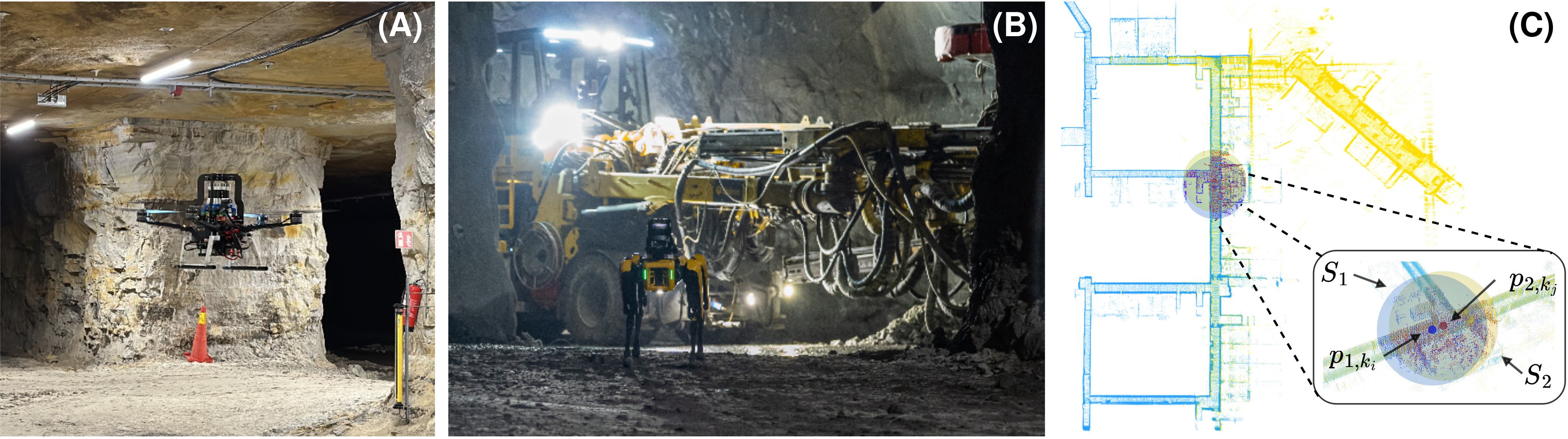}
    \caption{(A) One of the custom-built quad rotors that was utilized during the series of experiments in this article. (B) Spot in a construction area during the field trials. (Copyright NCC) (C) The outcome of the proposed framework from a larger scale indoor environment from Lule\r{a} University of Technology, where the two robots start together but explore different branches of the building.} \label{fig:concept}
    \vspace{-0.6cm}
\end{figure*}

However, map merging can be a challenging task, especially when dealing with large datasets or deploying it at large scale areas. The complexity of the map merging process is determined by a variety of factors, including the level of awareness about the agents' relative positions and orientations, as well as the accuracy and noise levels of the sensors equipped on each robot. Multi-modal systems~\cite{multi_modal} and varying sensor configurations on different robotic platforms, as seen on Fig.~\ref{fig:concept}, can further complicate the map merging process, making it difficult to directly fuse individual maps.

To address this challenge, 3D point cloud map merging is a key step in many robotics and computer vision tasks, including localization, navigation, and object recognition. There are several algorithms and techniques that can be used to perform 3D point cloud map merging, and the choice of method depends on the specific requirements and constraints of the application, as well as the characteristics of the point clouds being merged. Some common techniques for merging point cloud maps include point cloud registration and fusion, surface fitting, and feature matching. One of the main challenges in 3D point cloud map merging is accurately aligning the different point clouds, that is typically done by finding corresponding points in the different point clouds and using these correspondences to estimate the relative pose (i.e., position and orientation) of the point clouds. There are various methods that can be used to find correspondences, including feature matching and surface fitting. Once the relative pose has been estimated, the point clouds can be transformed and combined using techniques such as point cloud registration and fusion. The goal of 3D point cloud map merging is to produce a single, accurate, and complete map of the environment. In order to achieve this goal, it is important to minimize errors and inconsistencies in the merged map, which can be caused by noise, outliers, and other factors. There are various techniques that can be used to improve the accuracy and completeness of the merged map, including filtering, smoothing, and outlier removal. Overall, 3D point cloud map merging is a complex and challenging task that requires careful planning and execution. By effectively combining multiple point cloud maps, it is possible to create a more accurate and complete representation of an environment, which can have numerous practical applications in robotics and computer vision. Whether it is laser scans, structured light sensors, or other types of 3D data, the ability to merge point cloud maps is a valuable skill that can enable tackling a wide range of real-world problems.


\subsection{Related Work} \label{sec:related_work}

In this section, we delve into contemporary developments within the literature, focusing on two distinct areas. Our attention is initially directed towards the ongoing advancements in map merging algorithms. Subsequently, we turn our focus to the current state-of-the-art literature concerning place recognition descriptors. This sequential exploration is motivated by the direct correlation between place recognition descriptors and our proposed solution.\\


\subsubsection{\textbf{Map Merging}}

Current research on map merging primarily addresses the problem using methods based on 2D occupancy grid maps. These methods include those that use probability~\cite{robust-multi},~\cite{simultaneous-merging}, optimization~\cite{merging-occupancy},~\cite{on-merging}, and feature-based techniques~\cite{feature-semantic}. Feature-based methods often involve extracting point, line, or geometric features in order to match and merge local maps. Wang et al.~\cite{rtm} treated occupancy grid maps as images, extracted Scale-Invariant Feature Transform (SIFT) features~\cite{sift} and merged local maps using the Iterative Closest Point (ICP)~\cite{icp} scan matching algorithm. More recently, Sun et al.~\cite{subgraph}  proposed a maximum common subgraph-based algorithm, where Harris corner points~\cite{harris} were first extracted, then the maximum common subgraph was found using an iterative algorithm, and finally a transformation matrix was calculated based on the relationship between the corner points in order to merge the maps. Another method for resolving the issue of merging occupancy grid maps is presented by Park et al. in ~\cite{7479743}. This approach utilizes rectangular features to identify the best shared areas between maps. By determining the dimensions and connections of the largest empty rectangles, the algorithm is able to match orientations and scales, as well as find overlapping areas. One benefit of this method is that it is able to merge maps without requiring any information about the relative locations of the robots. However, occupancy grid maps have limitations, particularly in multi-robot systems with different types of robots, e.g. aerial and ground robots, where different occupancy maps might be created, due to different operating height and viewing angles of the sensors, therefore making the accurate overlap matching process impossible. In addition to this, with the size of the environment and the processing and storage power required to maintain an occupancy grid map increases with the size of the environment~\cite{slam-rao}, making it impractical for large-scale environments. In order to overcome these limitations, more recent studies have focused on extracting features from the environment and creating feature-based 3D maps, such as dense point clouds.

The main distinction between 3D map merging and fusion and 2D map matching is the higher dimensionality of the former, which leads to increased memory~\cite{Jessup2017RobustAE},~\cite{8451911}, bandwidth~\cite{cloud_based},~\cite{8451911}, and processing requirements~\cite{cloud_based},~\cite{Bonanni2017}. These issues are explicitly addressed in research on 3D map merging. Otherwise, the approaches used for 3D map matching are similar to those used for 2D maps, but they aim to reduce computational complexity by using structural features, matching submaps, or a combination of both. The use of dense point clouds is not only limited to overcoming the limitations of occupancy maps, but also commonly used to present a more detailed representation of the environment and extract feature points. These feature points can be applied in two ways, such as matching within a single robot's map generation or computing the transformation between maps generated by multiple robots. A crucial step in the map merging process is identifying stable feature points from point-based maps, as unstable feature points could lead to inaccurate matching results and negatively impact the overall map merging performance.

For such scenarios, researchers such as Konolige et al.~\cite{distributed} attempted to merge two local maps by matching features that were manually extracted from the maps, such as doors, junctions, and corners. These features were used to align the maps and merge them into a single map. Another approach, used by Sun et al.~\cite{sparse}, was to utilize the open source framework ORB\_SLAM2~\cite{orb2} to construct sparse point cloud maps. This was done by extracting FAST key points~\cite{fAST_ML} and BRIEF descriptors~\cite{brief} from the local maps, and then comparing these to an off-line dictionary using the bag of words method~\cite{bow}. However, this method can require more computational resources as more advanced features are typically used. Yue et al.~\cite{8451911} tried to reduce computational resources by reducing the search space for transformations by extracting structural edges from the voxel grid and using edge matching to guide the search, along with the use of local voxel information to refine the result. An alternative approach is to use plane features~\cite{planar_features}, which are extracted from the environment by fitting a plane model to multiple 3D points. These features are less affected by noise, more robust and faster to extract~\cite{point_plane}. When a robot encounters a large planar structure, it may only be able to see a portion of the plane in each frame due to limitations in the sensor's field of view. As a result, features from the same planar structure may appear in multiple frames and need to be merged to accurately reflect the structure in the environment. Similarly, Drwi\c ega~\cite{3d_map_server} proposed a 3D map merging scheme for multi-robot systems based on overlapping regions. The overlap estimation was performed using SHOT descriptors~\cite{shot} and the map alignment was done using SAC-IA. In another approach~\cite{3d-map-merge}, 3D map merging was achieved by first segmenting the map and using VFH descriptors~\cite{vfh}, and finally using KP-PDH descriptors which offer faster computational times compared to the SHOT descriptors. The proposed solution by Basso et al.~\cite{merging_3d_occu} involves using a keypoint detector and descriptor, as well as filters for keypoints and correspondences to merge 3D occupancy maps. The keypoint properties are determined using gradients from potential fields, and various image processing techniques are also employed to minimize noise in the map and enhance the accuracy of the transformation parameters.

The aforementioned approaches for map merging in autonomous multi-robot exploration have been found to be time-consuming, with overall processing times ranging from 15 to 250 seconds. This can be a significant issue, particularly for aerial platforms that need to react quickly to their environment. In order to address this, our proposed approach, FRAME, aims to minimize the processing time of the map merging algorithm by using deep learning-based descriptors. These descriptors are designed to provide fast and reliable results across a variety of environments and platforms. \\


\subsubsection{\textbf{Place Recognition}}

The process of 3D point cloud place recognition commonly involves the extraction of discriminative features. PointNetVLAD~\cite{uy2018pointnetvlad} pioneered an end-to-end trainable global descriptor by leveraging PointNet~\cite{charles2017pointnet} for local feature extraction and the NetVLAD aggregator~\cite{arandjelovic2016vlad} for generating global descriptors in LiDAR-based place recognition. Recognizing the limitations of PointNet in descriptive power, the alternative LPD-Net~\cite{liu2019lpdnet} was proposed. More recently, MinkLoc3D~\cite{komorowski2021mink} showcased enhanced efficiency and performance by employing sparse convolutions to capture point-level features effectively. Addressing global descriptor improvements, LoGG3D-Net~\cite{vidanapathirana2022logg3d} introduced a local consistency loss to ensure feature consistency within point clouds.

To improve computational efficiency, several strategies have been devised for processing larger scans, with many opting to transform them into an intermediary representation before subjecting them to deep neural network feature extraction. A notable contribution in this domain is the OREOS system, as detailed in~\cite{schaupp2019oreos}, specifically tailored for place recognition, with the additional functionality of estimating the yaw discrepancy between scans. The core methodology involves the projection of input data into a 2D range image using a spherical projection model.
Expanding on the concept of utilizing range images, the OverlapNet framework, introduced in~\cite{chen2020overlapnet}, goes a step further by benefiting on various cues, including range, normals, intensity, and semantic classes. These cues are effectively projected as spherical images, derived from the point cloud data, thereby contributing to the enhancement of the overall performance of the framework.
Taking strides towards a more advanced version, the OverlapTransformer, as explained in~\cite{ma2022overlaptrans}, not only provides a rotation-invariant representation but also exhibits faster inference capabilities. This is achieved by harnessing the transformative power of Transformer networks. However, it is worth noting that, unlike its predecessors, OverlapTransformer does not possess the capability to estimate yaw angles.

Several classical approaches for place recognition exist that do not rely on learning methodologies. One widely adopted method is Scan Context, as introduced by~\cite{giseop2018sc}. Scan Context encodes the maximum height of the point cloud in various bins, creating a 2D global descriptor for heightened discrimination. However, it is worth noting that this approach comes with an increased computational matching time.
In contrast to Scan Context, which relies solely on geometric information,~\cite{cop2018delight} propose Delight, incorporating the intensity readings of LiDAR into a series of histograms for a more comprehensive utilization of both intensity and geometric information. Building upon this idea,~\cite{han2020isc} extended Scan Context by leveraging both geometry and intensity. This extension outperforms descriptors based solely on geometry, employing the same space division method as utilized in Scan Context.
A recent advancement in this domain is the LiDAR-Iris proposed by~\cite{wang2020iris}, which employs the Fourier transform to generate a binary signature image. The process involves initially creating LiDAR-Iris images by expanding the bird's-eye view of the LiDAR scan into an image strip. Subsequently, Fourier transform is applied to these LiDAR-Iris images, facilitating spatial place recognition within the frequency domain. This innovative approach showcases the evolving landscape of classical techniques in the field of place recognition.


\begin{figure*}[!t]
    \centering
    \includegraphics[width=1.0\textwidth]{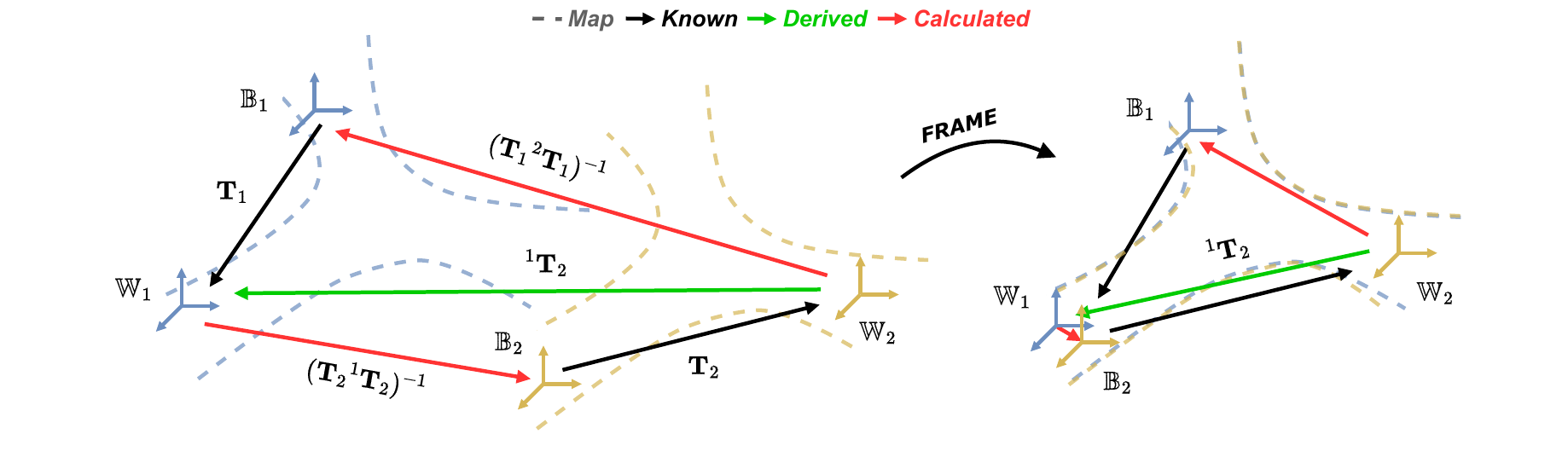}
    \caption{The coordinate frames for the map, denoted as $\mathbb{W}_ 1$ and $\mathbb{W}_2$, along with the robot frames $\mathbb{B}_1$ and $\mathbb{B}_2$, are depicted before and after the map alignment process. The transformations $\mathbf{T}_1$ and $\mathbf{T}_2$ represent known, non-static transforms between the robot and its static coordinate map frame. The spatial coordinate transform ${}^1\mathbf{T}_{2}$, derived from our proposed framework, facilitates the transformation of $\mathbb{W}_2$ to $\mathbb{W}_1$. All other transformations between each robot and map frame can be computed using the previously mentioned information.}
    \label{fig:frames}
    \vspace{-0.6cm}
\end{figure*}

\subsection{Overview of the proposed method}

\begin{figure*}[!b]
    \centering
    \includegraphics[width=\textwidth]{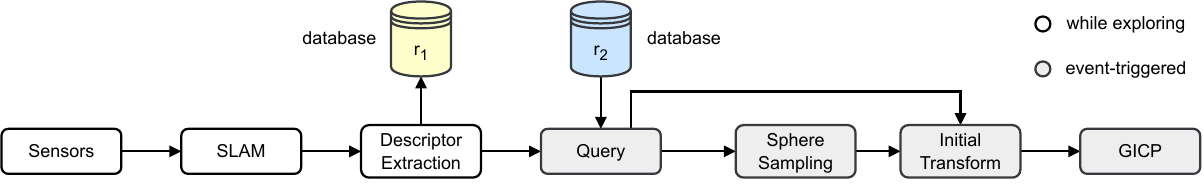}
    \caption{This figure illustrates the overall pipeline of the proposed framework. Both the SLAM process and Descriptor Extraction process are designed to be adaptable to different environments or use cases. White boxes denote the processes during exploration, while gray boxes indicate the merging process once it is initiated.}
    \label{fig:genral-arch}
\end{figure*}

As previously introduced in our conference article~\cite{stathoulopoulos2023frame}, we present a framework referred to as FRAME (Fast and Robust Autonomous Map-merging for Egocentric multi-robot exploration) that enables autonomous merging of 3D point cloud maps for egocentric multi-robot exploration. In the following paragraphs, we will elaborate on this approach and provide further evaluation in a series of field experiments. This is necessary as conference articles often have a page limit, which makes it challenging to delve into the details of complex concepts. FRAME addresses the challenge of aligning local maps by identifying acceptable spatial coordinate transformations, also known as map alignment, and can be used in real-time during multi-robot exploration missions, allowing for more efficient mapping by avoiding the need to explore already mapped areas. The method can also determine the relative positions of the robot, as shown in Fig.~\ref{fig:frames}, which for many existing map alignment methods, is a limiting prerequisite. In contrast to existing map merging methods, which often require identical map formats, an initial guess, and a high map overlap ratio~\cite{alignment, planar_features}, FRAME only assumes partial overlap, which is detected using place-specific descriptors learned by the system. This eliminates the need for time-consuming global feature extraction. Then, we use the learned orientation regression descriptors along the place-specific descriptors to obtain an initial homogeneous rigid transform, which is subsequently employed as a prior to enhance the speed and precision of the registration algorithm.
It is worth noting that the pipeline of FRAME supports a variety of place recognition frameworks for use in the descriptor extraction module. This feature allows for a more versatile architecture, accommodating different environments and applications. An overview if the general pipeline of the map merging process is depicted in Fig.~\ref{fig:genral-arch}.

\subsection{Contributions} \label{contributions}

In summary, the contributions of this work are:
\begin{enumerate} 
    \item[(a)] Introducing a novel approach for combining 3D point cloud maps in the context of egocentric multi-robot exploration, that unlike most methods, offers fast processing times and can be used in real-time. This allows for more efficient mapping by avoiding redundant exploration of previously mapped areas and faster exploration since it reduces the overall exploration time, thereby increasing the coverage area.
    \item[(b)] The proposed approach relies on the detection of overlapping regions and leverages state-of-the-art learned place recognition descriptors, avoiding of the typically time-consuming global feature extraction and matching process associated with handcrafted descriptors. Unlike other methods that necessitate an initial guess, our proposed framework autonomously determines the initial estimate, enhancing robot autonomy and decreasing reliance on human input. In contrast, our approach utilizes the trajectory and descriptors, which occupy only a few kilobytes for offline use, thereby contributing to increased autonomy in multi-robot exploration missions.
    \item[(c)] As an extension to the conference article, we revisited our work, establishing a more generic and modular approach. This evolution is evidenced through the incorporation of a place recognition comparison, runtime analysis, GICP fail case analysis, and a detailed discussion on the crucial aspect of selecting components tailored to different scenarios. We elaborate on our rationale behind the selection of descriptors for the field experiments, providing insights into the considerations driving our choices.
    \item[(d)] Enhancing the modular approach, we introduced an adaptive keyframe sampling module and automated sphere sampling radius selection. Notably, these additions eliminate the need for manual tuning, contributing to the adaptability of our framework across diverse environments.
    \item[(e)] Last but not least, in this iteration of FRAME, we introduced a more formalized problem formulation. This formulation accommodates any number of maps to be merged, setting a robust foundation for future works and expanding the applicability of our research.
\end{enumerate} 

The proposed approach is demonstrated to be effective through multiple field exploration missions in a variety of harsh underground environments, showing potential for real-world applications.
The remainder of this article is divided into the following sections. 
First, in Section~\ref{sec:methodology} we go through the problem formulation and then the proposed 3D point cloud map merging framework is introduced. Starting the experimental evaluation in Section~\ref{sec:preliminary}, we conduct a preliminary evaluation to test various submodules of the pipeline before proceeding to the field experiments. In Section~\ref{sec:setup}, the datasets, the robotic platforms and the metrics are presented. Section~\ref{sec:results} discusses the experimental results, in three parts. Initially in subsection~\ref{subsec:evaluation} we evaluate, in a variety of subterranean environments and at large-scale, then in subsection~\ref{subsec:autonomous} we demonstrate FRAME as an integrated part of fully autonomous missions, and then in subsection~\ref{subsec:comparisons} we compare with other available methods. Before concluding, Section~\ref{sec:limitations} addresses the current limitations and outlines the areas of future work. Last but not least, the article is summarized and concluded in Section~\ref{sec:conclusions}.


\section{Methodology} \label{sec:methodology}

The goal of this research is to create a new system for combining 3D point cloud maps in real-time, allowing multiple robots working on an exploration task to seamlessly merge their maps as long as they can communicate and their maps overlap. The merging process should be fast enough in order not to disturb the exploration and affect the general mission. Any delay caused directly affects the exploration time, since the batteries have a limited amount of energy they can provide. The mobile robots are equipped with a 3D LiDAR and an IMU, performing SLAM individually while exploring.
This section presents a framework for merging maps using deep learning-based descriptors derived from raw LiDAR scans, and can be summarized in the algorithm~(\ref{alg:while_exploration}). These descriptors are compact, high-dimensional vectors that capture features of the environment~\cite{uy2018pointnetvlad}. 
This framework is an extension of the 3DEG system~\cite{stathoulopoulos20223deg}, which was previously used for relocalization, and utilizes the same place recognition and yaw discrepancy regression descriptors to estimate the overlapping regions between two point cloud maps, eliminating the need for manual alignment. At this point, it is important to emphasize that the architecture of FRAME remains independent of the place recognition framework used to extract descriptors, as long as these descriptors meet specific requirements: (a) the ability to support fast search structures like a \textit{k}-d tree and (b) possessing yaw regression capabilities. Further elaboration on these aspects and a discussion of various place recognition frameworks can be found in Section~\ref{sec:preliminary}. The map merging algorithm uses two data structures containing the map descriptors, which are queried with each other to find the pair with the smallest distance. The resulting \black{indices} provide the corresponding trajectory points and initial transform, which roughly aligns the two point cloud maps. Finally, the \texttt{fast\_gicp} registration algorithm~\cite{fast_gicp} is used to refine the alignment and produce the final merged map. In order to continue with a more detailed description of the framework, we start by defining the problem and then further explaining each step.

\subsection{Problem Formulation} \label{subsec:problem_formulation}
\begin{figure*}[b!] 
    \centering
    \includegraphics[width=\textwidth]{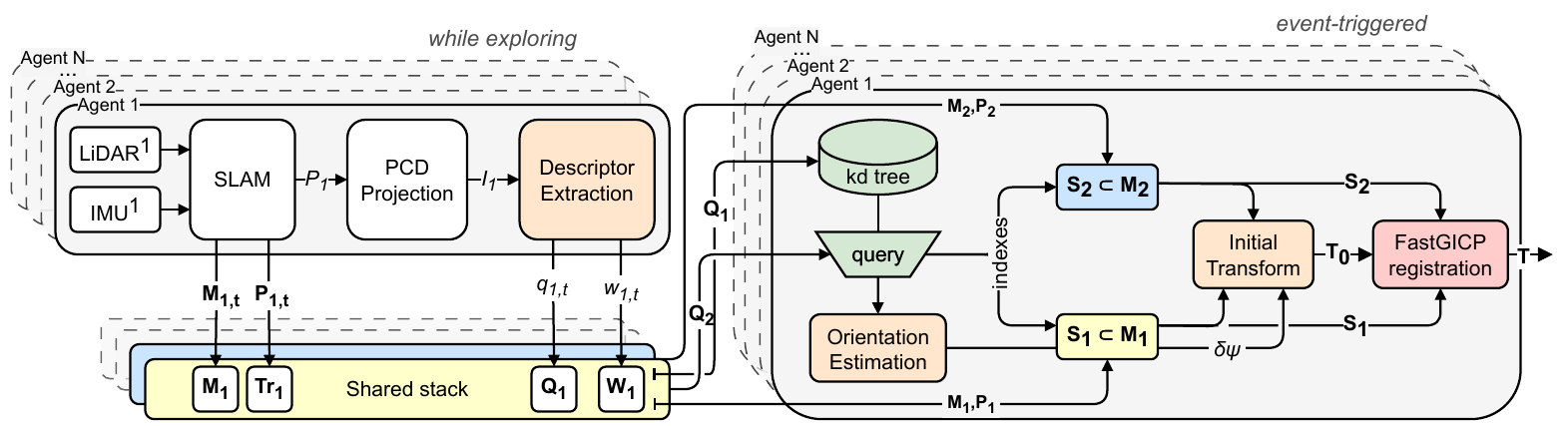}
    \caption{The overall map merging pipeline of FRAME. While the robots $\mathbf{r}_1, \mathbf{r}_2$ explore the surroundings, they collect the vector sets $\mathbf{Q}$ and $\mathbf{W}$. A predefined event will trigger the merging process, and as an egocentric approach each robot will create its own merged map $\mathbf{M}$, maintaining its local map frame as the global frame.}
    \label{fig:architecture}
    \vspace{-0.6cm}
\end{figure*}
Considering a system comprising of $N \in \mathbb{N}$ robots operating in 3D space $\mathbb{R}^3$, denoted as:
\begin{equation}
    \mathbf{R} = \{\mathbf{r}_1, \mathbf{r}_2, \ldots, \mathbf{r}_N\}
\end{equation}
Each robot $\mathbf{r}$ constructs a map ${}^{n}\mathbf{{M}}_r$ where its points $\mathbf{m} \in \mathbb{R}^3$ are registered with respect to its local world coordinate frame $\mathbb{W}_n$. The non-static body frame of the robot is designated as $\mathbb{B}_n$. The point cloud map for each robot is represented as:
\begin{equation}
    {}^{n}\mathbf{{M}}_r = \big\{{}^{n}\mathbf{{m}}_{r,1}, {}^{n}\mathbf{{m}}_{r,2}, \ldots, {}^{n}\mathbf{{m}}_{r,k}\big\} \; \text{with} \:\: k \in \mathbb{N},
\end{equation}
where the subscript $r$ specifies the agent and the superscript $n$ denotes the reference coordinate frame $\mathbb{W}_n$. The map merging process aims to combine multiple maps into a unified representation, the global map:
\begin{equation}
\begin{split}
    {}^{n}\mathbf{{M}}_G &= \big\{ {}^{n}\mathbf{M}_1, {}^{n}\mathbf{M}_2, \ldots,  {}^{n}\mathbf{M}_N \big\} \\
    &= \big\{ {}^{n}\mathbf{{T}}_1 {}^{1}\mathbf{M}_1, {}^{n}\mathbf{{T}}_2 {}^{2}\mathbf{M}_2, \ldots, {}^{n}\mathbf{{T}}_N {}^{N}\mathbf{M}_N \big\}
\end{split}
\end{equation}
Here, ${}^B\mathbf{{T}}_A: \mathbb{R}^3 \rightarrow \mathbb{R}^3$ represents the homogeneous rigid transformation of the special Euclidean group $SE(3)$, transforming frame $\mathbb{A}$ to frame $\mathbb{B}$:
\begin{equation}
    {}^B\mathbf{{T}}_A = \left[ \begin{array}{cc}
         R_z(\psi) & {}^B\mathbf{{t}}_A \\
         O_{1\times3} & 1
    \end{array} \right] \in SE(3)
\end{equation}
The rotation matrix $R_z(\psi) \in SO(3)$ represents the rotation about the z-axis by angle $\psi$ and ${}^B\mathbf{{t}}_A$ is the translation vector.
\begin{equation} 
    R_z(\psi) = \left[ \begin{array}{ccc}
         \cos{\psi} & -\sin{\psi} & 0 \\
         \sin{\psi} & \cos{\psi} & 0 \\
         0 & 0 & 1
    \end{array} \right] \:\: \text{and} \:\: {}^{B}\mathbf{t}_A = \left[ \begin{array}{ccc}
         t_x \\
         t_y \\
         t_z
    \end{array} \right]
\end{equation}
For simplicity and without loss of generality, each agent assumes it is agent number one. The objective is for each agent to transform all other maps to its static global frame $\mathbb{W} = \mathbb{W}_1$. The map merging function $f_m$ is defined as:
\begin{equation}
    f_m : \mathbb{R}^3 \times \mathbb{R}^3 \rightarrow \mathbb{R}^3
\end{equation}
In more detail, we define it as the function that unites all maps after being correctly transformed to the common global frame. The only constraint is that the sequential maps should have an overlap between them.
\begin{equation} \label{eq:f_m-general}
\begin{split}
    {}^{1}\mathbf{{M}}_G &= f_m \big( {}^{1}\mathbf{M}_1, {}^{2}\mathbf{M}_2, \ldots, {}^{N}\mathbf{M}_N \big) = {}^{1}\mathbf{M}_1 \bigcup_{n=2}^{N} {}^{1}\mathbf{T}_n {}^{n}\mathbf{M}_n, \\
    &\text{where} \;\; \black{{}^{1}\mathbf{V}_{G,n-1}} \cap \black{{}^{1}\mathbf{V}_n} \neq \O, \;\; \forall \; n \in \left[2,N\right] \subset \mathbb{N}
\end{split}
\end{equation}
\black{In this context, ${}^n\mathbf{V}_r$ is introduced as the volume covered by ${}^n\mathbf{M}_r$. This definition arises from our theoretical understanding of maps as sets of points, which inherently lack continuity. Consequently, there exists a possibility that the intersection of two point lists may yield a zero set. To address this, we define the volume of the maps and subsequently conduct the overlap check within these volumes.}
To simplify map merging, the assumption is made to merge two maps at a time. Multiple maps can be merged sequentially, provided there is overlap as defined above. At a given time step $k$, two maps ${}^{1}\mathbf{M}_1$ and ${}^{2}\mathbf{M}_2$, represented as sets of points $\mathbf{m} \in \mathbb{R}^3$, along with the corresponding trajectories ${}^1\mathbf{P}_1$ and ${}^2\mathbf{P}_2$, represented as sets of poses $ {}^n\mathbf{p}_k = \left[x_k, y_k, z_k\right]^\mathsf{T}_n \in \mathbb{R}^3$, are considered. The maps are defined as:
\begin{equation} \label{eq:maps}
\begin{split}
    {}^{1}\mathbf{M}_1 &= \big\{ \mathbf{m}_{1,1}, \mathbf{m}_{1,2}, \dots, \mathbf{m}_{1,n_1} \big\}, \;\\
    {}^{2}\mathbf{M}_2 &= \big\{ \mathbf{m}_{2,1}, \mathbf{m}_{2,2}, \dots, \mathbf{m}_{2,n_2} \big\}, \; \text{with} \:\: n_1, n_2 \in \mathbb{N}
\end{split}
\end{equation}
Similarly, the trajectories are denoted as:
\begin{align} \label{eq:trajectories}
\begin{split}
    {}^{1}\mathbf{P}_1 &= \big\{ \mathbf{p}_{1,1}, \mathbf{p}_{1,2}, \dots, \mathbf{p}_{1,k_1} \big\}, \; \\
    {}^{2}\mathbf{P}_2 &= \big\{ \mathbf{p}_{2,1}, \mathbf{p}_{2,2}, \dots, \mathbf{p}_{2,k_2} \big\}, \; \text{with} \:\: k_1, k_2 \in \mathbb{N}
\end{split}
\end{align}
The process of combining two maps can be described from Eq.~(\ref{eq:f_m-general}) for $N=2$ as:
\begin{equation} \label{eq:merged}
\begin{split}
    {}^{1}\mathbf{M}_G &= f_m \big( {}^{1}\mathbf{M}_1, {}^{2}\mathbf{M}_2 \big) = {}^{1}\mathbf{M}_1 \cup {}^{1}\mathbf{M}_2\\
    &= {}^{1}\mathbf{M}_1 \cup {}^{1}\mathbf{T}_{2} {}^{2}\mathbf{M}_2,
    \text{with} \:\: \black{{}^{1}\mathbf{V}_1 \cap {}^{1}\mathbf{V}_2} \neq \O
\end{split}
\end{equation}
The proposed approach breaks the problem down into two parts. The first step is to locate the two overlapping submaps and find an initial transform $\mathbf{T}_0 \in SE(3)$. Then, using this initial transform as an initial condition, a point cloud registration algorithm is used to improve the alignment and provide the final transform ${}^{1}\mathbf{T}_{2}$, as shown in Fig.~\ref{fig:architecture}.

\subsection{While exploring} \label{subsec:while_exploring}

When the mission begins, each robot $r$ independently explores its surroundings and gathers information that will be used later on to identify and merge overlapping regions. In order to do so, first the LiDAR scans are transformed into depth images through the point cloud projection module~(\ref{subsubsec:pcd_projection}) and then from these depth images the descriptor extraction module~(\ref{subsubsec:descriptor_extraction}) extracts descriptive vectors for place recognition and yaw discrepancy regression purposes. For gathering the information, we leverage an adaptive spatial keyframe sampling scheme, described in~(\ref{subsubsec:keyframe_sampling}).

 \begin{figure}[!b]
    \vspace{-0.6cm}
    \includegraphics[width=\columnwidth]{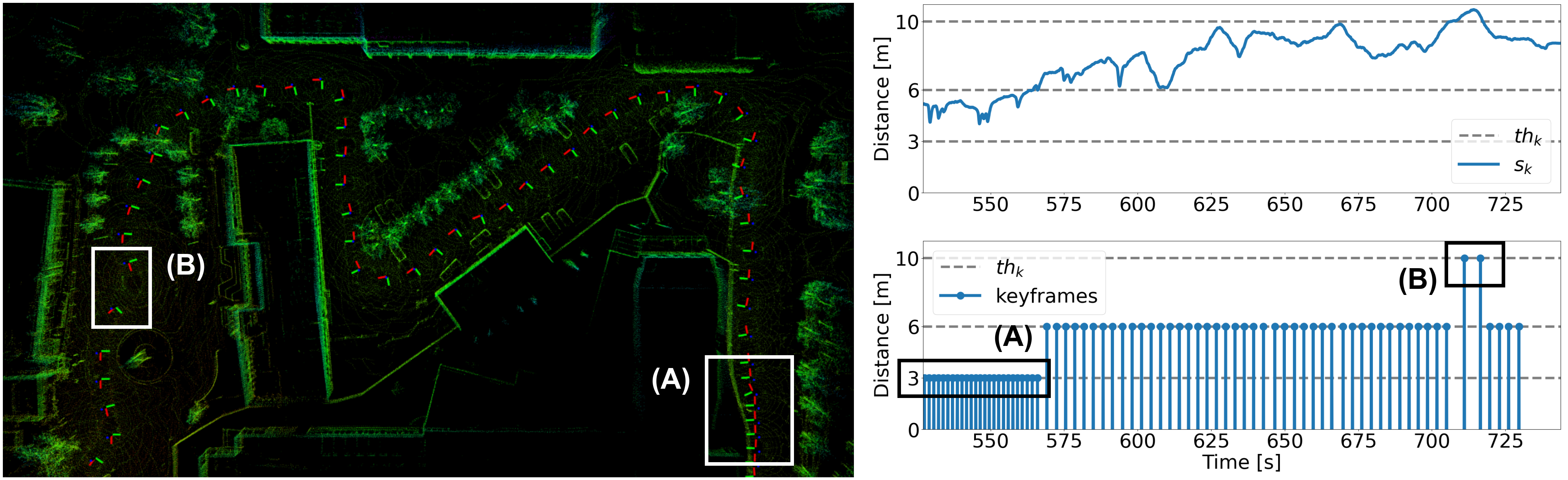}
    \caption{An example of the adaptive keyframe sampling thresholds: On the left, keyframes are illustrated on the map, while on the right, a plot depicts the variations in spaciousness alongside keyframe sampling. The threshold dynamically transitions from a smaller value, as depicted in (A), to a larger value, as demonstrated in (B).} \label{fig:keyframes}
 \end{figure} 
 
\subsubsection{\textbf{Adaptive Keyframe Sampling}}
\label{subsubsec:keyframe_sampling}

Keyframes are sampling positions that are commonly used in constructing global or local graphs. In most works in the current literature~\cite{LIO-SAM, LOCUS, lvisam2021shan}, keyframe nodes are dropped in a fixed threshold manner, e.g. every 2 \unit{meters} of translational displacement or every 15$^o$ of rotational change. Inspired by~\cite{DLO}, we adopt a similar logic of adaptively changing the translation threshold of sampling distance based on the spaciousness of the current point cloud scan $P_{r,k}$ since in large-scale surroundings, the characteristics detected by the point cloud scan remain noticeable for a longer duration and can be relied upon. Conversely, in confined or limited spaces, it is necessary to use a lower threshold to consistently capture the small-scale attributes such as tight corners. The spaciousness is defined as:
\begin{equation} \label{eq:spaciousness}
s_k = \alpha s_{k-1} + \beta S_k,
\end{equation}
where $S_k$ is the mean Euclidean point distance from the center of the point cloud to each point, and $s_k$ is the smoothed signal that defines the sampling threshold $th_k$. The constants $\alpha$ and $\beta$ are set to $0.9$ and, $0.1$ respectively.
\begin{equation} \label{eq:threshold}
th_k = \left\{
            \begin{array}{rl}
                 10 \text{m } & \text{if }\; s_k > 10 \text{m } \\
                 6 \text{m } & \text{if }\; 6 \text{m } < s_k \leq 10 \text{m } \\
                 3 \text{m } & \text{if }\; 3 \text{m } < s_k \leq 6 \text{m } \\
                 s_k & \text{if }\; 0 \text{m } < s_k \leq 3 \text{m }  \\
            \end{array}
        \right.
\end{equation}
In addition to Eq.~(\ref{eq:threshold}), a rotational threshold is held fixed at $30^o$ and a \textit{k}-d tree is constructed to search for the nearest neighbors and discard new keyframes that are within the set threshold, avoiding duplicate keyframes. Fig.~\ref{fig:keyframes} illustrates an example of the sampling process as well as the corresponding plot of the adaptive threshold.

\subsubsection{\textbf{Point Cloud Projection}} \label{subsubsec:pcd_projection}

The primary function of the Point Cloud Projection submodule is to convert the LiDAR point cloud scan data $P_{r,k}$ into a 2D depth image $I_{r,k}$ using a spherical projection model, performed for each time step $k$. \black{A common approach, as seen in prior works~\cite{chen2020overlapnet,ma2022overlaptrans, stathoulopoulos2024recnet}, involves transforming the point cloud $P_{r,k}$ into what is typically referred to as a \textit{vertex} map $\mathcal{V}:\mathbb{R}^2 \rightarrow \mathbb{R}^3$. In this mapping representation, each pixel corresponds to the nearest 3D point. A mapping function $\Pi:\mathbb{R}^3 \rightarrow \mathbb{R}^2$ is then applied to transform each point $p_i = (x, y, z) \in \mathbb{R}^3$ into spherical coordinates, which are subsequently mapped to image coordinates $I=(u, v) \in \mathbb{R}^2$. This process is denoted as:}
\begin{equation}
    \Bigg(\begin{array}{c}
       u  \\
       v  
    \end{array}\Bigg) = 
    \Bigg(\begin{array}{c}
        \frac{1}{2}\left[1 - \arctan{(yx^{-1})} \pi^{-1}\right]w \\
        \left[1 - (\arcsin{(zr^{-1})} + \text{f}_{\text{up}}) \text{f}^{-1}\right]h
    \end{array} \Bigg),
\end{equation}

\black{where $r = \|p\|_2$ denotes the range, $\text{f} = \text{f}_{\text{up}} + \text{f}_{\text{down}}$ is the vertical field-of-view and $w$, $h$ represent the width and height dimensions of the resultant vertex map $\mathcal{V}$.}
One of the significant advantages of converting LiDAR point cloud scans into range images, as opposed to using the raw point cloud data, is the ability to obtain a $360^o$ view of the environment. This panoramic view enables us to produce orientation-invariant descriptors that can be used for various applications, such as object detection and classification. When using a 2D CNN on these range images, the computational efficiency and ability to handle large datasets can be particularly beneficial. Additionally, 2D CNNs are well-suited for capturing translational invariance in the data, which is useful for detecting and classifying features from different viewpoints. However, there are some limitations to using range images. One of the main drawbacks is that they capture a less dense view of the surroundings compared to using a depth sensor. As a result, it may be challenging to extract detailed features from the converted range images, especially for tasks that require high accuracy and precision.

\begin{figure*}[!t]
    \centering
    \includegraphics[width=\textwidth]{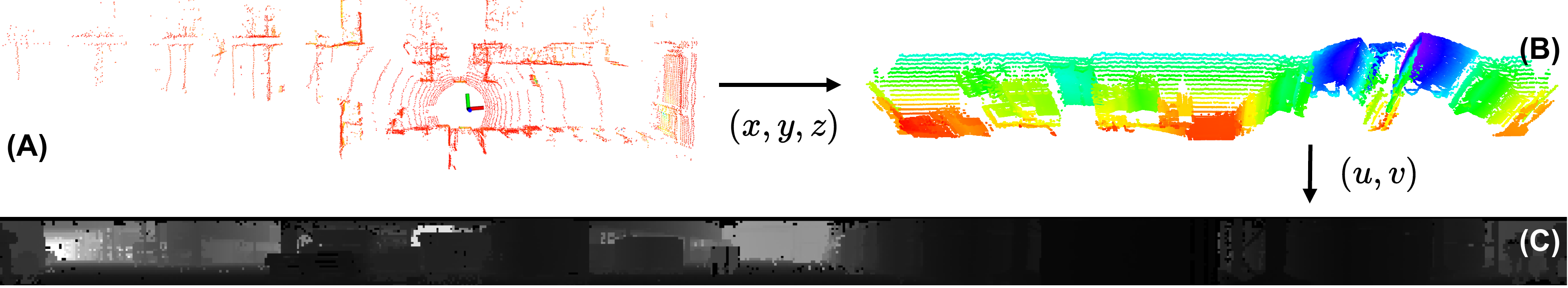}
    \caption{(A) The raw LiDAR scan $P_{r,k}$. (B) The outcome of the spherical projection. (C) The depth image $I_{r,k}$ after mapping the range of each point to each image pixel.}
    \label{fig:projection}
    \vspace{-0.6cm}
\end{figure*}

\subsubsection{\textbf{Descriptor Extraction}} \label{subsubsec:descriptor_extraction}

The Descriptor Extraction submodule takes as input the aforementioned depth images, $I_{r,k}$ and derives a compact $2\times64$ vector representation that captures the surrounding topological characteristics and are utilized for place recognition and orientation regression. The sets of vectors obtained in this manner are referred to as:
\begin{align}
    \mathbf{Q}_r = \, \{\vec q \in \mathbb{R}^{64}, \, n \in \mathbb{N}&: \vec q_{r,1}, \vec q_{r,2}, \dots, \vec q_{r,n} \} \\
    \mathbf{W}_r = \, \{\vec w \in \mathbb{R}^{64}, \, n \in \mathbb{N}&: \vec w_{r,1}, \vec w_{r,2}, \dots, \vec w_{r,n} \}
\end{align}
To be more precise, the vector denoted as $\vec q$ is a place-dependent vector that is independent of orientation. It is used to query for similar point clouds. On the other hand, the vector $\vec w$ is orientation-specific and is used for regressing the yaw discrepancy between two point clouds. Both vectors are based on OREOS~\cite{schaupp2019oreos}, and further information can be found in our extension 3DEG~\cite{stathoulopoulos20223deg}. These vectors, along with the corresponding pose $\mathbf{p}_{r,k}$ of the robot with respect to its local map frame $\mathbb{W}_r$, are collected and stacked in order to be used when the map merging process is triggered. 

\subsection{Event-triggered} \label{subsec:even_triggered}

Depending on the mission requirements, the merging of maps can be initiated by a predefined event. For instance, in a multi-agent centralized approach, the 3DEG descriptors can be used to classify the descriptors and initiate the merging process when the agents pass from a junction, that usually contains robust features for point cloud registration. As  another example, in a decentralized strategy, the merging process can commence upon the two robots establishing a communication connection. Once the connection is established, the event will activate the map merging process, which consists of the following submodules.

\subsubsection{\textbf{Overlapping Regions Selection}} \label{subsubsec:overlap_estimation}

Once the map merging process has been triggered, the first step is to use the vector sets $\mathbf{Q}_1$ and $\mathbf{Q}_2$ to identify the two overlapping regions, which are defined as $\mathbf{S}_1 \subset \mathbf{M}_1$ and $\mathbf{S}_2 \subset \mathbf{M}_2$. To accomplish this in an egocentric approach, each robot will create a \textit{k}-d tree using its own set of $\mathbf{Q}_1$ vectors and query it with the  vector set $\mathbf{Q}_2$ from the other robot. This method enables us to locate the pair of vectors $\vec q_{1,i}$ and $\vec q_{2,j}$ that have the minimum distance between them in the vector space, and thus originated from two similar point clouds.
\begin{equation} \label{eq:argmin} 
    (k_i,k_j) = \operatorname*{arg\,min}_{(i,j) \, \in \, \mathbb{N}}f\big(\mathbf{Q}_{1,i},\mathbf{Q}_{2,j}\big)
\end{equation}
The indices $i,j$ provide information on which time instance $k_i, k_j$ was selected from each vector set and consequently what was the position $\mathbf{p}_{1,k_i}$ and $\mathbf{p}_{2,k_j}$ for each robot. This information can be used for the next step, which is determining the homogeneous initial transform $\mathbf{T}_0 \in SE(3)$ between the past robot frames $\mathbb{B}_{1,k_i}$ and $\mathbb{B}_{2,k_j}$.
The function we are trying to minimize in Eq.~(\ref{eq:argmin}) is directly correlated to the loss function utilized to train the neural network for the description extraction and can be described as:
\begin{equation}
    L(d_S,d_D) = \underbrace{||f(I_A)-f(I_S)||^2}_{d_S} - \underbrace{||f(I_A)-f(I_D)||^2}_{d_D} + \;c, 
\end{equation}
where $I_A,\,I_S$ and $I_D$ represent the anchor, similar and dissimilar depth images respectively, used for training with the triplet loss method. We also define $d_S$ as the Euclidean distances between the descriptors $\vec q_A$ from $I_A$ and the descriptors $\vec q_S$ from $I_S$. 
The same applies for $d_D$ and the descriptors $q_D$ from $I_D$. 
The loss function is designed so that similar and dissimilar point cloud pairs are pushed close together and far apart in the derived vector space. 
The parameter $c$ is a margin distance for distinguishing between similar and dissimilar pairs. 

\begin{figure}[!b]
    \centering
    \includegraphics[width=1.\columnwidth]{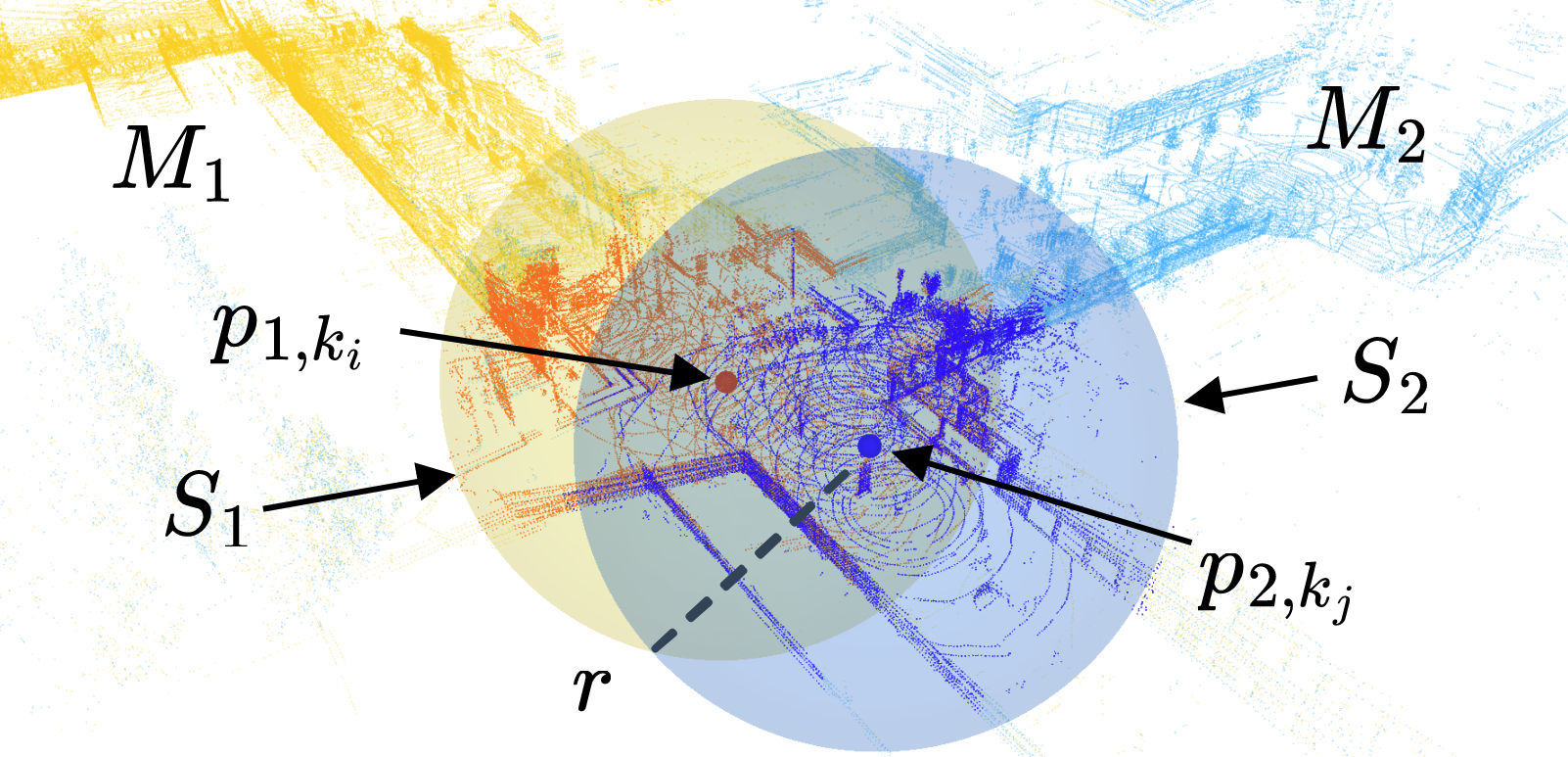}
    \caption{The merged map $\mathbf{M}$ after the alignment of $\mathbf{M}_1$ and $\mathbf{M}_2$, the spheres $\mathbf{S}_1$ and $\mathbf{S}_2$ and their corresponding centers $\mathbf{p}_{1,k_i}$ and $\mathbf{p}_{2,k_j}$ of the overlapping regions. The spheres encompass the highlighted points that are used as an input to the GICP algorithm.}
    \label{fig:spheres}
\end{figure}

\subsubsection{\textbf{Initial Transform}} \label{subsubsec:initial_transform}

In order to obtain the complete homogeneous initial transform $\mathbf{T}_0$, we utilize the orientation-specific vectors $w_{1,k_i}$ and $w_{2,k_j}$ as input for the \textit{orientation estimator} module, which predicts a yaw discrepancy angle $\delta \psi$. The \textit{orientation estimator} is a component of the descriptor extraction process and has been explained in greater detail in~\cite{schaupp2019oreos, stathoulopoulos20223deg}. The initial transform $\mathbf{T}_0$ is then generated to align the two local frames $\mathbb{B}_{1,k_i}$ and $\mathbb{B}_{2,k_j}$.
\begin{equation}
    \label{eq:T0}
    \mathbf{T}_0 = \left[ \begin{array}{cc}
         R_{z}(\delta \psi) & \mathbf{p}_{1,k_i}-\mathbf{p}_{2,k_j}\\
         O_{1\times3} & 1
    \end{array} \right]
\end{equation} 
Due to the fact that the translation component of Eq.~(\ref{eq:T0}) relies on the trajectory points, it is not possible to ensure that the two robots have explored precisely the same coordinates on the map. As a result, it is not feasible to obtain a perfectly accurate alignment of the two point cloud maps.

\subsubsection{\textbf{Refined Alignment}} \label{subsubsec:refined_alignment}

To attain an ultimate and enhanced alignment, the initial transform $\mathbf{T}_0$ is employed as a prior in the \texttt{fast\_gicp} registration algorithm~\cite{fast_gicp}, which leads to a considerably quicker convergence. The General Iterative Closest Point (GICP) algorithm~\cite{GICP} considers the estimation of the transformation matrix $\mathbf{T}$, which aligns two sets of points $\mathcal{A}=\{a_0, \ldots, a_N\}$ and $\mathcal{B}=\{b_0, \ldots, b_N\}$. The transformation error is defined as:
\begin{equation}
    \hat d_i = \hat b_i - \mathbf{T}\,\hat a_i,
\end{equation}
where each point was sampled as a Gaussian distribution $a_i \sim \mathcal{N}(\hat a_i, C^A_i), b_i \sim \mathcal{N}(\hat b_i, C^B_i)$. The distribution of $d_i$ is yielded by the reproductive property of the Gaussian distribution as:
\begin{equation} \label{eq:likelihood}
    d_i = \mathcal{N}(0, C_i^B + \mathbf{T}\,C_i^A\,\mathbf{T}^\mathsf{T})
\end{equation}
The GICP algorithm yields the transformation $\mathbf{T}$ that maximizes the logarithmic likelihood of Eq.~(\ref{eq:likelihood}):
\begin{equation}
    \mathbf{T} =  \operatorname*{arg\,min}_{\mathbb{T}} \sum d_i^\mathsf{T} (C_i^B + \mathbf{T}\,C_i^A\,\mathbf{T}^\mathsf{T})^{-1}\,d_i
\end{equation}
Additionally, to reduce computational time further, we utilize only a portion of the two point cloud maps as input to the registration algorithm. We identify the centers of the overlapping regions as points $\mathbf{p}_{1,k_i}$ and $\mathbf{p}_{2,k_j}$ and then sample the points inside two spherical regions $\mathbf{S}_1$ and $\mathbf{S}_2$ with radius $r$. These spherical regions are defined as:
\begin{equation} \label{eq:sampling}
    \mathbf{S} = \{ \mathbf{m}, \mathbf{p} \in \mathbb{R}^3 : ||\mathbf{m}-\mathbf{p}_k||^2 \leq r^2 \}
\end{equation}
As shown in Fig.~\ref{fig:spheres}, the alignment of the points within the spherical regions results in the creation of the final and merged global map.


\subsubsection{\textbf{Sphere Radius Selection}} \label{subsec:sphere_radius}

As mentioned earlier, only a portion of the point cloud maps is utilized to run the GICP algorithm, aiming to reduce computational time. The sampling radius $r$ from Eq.~(\ref{eq:sampling}) proves crucial for the successful convergence of the registration algorithm and has a subtle impact on computational times, as we will discuss in Section~\ref{sec:preliminary} and~\ref{sec:results}. Throughout the experimental evaluation, manual tuning becomes necessary for each environment due to variations in tunnel openings and in order to address this variability, we propose an adaptive sampling radius to automate the selection process. A straightforward approach involves directly utilizing the spaciousness $s_k$ defined for the adaptive keyframe sampling in Eq.~(\ref{eq:spaciousness}). However, this method would require an additional array to store the spaciousness corresponding to each pose accumulated in the trajectory array. Given that the map merging process assumes only the full point cloud map and the trajectory, lacking individual LiDAR scans for spaciousness calculation, we adopt an alternative strategy. Each robot can get the spaciousness for its self and then approximate the spaciousness of the other agent based on their trajectory, in order to avoid redundant data exchange. We use the \black{indices} selected for overlapping regions $k_i$ and $k_j$ from Eq.~(\ref{eq:argmin}), along with the poses corresponding to one step back, to determine the threshold set for that instance. With the threshold from Eq.~(\ref{eq:threshold}), we can approximate the spaciousness. Given the \black{indices} $k_i$ and $k_j$ from the trajectories $\mathbf{P}_1$ and $\mathbf{P}_2$, we extract the corresponding poses $\mathbf{p}_{1,k_{i-1}}, \mathbf{p}_{1,k_i}$ and $\mathbf{p}_{2,k_{j-1}}, \mathbf{p}_{2,k_j}$. We then calculate their distances, denoted as $l_1$ and $l_2$, as follows:

\begin{equation} \label{eq:distance}
l_1 = \left\{
            \begin{array}{rl}
                  ||\mathbf{p}_{1,k_i} - \mathbf{p}_{1,k_i+1}||_2, & \text{if }\; k_i = 0  \\
                  ||\mathbf{p}_{1,k_i} - \mathbf{p}_{1,k_i-1}||_2, & \text{else } \\
            \end{array}
        \right. 
\end{equation}
\begin{center}
    and
\end{center}
\begin{equation} 
l_2 = \left\{
            \begin{array}{rl}
                  ||\mathbf{p}_{2,k_j} - \mathbf{p}_{2,k_j+1}||_2, & \text{if }\; k_j = 0  \\
                  ||\mathbf{p}_{2,k_j} - \mathbf{p}_{2,k_j-1}||_2, & \text{else }\;  \\
            \end{array}
        \right.
\end{equation}

If the \black{indices} happen to be the first ones in the list, we use the next pose instead of the previous. After obtaining the distance between the poses, as per Eq.~(\ref{eq:threshold}), we can estimate the rough spaciousness. We define the corresponding radii for each threshold based on our experience and experimentation as follows:
\begin{equation} \label{eq:sphere_threshold}
r_{r} = \left\{
            \begin{array}{rl}
                 25 \text{m, } & \text{if }\; l_r \geq 10 \text{m } \\
                 15 \text{m, } & \text{if }\; l_r = 6 \text{m } \\
                 10 \text{m, } & \text{if }\; l_r = 3 \text{m } \\
                 2\,l_r \text{, } & \text{if }\; l_r < 3 \text{m }  \\
            \end{array}
        \right.
\end{equation}
To conclude this section, it is worth noting that the radius could be fixed at a larger value to cover most cases. However, we have observed rare instances where increasing the radius includes points from the map that may confuse the registration algorithm by introducing incorrect correspondences. In the map merging process, we sample the two spheres from the point cloud maps without using the LiDAR scans. This approach captures points that would otherwise be out of the sensor's field of view due to obstructions. These additional points can potentially lead to the registration algorithm failing.

\SetKwComment{Comment}{/* }{ */}

\begin{algorithm}[h!]
\caption{The overall FRAME algorithm}\label{alg:while_exploration}
\SetKwFunction{predict}{predict}
\SetKwFunction{Im}{Im}
\SetKwFunction{append}{.append}
\SetKwFunction{slam}{SLAM}
\SetKwFunction{junction}{junctionDetected}
\SetKwFunction{communication}{communicationEstablished}
\SetKwFunction{kd}{kdTree}
\SetKwFunction{query}{.query}
\SetKwFunction{argmin}{argmin}
\SetKwFunction{orientation}{yawDiff}
\SetKwFunction{extract}{extractSphere}
\SetKwFunction{rot}{Rot}
\SetKwFunction{tran}{Tran}
\SetKwFunction{gicp}{fast\_gicp}
\SetKwFunction{radius}{adaptiveRadius}
\SetKwFunction{condition}{condition}

\KwData{$P,\,I,\,\mathbf{p}$ \Comment*[r]{pcd, image, pose}} 
\KwResult{$\mathbf{Q},\,\mathbf{W},\,\mathbf{M},\,\mathbf{P}$\Comment*[r]{vector stacks, map and trajectory}}
\While{exploring}{
    $P_i \gets$  \slam{$lidar$,\,$imu$}\Comment*[r]{registered scan}
    $S_i \gets ||P_i||$\Comment*[r]{mean Euclidean distance for every point}
    $s_i \gets \alpha s_{i-1} + \beta S_i$\;
    $th_i \gets$ \condition{$s_i$}\;
  \If{$\;||\mathbf{p}_{1,i} - \mathbf{p}_{1,i-1}||_2 \geq th_i\;$}{
    $I_{i} \gets$ \Im{$P_{i}$}\Comment*[r]{pcd projection}
    $q_{1,i}, w_{1,i} \gets$ \predict{$I_{i}$}\Comment*[r]{extract descriptors}
    $\mathbf{Q}_1$\append{$q_{1,i}$}\;
    $\mathbf{W}_1$\append{$w_{1,i}$}\;
    $\mathbf{M}_1$\append{$P_{i}$}\;
    $\mathbf{P}_1$\append{$\mathbf{p}_{1,i}$}\;
  }{\If{ \junction{} or \communication{} }{
      $tree \gets$ \kd{$\mathbf{Q}_1$}\;
      $dist,\,ind \gets tree$\query{$\mathbf{Q}_2$}\;
      $k_i \gets$ \argmin{$dist$}\Comment*[r]{get index of minimum distance pair}
      $k_j \gets ind_{k_i}$\;
      $\delta\psi \gets$ \orientation{$ \mathbf{W}_{1,k_i}, \mathbf{W}_{2,k_j}$}\Comment*[r]{orientation estimation}
      $r_1 \gets$ \radius{$\mathbf{P}_1,k_i$}\Comment*[r]{get adaptive radius for each sphere}
      $r_2 \gets$ \radius{$\mathbf{P}_2,k_j$}\;
      $\mathbf{S}_1 \gets$\extract{$\mathbf{M}_1,$$r_1$}\;
      $\mathbf{S}_2 \gets$\extract{$\mathbf{M}_2,$$r_2$}\;
      \rot{$\mathbf{T}_0$} $\gets \delta\psi$\Comment*[r]{initial rotation}
      \tran{$\mathbf{T}_0$} $\gets [\,\mathbf{p}_{1,k_i}-\mathbf{p}_{2,k_j}\,]$\Comment*[r]{initial translation}
      $\mathbf{T} \gets$\gicp{$\mathbf{S}_1,\mathbf{S}_2,\mathbf{T}_0$}\Comment*[r]{GICP with initial guess}
      $\mathbf{M}_2 \gets  \mathbf{T} \, \mathbf{M}_2$
    }
  }
}
\end{algorithm}


\section{Preliminary Evaluation} \label{sec:preliminary}

In this section, we initiate a preliminary assessment using the publicly available KITTI dataset~\cite{Geiger2012CVPR}, setting the stage for subsequent experimentation in scaled-down subterranean environments and, ultimately, real-world mines. Our initial focus centers on evaluating the place recognition performance of various frameworks, including the one employed in our field evaluation. We provide valuable insights into the rationale behind our choice of descriptors, while at the same time highlighting that our map merging framework boasts modularity, accommodating arbitrary descriptors. To demonstrate this versatility, we employ the OverlapTransformer~\cite{ma2022overlaptrans} to seamlessly merge two maps from the KITTI dataset. Following the analysis of the place recognition performance, we delve into a runtime analysis. This analysis is of high importance, considering the vital role played by the description extraction and querying process in the computational time of the overall framework. Concluding this preliminary evaluation, we turn our attention to tests aimed at \black{understanding} the limits of the \texttt{fast\_gicp} algorithm. This entails evaluating performance across varying yaw discrepancies, an expanding sphere sampling radius, and increasing distances between selected pose candidates. This multifaceted evaluation approach not only contributes to a comprehensive understanding of the algorithm's capabilities and limitations but also facilitates an understanding of how the performance of preceding components influences the inputs to this algorithm in real-world scenarios.

\subsection{Place Recognition Analysis}~\label{subsec:place_recognition}

The place recognition analysis compares state-of-the-art place recognition frameworks, that includes, \black{learning-based} algorithms that use a projected point cloud as a 2D range image, as well as others that incorporate multiple cues like range images, intensity images, normals images and semantic images. Additionally, we explore frameworks leveraging the latest advancements in deep learning, specifically the transformative capabilities of transformers~\cite{attentionTrans}. Classical handcrafted descriptors are also considered in this comparative evaluation. The decision to use 2D range images as input, instead of directly employing 3D point clouds, is grounded in the current higher computational demands associated with the latter representation. This consideration aligns with our aim to deploy these algorithms \black{on} mobile robots, including UAVs, where efficiency is paramount, and real-time processing on CPU-only computational units is a key requirement. For our evaluation, we follow established practices, drawing inspiration from methodologies outlined in notable works such as~\cite{schaupp2019oreos, chen2020overlapnet, ma2022overlaptrans}. The training data is sourced from the KITTI odometry benchmark~\cite{Geiger2012CVPR}, providing LiDAR scans captured by a Velodyne HDL-64E in urban areas around Karlsruhe, Germany. Our experimental setup mirrors that of the aforementioned articles, utilizing sequence 00 for evaluation, sequences 03-10 for training, and sequence 02 for validation, thereby ensuring alignment with recognized benchmarks in the field.

\begin{figure*}[t!]
    \centering
    \includegraphics[width=\textwidth]{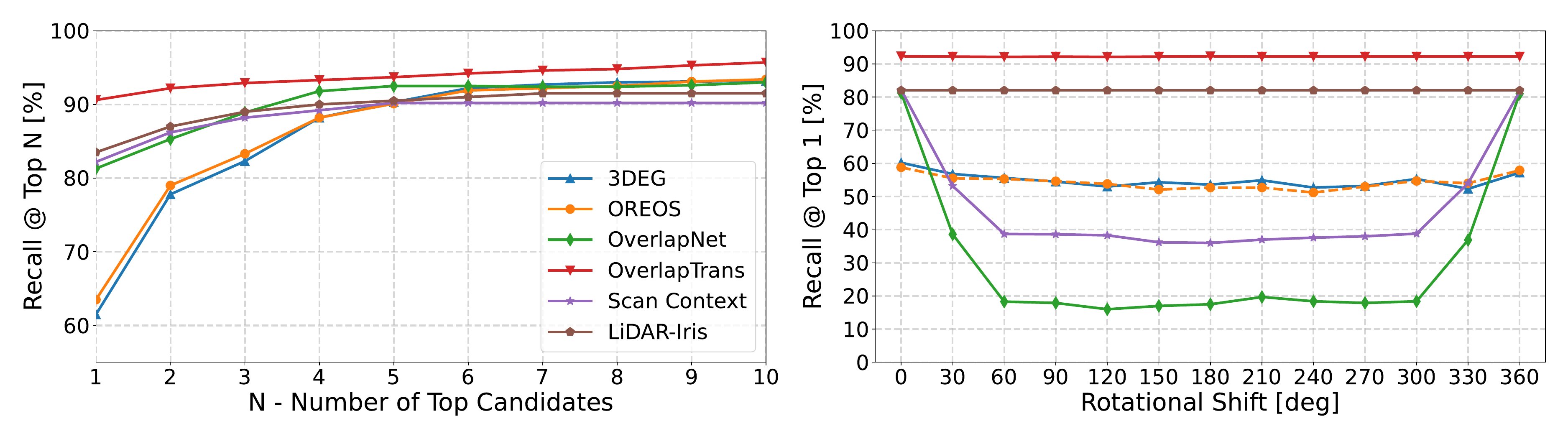}
    \caption{Place recognition results on the KITTI dataset: The left side illustrates the recall percentage for an increasing number of N candidates, while the right side depicts the recall percentage for the top 1 candidate across an increasing rotational shift. The coloring scheme shown on the legend of the left plot holds for both figures.}
    \label{fig:top-k}
    \vspace{-0.5cm}
\end{figure*}

Commencing with Fig.~\ref{fig:top-k}, the recall percentages for the top candidates across various algorithms are presented, showcasing their respective capabilities in place recognition, as well as the performance for the top 1 candidate when we induce rotational shift to the querying scan, testing their yaw invariance capabilities. Notably, the state-of-the-art OverlapTransformer~\cite{ma2022overlaptrans} stands out with superior performance, while OREOS~\cite{schaupp2019oreos} and 3DEG~\cite{stathoulopoulos20223deg} exhibit comparatively lower performance, particularly for the top 1 candidate. Among the non-learned based frameworks, LiDAR-Iris~\cite{wang2020iris} and Scan Context~\cite{giseop2018sc} demonstrate creditable performance in candidate retrieval, although Scan Context falls short in achieving yaw invariance. At this point, it is worth noting that according to the corresponding authors, there are limitations in generalizing to narrower environments with repeated structures, such those present in the Complex Urban LiDAR dataset~\cite{jjeong-2019-ijrr}, akin to challenges posed by subterranean environments. These limitations arise from the failure of the projection and final encoding stages to capture sufficient details. OverlapTransformer, while excelling in the KITTI dataset, presents challenges for subterranean applications. Its reliance on large datasets for effective training, as a transformer based network~\cite{LIN2022111}, coupled with the lack of open-source datasets for subterranean environments, hinders its applicability. Additionally, the absence of a yaw discrepancy module poses a drawback, a crucial element we will elaborate on later in the context of successful map merging processes. OverlapNet~\cite{chen2020overlapnet}, in contrast, is a Fully Convolutional Network (FCN). While it relies on semantic information, which may be limited or less useful in subterranean environments with sparse semantics, the authors have showcased its commendable performance using solely range images in their ablation study. However, a drawback lies in its limited robustness in scenarios involving rotational shift. While OREOS may not exhibit the best performance, it stands out for maintaining yaw invariance, and its network architecture is notably lightweight. Building upon this foundation, 3DEG extends the work by incorporating topological semantic information. As demonstrated by the authors, this extension enhances performance in subterranean environments and introduces a module that serves as a triggering device for the map merging process. This module identifies junctions, which are inherently more feature-rich than straight, repeating tunnels, contributing to the overall effectiveness of the map merging framework.

\begin{figure*}[b!]
    \begin{subfigure}{0.25\textwidth}
    \includegraphics[width=\textwidth]{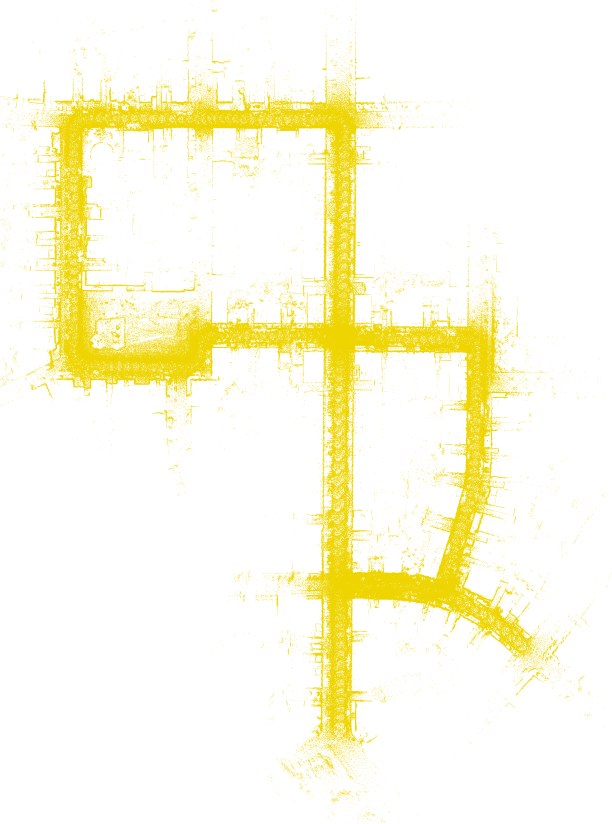}
    \caption{seq. 00 $|$ scans $0000-1700$}
    \end{subfigure}
    \begin{subfigure}{0.36\textwidth}
    \includegraphics[width=\textwidth]{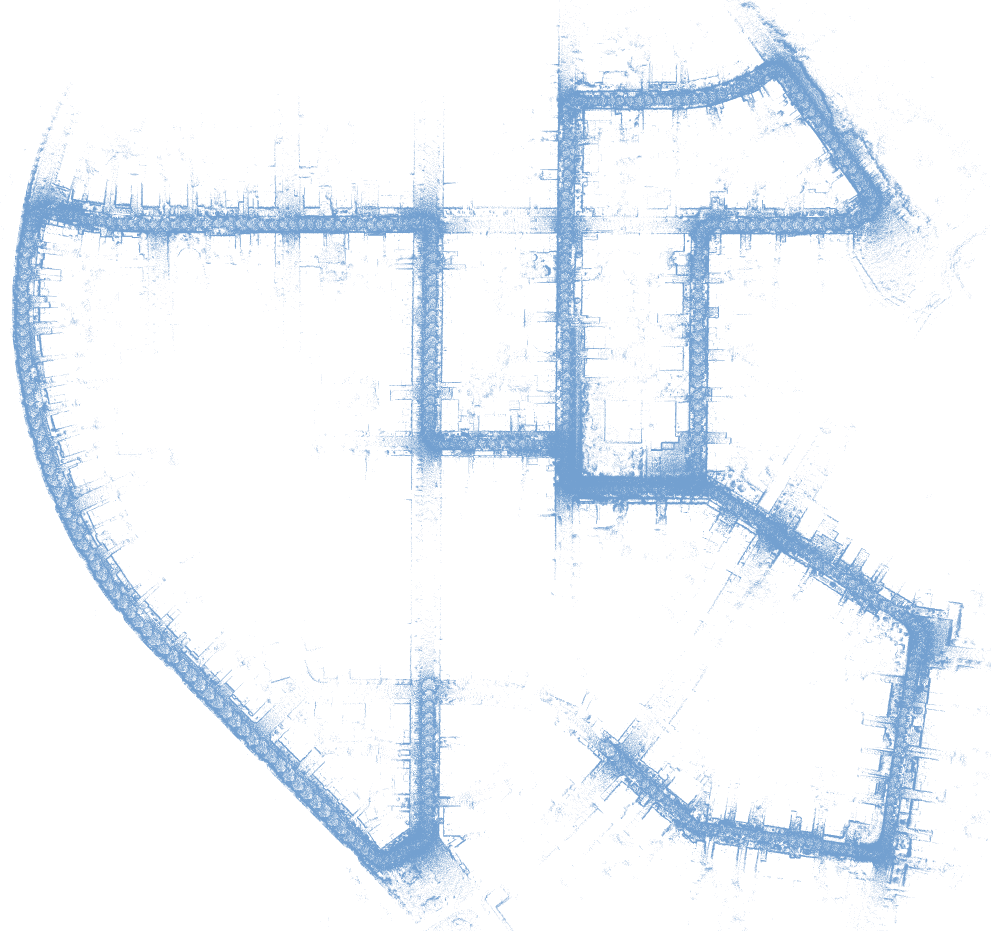}
    \caption{seq. 00 $|$ scans $1701-4540$}
    \end{subfigure}
    \begin{subfigure}{0.39\textwidth}
    \includegraphics[width=\textwidth]{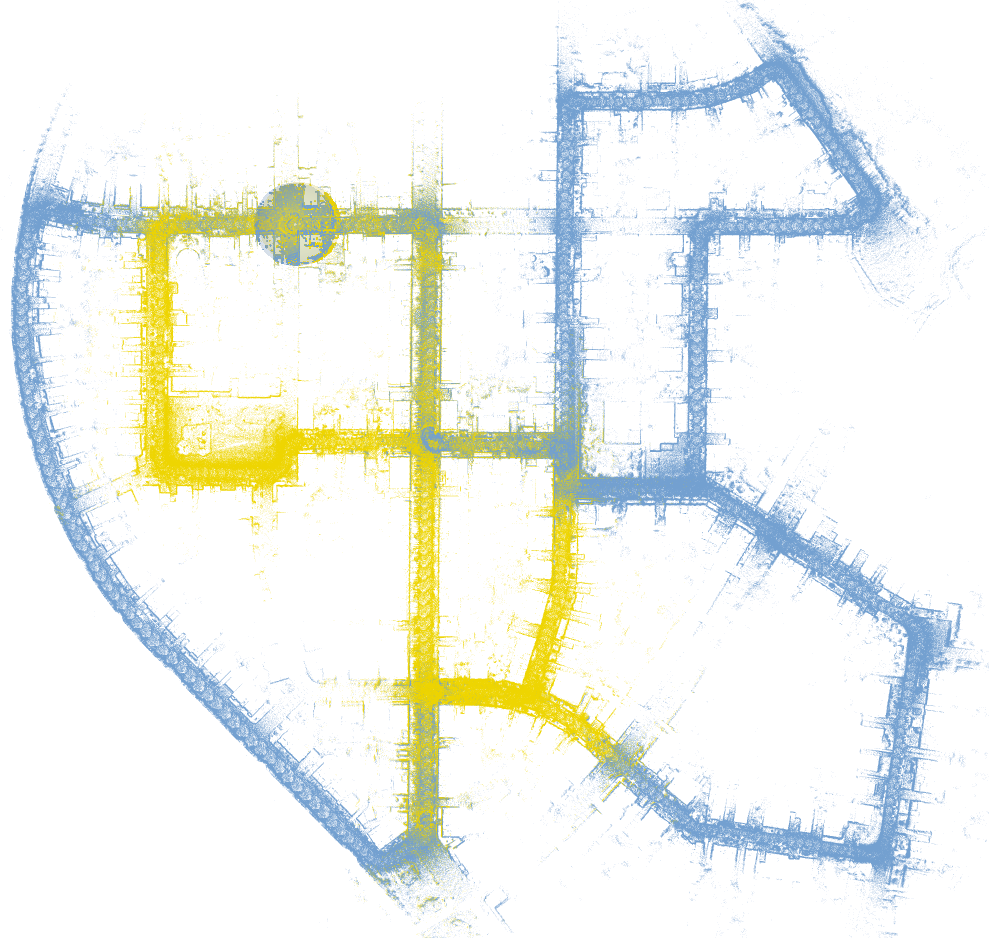} 
    \caption{seq. 00 $|$ Merged}
    \end{subfigure}
    \caption{The FRAME algorithm on the 00 sequence of KITTI using the OverlapTransformer descriptors to identify the overlapping regions. (a) The first map $\mathbf{M}_1$ is made of the first 1700 sequences. (b) The second map $\mathbf{M}_2$ is consisted of the rest of the scans. (c) The final merged map $\mathbf{M} = \mathbf{M}_1 \cup {}^1\mathbf{T}_{2} \mathbf{M}_2$.}
    \label{fig:kitti}
\end{figure*}
Furthermore, and prior to delving into the runtime per component analysis, which further illuminates the performance and suitability of each descriptor for map merging scenarios, we underscore the modularity of our framework to accommodate any descriptor architecture. Leveraging the excellent performance of OverlapTransformer, we showcase its application in merging two maps from sequence 00 of the KITTI dataset. For this demonstration, we divide the sequence into two segments: the first map comprises scans from 0000 to 1700, totaling $34.33\cdot10^6$ points, while the second map encompasses scans from 1701 to 4540, totaling $57.75\cdot10^6$ points. These two maps capture different perspectives of the city with multiple overlaps. Extracting descriptors for each scan, we then query the two vector sets to identify overlapping regions with the minimum distance between the descriptors. Then, employing the automated radius selection, the radius is set at $r=25$ \unit{meters}, and finally the two overlapping spheres are aligned and the transformation matrix $\mathbf{T}$ is derived. Fig.~\ref{fig:kitti} visually portrays both maps and the final merged map.

\subsection{Runtime Analysis}~\label{subsec:runtime}

\begin{table*}[b!]
\centering
\caption{Per-component runtime analysis of various frameworks, for the 00 sequence of the KITTI dataset. The analysis includes the time taken to extract the descriptors, including any preprocessing steps, the query time to find the overlapping regions, the sphere sampling time to extract the submaps and finally the time for the GICP registration.}
\label{table:runtime}
\resizebox{0.85\textwidth}{!}{%
\begin{tblr}{
  row{even} = {},
  row{3} = {},
  row{5} = {},
  row{7} = {},
  column{even} = {c},
  column{3} = {c},
  column{5} = {c},
  cell{1}{2} = {},
  cell{1}{3} = {},
  cell{1}{4} = {},
  cell{1}{5} = {},
  cell{1}{6} = {},
  cell{2}{4} = {r=6}{},
  cell{2}{5} = {r=6}{},
  hline{1-2,8} = {-}{0.08em},
}
\textcolor{black}{} & Descriptor Extr. [ms]  & Query [ms]        & Sphere Sampling [ms] & GICP [ms] & Total [ms]        \\
LiDAR-Iris~\cite{wang2020iris}         & 74.49 + 61.36          & 16.76 $\cdot10^3$ & 1271.01       & 0.132          & 18.17 $\cdot10^3$ \\
Scan Context~\cite{giseop2018sc}       & 84.99 + 93.63          & 857.91            &               &                & 2307.67           \\
OverlapTrans~\cite{ma2022overlaptrans}       & 30.16 + 24.95          & 1139.17           &               &                & 2465.42           \\
OverlapNet~\cite{chen2020overlapnet}         & 74.11 + 18.63          & 65.70 $\cdot10^3$ &               &                & 66.97 $\cdot10^3$ \\
OREOS~\cite{schaupp2019oreos}              & \textbf{19.19 + 12.43} & \textbf{272.94}   &               &                & \textbf{1575.70}  \\
3DEG~\cite{stathoulopoulos20223deg}               & \textbf{15.52 + 14.36} & \textbf{289.07}            &               &                & \textbf{1590.09}  
\end{tblr}
}
\end{table*}

The per-component analysis offers a better understanding of the computational performance of each module and its suitability for the map merging process. Given our objective of developing a solution capable of online execution on mobile robots, reliant solely on a CPU, the results presented in Table~\ref{table:runtime} stem from the earlier-described evaluation on the KITTI dataset. These evaluations were conducted on an 11th Gen Intel® Core™ i7-1165G7 @ 2.80GHz CPU, specifications akin to the onboard computers of the mobile robots employed in our field evaluations.
The reported results encompass various time metrics, including the time required for descriptor extraction — incorporating any preprocessing steps, denoted in the format of $x+y$, where $x$ represents preprocessing and $y$ corresponds to descriptor extraction. Additionally, the analysis includes query time for searching between the two vector sets to identify overlapping regions, sphere sampling time for extracting submaps, the time required for the GICP algorithm to align the maps, and finally, the total time in milliseconds. These insights contribute to the assessment of each component's efficiency in achieving the real-time processing goals essential for practical deployment on mobile robots.

Examining the non \black{learning-based} descriptors, particularly in terms of descriptor extraction times, reveals higher computational requirements in both the preprocessing step and the extraction time. This is primarily due to the manual handcrafting nature of these descriptors. Scan Context and LiDAR-Iris, for instance, involve partitioning the point cloud into rings and sectors, calculating azimuthal and radial bins. LiDAR-Iris introduces additional steps, including a Fourier transform for translation invariance and LoG-Gabor filters~\cite{fischer2007loggabor} for binary feature extraction. Notably, OverlapNet experiences an increase in preprocessing time due to the necessity of calculating the normals of the point cloud to create the normals image. In contrast, OverlapTransformer demonstrates relatively low processing times, noteworthy given its CPU execution. Its three-step process involves a FCN layer, a transformer, and an MLP+NetVLAD layer to generate global descriptors. Moving to OREOS and 3DEG, they exhibit the lowest computational times, aligning with expectations due to their relatively simple architectures. Both employ a Convolutional Neural Network (CNN) for feature extraction and a Multi-Layer Perceptron (MLP) for the yaw regression. Remarkably, all these frameworks boast processing times good for deployment in the map merging process. They operate at frequencies ranging from 5 to 30 \unit{Hz}, which is more than enough, considering the speed of a mobile robot (1 to 3~\unit[m$/$s]) and the sampling distance thresholds outlined in subsection~\ref{subsubsec:keyframe_sampling}.

Transitioning to the querying times, a critical aspect differentiating our map merging approach from traditional place recognition becomes apparent. While place recognition involves a one-to-all comparison, where the current scan is matched against all previous scans in the querying stack, the map merging process adopts an all-to-all approach. Here, every scan in one stack must be compared to all scans in the other stack. The efficiency of this process is crucial, and inefficient querying can significantly escalate computational times, as evident in the case of LiDAR-Iris and OverlapNet. LiDAR-Iris proves inefficient due to its binary feature-saving approach in a database, requiring Hamming distance calculations and manual threshold setting for loop closures. This becomes time-consuming when comparing every combination between vector stacks. Scan Context improves the search time by employing a two-phase search algorithm using ring keys, along with a \textit{k}-d tree for querying. On the other hand, OverlapNet demonstrates inefficiency as it runs every pair of samples through the delta head for similarity score comparison, a time-consuming process in an all-to-all comparison scenario. In addition, the dimensionality of the data used for querying is a notable consideration. OverlapTransformer, for instance, offers two options for querying. First, it allows the computation of similarity between two descriptors, resulting in a slower total querying time of 4.04~\unit{seconds}. Alternatively, it can create a \textit{k}-d tree for rapid search, taking advantage of its compact $1\times256$ descriptor, totaling at a 1.1~\unit{seconds} as denoted in Table~\ref{table:runtime}. In contrast, OverlapNet's larger $1\times360\times128$ multi-dimensional vector is not as efficient and practical for constructing a \textit{k}-d tree. The \textit{k}-d tree requires $n \times m$ dimensional data, where $n$ is the number of samples and $m$ is the number of features. OREOS and 3DEG, with their small $1\times64$ feature vectors, support fast querying with a \textit{k}-d tree, resulting in the fastest querying times. Their efficient implementation makes them well-suited for the map merging process, particularly in scenarios where quick and accurate comparisons between vector stacks are paramount.

Concluding the per-component runtime analysis, it is noteworthy that the sphere sampling process and the GICP are inherently independent of the descriptors used. However, the sphere sampling step can currently be considered a bottleneck. Extracting submaps from large point cloud maps involves an initial voxel downsampling of the point clouds, with this step demanding the most time within the entire process. Opting not to downsample the point cloud maps would shift the bottleneck to the distance checking phase for feature extraction, as the volume of points becomes too cumbersome to handle. Typically, a voxel downsampling factor of 0.5 is employed, striking a balance between achieving dense enough submaps and aiding the GICP algorithm in delivering both fast and accurate results. Ultimately, the fastest times for the entire map merging process hover just above 1.5 seconds. Given the substantial number of points and the expansive scale of the dataset, this processing speed can be deemed fast and efficient.

\subsection{GICP - Failure Case Analysis}

\begin{figure*}[!b]
    \centering
    \includegraphics[width=\textwidth]{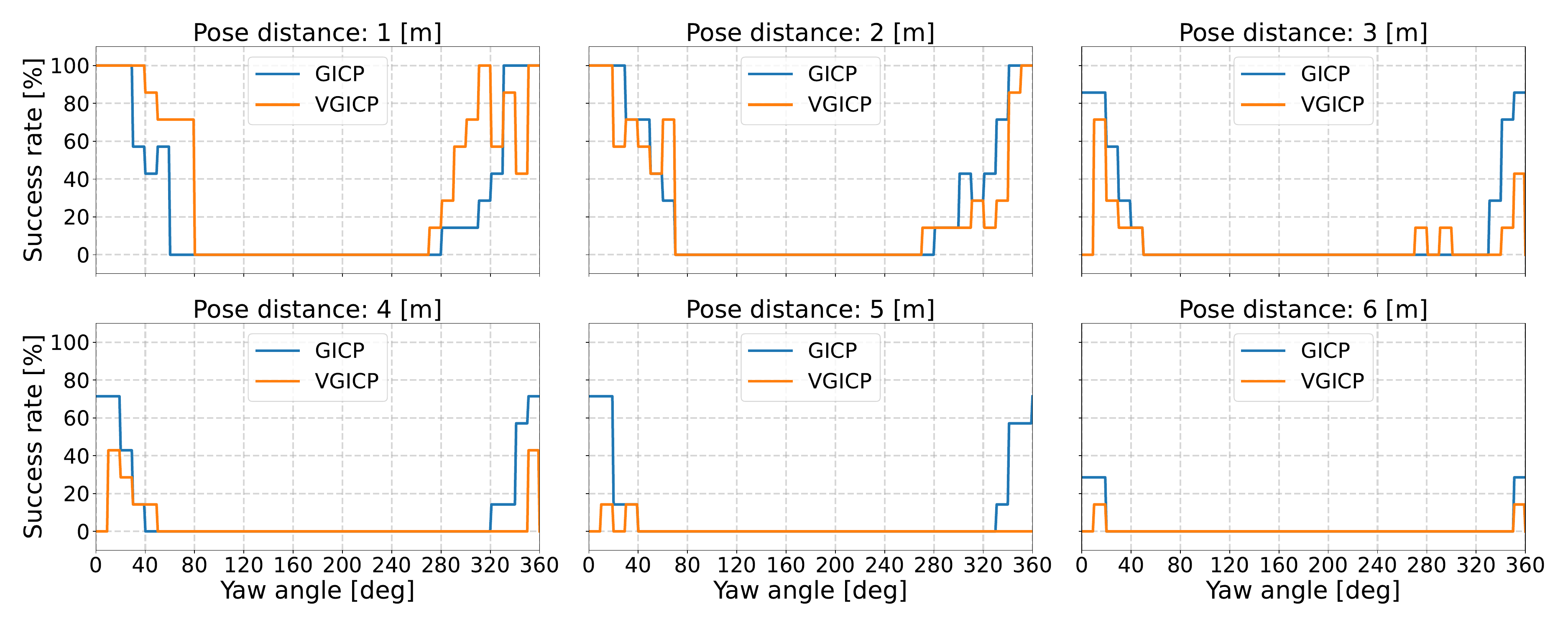}
    \caption{Success rate for the GICP and VGICP algorithms as rotational shift in the yaw axis between the two point clouds is introduced. Subfigures indicate the result for an increasing distance between the point cloud centers.}
    \label{fig:gicp_angles}
\end{figure*}
\begin{figure*}[!t]
    \centering
    \includegraphics[width=\textwidth]{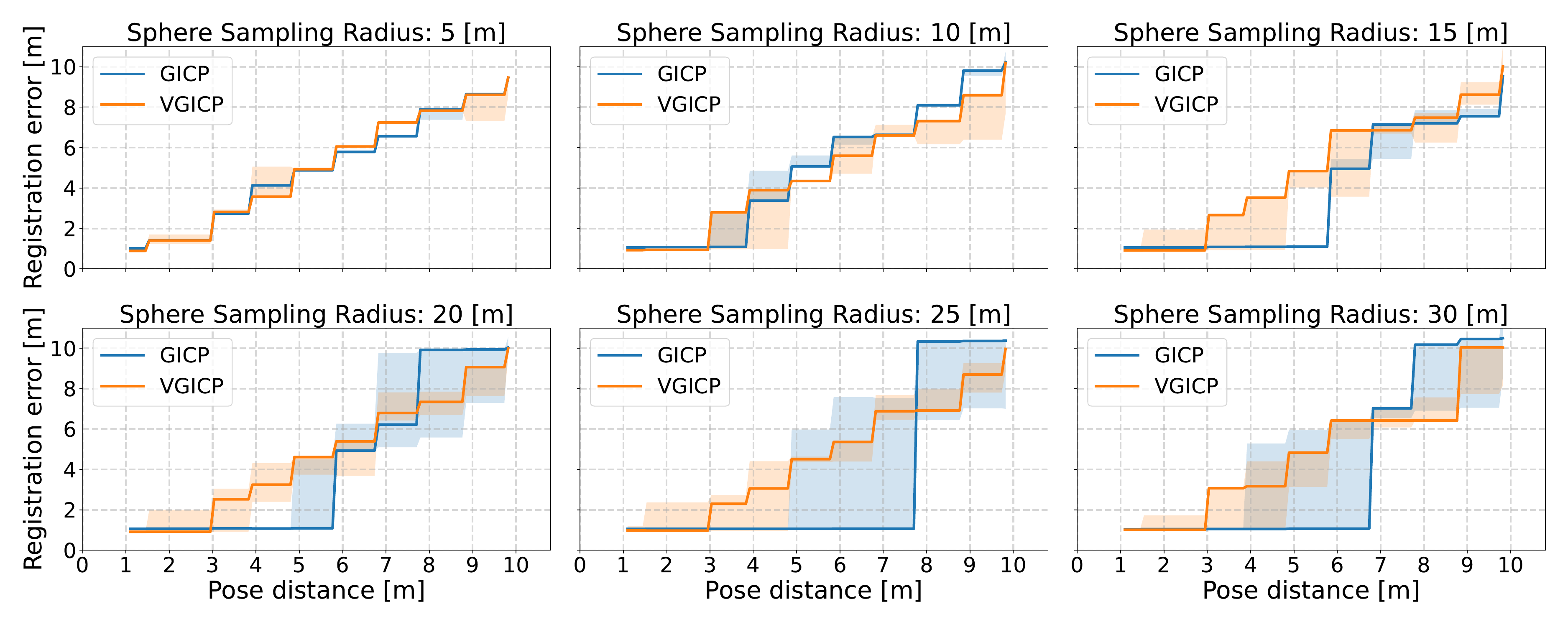}
    \caption{Point cloud registration error for the GICP and VGICP algorithms as the distance between point cloud centers increases. Subfigures indicate varying sphere sampling radii, with colored segments representing min and max deviations for different rotational shifts imposed on the second point cloud.}
    \label{fig:gicp_errors}
    \vspace{-0.6cm}
\end{figure*}
\begin{figure*}[!b]
    \centering
    \includegraphics[width=\textwidth]{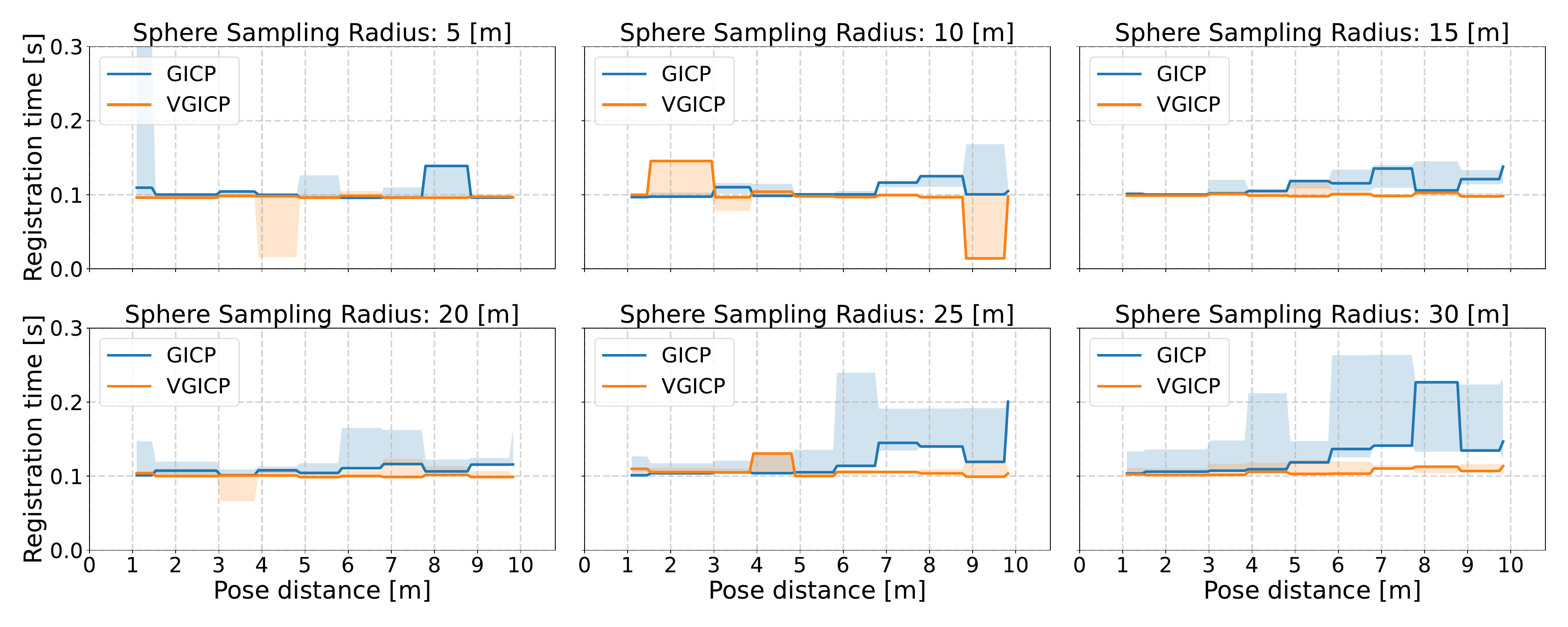}
    \caption{Processing time for the GICP and VGICP algorithms as the distance between point cloud centers increases. Subfigures indicate varying sphere sampling radii, with colored segments representing min and max deviations for different rotational shifts imposed on the second point cloud.}
    \label{fig:gicp_times}
\end{figure*}

\black{In this subsection, we conduct a thorough analysis of the General Iterative Closest Point (GICP) algorithm, which plays a pivotal role as the final component in the pipeline. Our objective is to systematically explore the algorithm's limitations and boundaries to gain insights into performance expectations for preceding components. This investigation is crucial for ensuring overall accuracy and provides valuable guidance for optimizing earlier stages of the map merging process.
We utilize implementations of both GICP and Voxelized General Iterative Closest Point (VGICP) provided by the \texttt{fast\_gicp} package\footnote{\url{https://github.com/SMRT-AIST/fast_gicp.git}}. Leveraging the KITTI dataset's 00 sequence, and employing the OverlapTransformer to identify overlapping regions denoted by trajectory indices $k_i$ and $k_j$, we begin with corresponding poses $\mathbf{p}_{1,k_i}$ and $\mathbf{p}_{2,k_j}$ approximately 1 meter apart.
To generate the plots illustrated in Fig.\ref{fig:gicp_errors} and Fig.\ref{fig:gicp_times}, we maintain the pose $\mathbf{p}_{1,k_i}$ while gradually increasing the $k_j$ index, incrementally expanding the distance between poses to up to 10 meters. This systematic sampling of poses aims to identify the threshold beyond which the GICP algorithm fails to accurately align due to a lack of overlapping features. Results presented in the figures are averages from various overlapping regions and index pairs.
Subplots in the figures correspond to increasing sphere sampling radii, which significantly influence the amount of information available in submaps. Additionally, in Fig.~\ref{fig:gicp_angles}, we assess GICP and VGICP performance in handling rotational shifts between input point clouds. These experiments provide insights into the robustness and limitations of the GICP algorithm under various conditions, laying the groundwork for refining the map merging process.}

Starting from Fig~\ref{fig:gicp_angles}, we evaluate the robustness of the GICP and VGICP algorithms in the presence of rotational shifts along the yaw axis. The plots illustrate the average success rate across all sphere sampling radii, ranging from 5 to 30 \unit{meters}. Examining Fig.~\ref{fig:gicp_angles} alongside Fig.~\ref{fig:gicp_errors}, we observe that GICP can effectively handle angle discrepancies of up to 20-30 \unit{degrees} when poses are within 3-6 \unit{meters}. Similarly, VGICP demonstrates the ability to manage the same rotational shifts but for poses with smaller distances between them. Moving on to Fig.~\ref{fig:gicp_errors} and Fig.~\ref{fig:gicp_times}, GICP exhibits greater robustness to increasing distances compared to VGICP, which struggles with distances exceeding 3 \unit{meters}. Notably, the 5-\unit{meter} radius fails to provide sufficient information for correct point cloud alignment as the distance increases beyond a \unit{meter}. The 25-\unit{meter} radius showcases the best performance, allowing for shifts up to approximately 8 \unit{meters}, followed by a slight drop to 5.5-6.5 \unit{meters} for the 20 and 30-\unit{meter} radii. These findings align with the discussion in subsection~\ref{subsec:sphere_radius}, emphasizing that excessively high radii may introduce confusion to the registration algorithm by introducing incorrect correspondences. Shaded segments in the plots denote the minimum and maximum deviations for added rotational shift, reaching up to 30 \unit{degrees}, as demonstrated in Fig.~\ref{fig:gicp_angles}. Concerning processing times (Fig.~\ref{fig:gicp_times}), they are maintained low as long as the algorithms reach convergence. Although some scenarios show a slight increase in times with an expanding radius, it does not amount to a significant delay. These results highlight the robustness and broader margin for the distance between queried poses offered by GICP. It is evident that the yaw estimation module in the map merging pipeline needs to regress the yaw discrepancy to less than 20-30 \unit{degrees} and the queried poses need to be within 3-5 \unit{meters}, for the GICP to deliver a fast and accurate result.

\subsection{Place Recognition Descriptor Choice}

Concluding the preliminary evaluation and transitioning to the field experiments, it is important to elaborate on our choice of place recognition method. Throughout this section, we emphasized various properties that descriptors must possess to seamlessly integrate into the map merging pipeline. Summarizing these considerations, the descriptors should be single-dimensional vectors to facilitate efficient \textit{k}-d tree querying, preventing unwanted delays during the matching phase. Additionally, the presence of a yaw regression module is essential, as demonstrated in our analysis, to ensure the GICP algorithm produces accurate results. The identified limits for the distance between two poses, approximately 5 \unit{meters}, are not an issue for what most place recognition frameworks offer. Examining the descriptors discussed and compared in subsection~\ref{subsec:place_recognition} and~\ref{subsec:runtime}, handcrafted methods prove unsuitable for subterranean environments due to their inability to capture sufficient details and features in the narrow spaces. OverlapNet's inefficient querying process and reliance on iterative matching through the delta head render it unsuitable for our map merging application. OverlapTransformer, while excelling in large-scale urban environments, lacks a yaw regression module, and the extensive data requirements for training a transformer pose challenges in subterranean scenarios with limited available data. On the other hand, OREOS and 3DEG, while not achieving the highest recall, offer fast querying times and include a yaw regression module. The added benefit of 3DEG lies in its incorporation of a classification module, capable of enhancing performance and triggering the map merging process. To further improve performance in such scenarios, coupling LiDAR data with other types of data is a viable option. For instance, the authors in~\cite{stathoulopoulos2023redundant} demonstrated the incorporation of Wi-Fi data, providing a redundant and more robust performance, particularly relevant in subterranean environments and modern mines equipped with Wi-Fi nodes. The proposed map merging framework stands out for its modularity, allowing for interchangeable descriptors. This flexibility accommodates various methods to enhance place recognition performance, depending on the specific scenario and application requirements.


\section{Field Evaluation Set-up} \label{sec:setup}

Before delving into the experimental field evaluation, offering a comprehensive analysis of the proposed map merging method, we introduce the various real-world scenarios where our experiments took place, along with the robotic platforms employed. In this section, we also introduce the metric error used for evaluation, provide a brief overview of the descriptor extraction method's training process, and outline the specifications of the computing hardware utilized.

\subsection{Subterranean Environments}

\begin{figure*}[t!]
    \centering
    \includegraphics[width=0.95\linewidth]{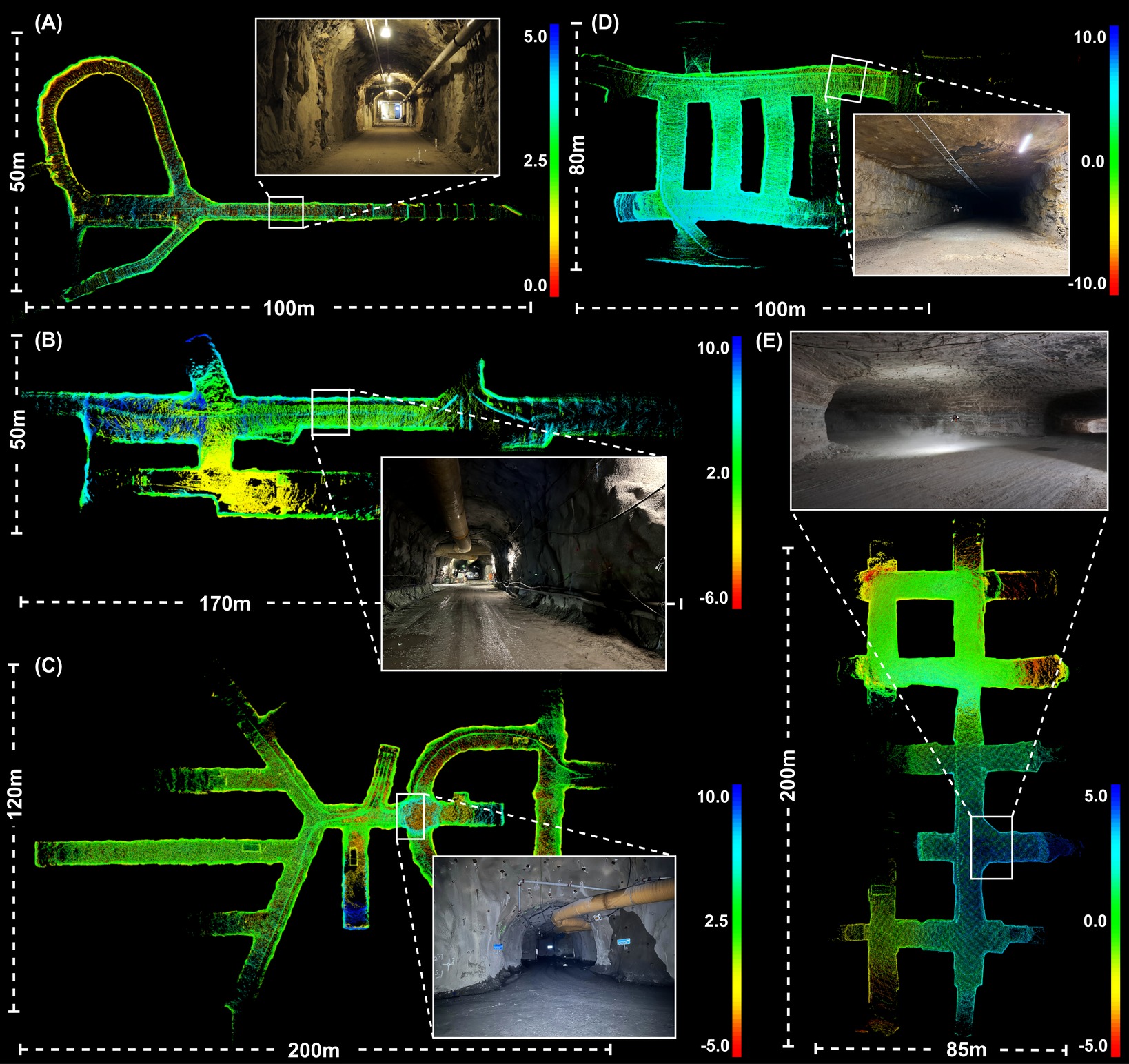}
    \caption{(A) Underground tunnel system, (B) Construction site at new metro line, (C) Testing mining site, (D) Subterranean mining site and (E) Large scale underground mine.} \label{fig:tunnels}
    \vspace{-0.6cm}
\end{figure*}

In order to test the performance of the proposed map merging approach in real-life conditions, where sensor noise, GNSS-denied localization, self-similarity and dust occur, it was evaluated in a series of large scale field experiments, in five different real-world, harsh subterranean environments. As a starting point, the first environment~\cite{KOVAL2022104168} is from an underground tunnel located in Lule\r{a}, Sweden, as seen on Fig~\ref{fig:tunnels}A and discussed in subsection~\ref{subsubsec:mjolk}. It consists of a long main entrance tunnel that leads to a wider junction with three branches that loop around and connect, making it the appropriate area to start the evaluation. For the second experiment, we already start evaluating in real operational conditions. The second environment is depicted on Fig.~\ref{fig:tunnels}B and discussed in subsection~\ref{subsubsec:NCC}, and it is a construction site from a new metro station in Stockholm, Sweden. This tunnel features wide passages ranging from $8-10$ \unit{meters}, as the two branches, upper and lower, of the area are explored. The third environment, to try out, is split between two different areas. Both areas are from the LKAB underground mining site in Kiruna, Sweden, depicted on Fig.~\ref{fig:tunnels}C, the largest underground iron mine in the world. The first one is part of the main Kiruna mine and consists of two highly similar areas where the goal is to challenge the ability to detect overlapping areas, discussed in subsection~\ref{subsubsec:lkab_drift}, while the second one is part of the Konsuln test mine and covers a larger and more complex area, providing more insights about the scalability of the algorithm, discussed in subsection~\ref{subsubsec:lkab_multi}. The next environment is again split into two different evaluation tests, shown on Fig.~\ref{fig:tunnels}D. Both are from the Epiroc underground test-mine facility in \"{O}rebro, Sweden, and unlike the previous environments, it features wider, squared tunnels, up to $10-12$ \unit{meters} wide, with multiple junctions and featureless, self-similar walls, discussed in subsection~\ref{subsubsec:epiroc}. The second part of this field test, in particular, offers the opportunity to test the merging of more than two maps, where some of them do not have overlapping trajectories and are limited to overlapping point clouds, discussed in subsection~\ref{subsubsec:epiroc_multi}. The final evaluation field test took place in the K+S operational mining facility located in Germany, Fig.~\ref{fig:tunnels}E, offering similar structured tunnels as the previous one but in greater scale, providing once more the opportunity to evaluate the scalability of the algorithm as discussed in subsection~\ref{subsubsec:kps}.

\subsection{Robotic Platforms}

\begin{figure*}[!t]
    \begin{subfigure}{0.23\textwidth}
    \includegraphics[width=\textwidth]{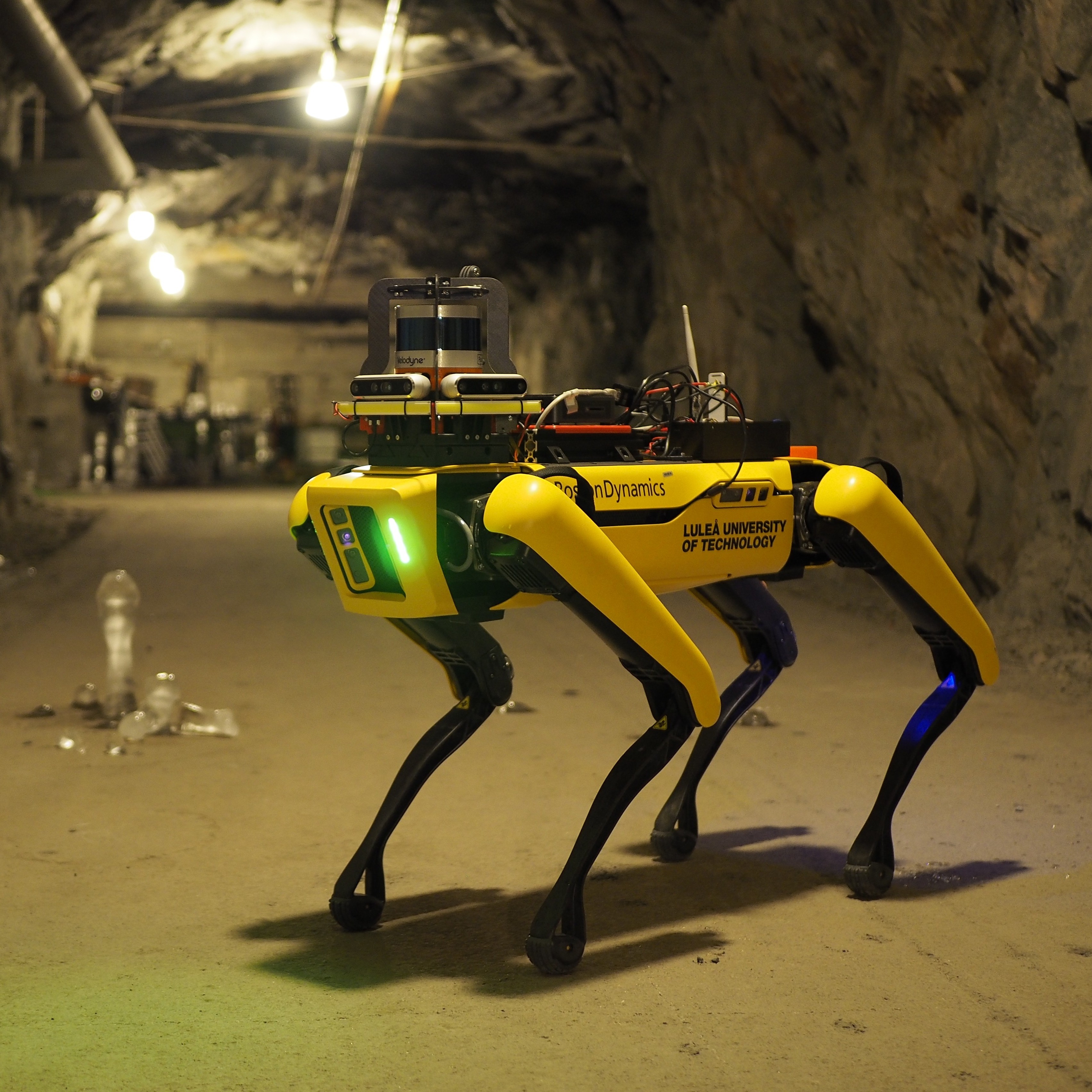} 
    \caption{Spot from BD} \label{subfig:spot}
    \end{subfigure} 
    \begin{subfigure}{0.24\textwidth}
    \includegraphics[width=\textwidth]{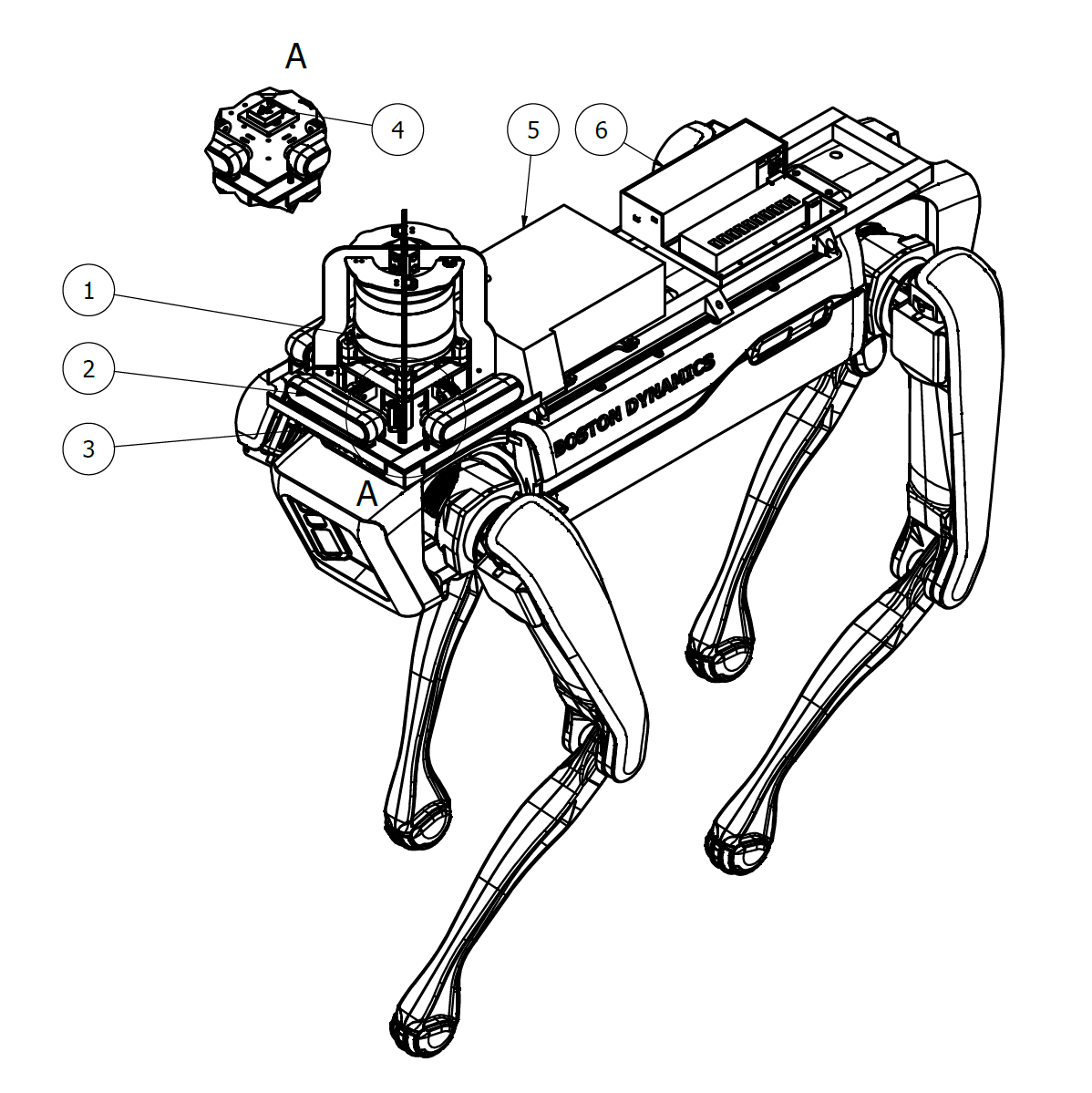}
    \caption{drawing of Spot} \label{subfig:spot_drawing}
    \end{subfigure}
    \begin{subfigure}{0.23\textwidth}
    \includegraphics[width=\textwidth]{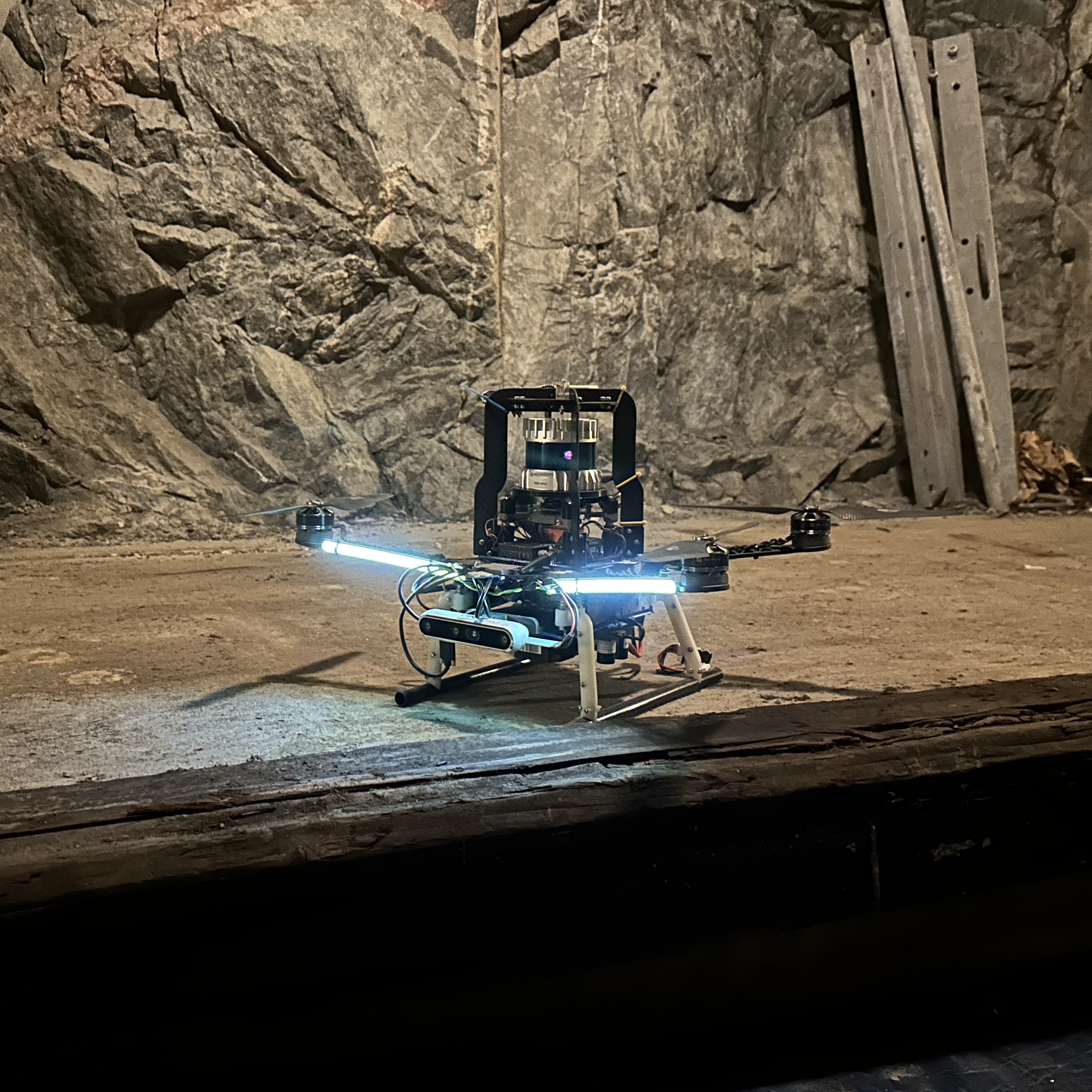}
    \caption{custom-build quadrotor} \label{subfig:shafter}
    \end{subfigure}
    \begin{subfigure}{0.28\textwidth}
    \includegraphics[width=\textwidth]{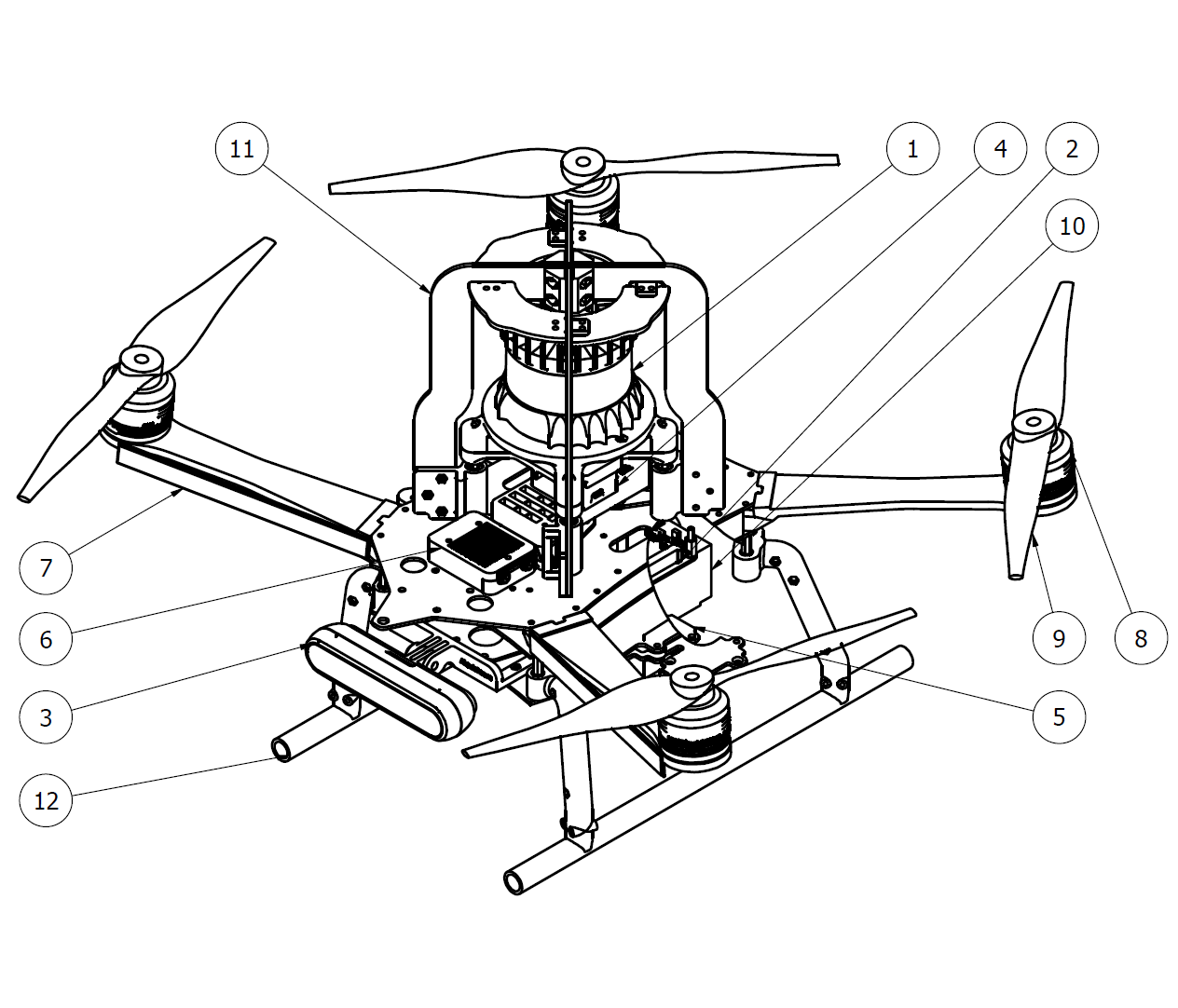}
    \caption{drawing of the quadrotor} \label{subfig:shafter_drawing}
    \end{subfigure} 
    \caption{(a) Spot from Boston Dynamics during the field experiments. (b) (1) Velodyne Puck Hi-Res 3D LiDAR, (2) Intel Realsense D455, (3) LED strips, (4) IMU Vectornav VN-100, (5) Spot Core and Intel NUC, (6) USB hub and external battery compartment. (c) The custom-built quadrotor during one of the field experiments. (d) (1) Ouster OS1-32 3D LiDAR, (2) Intel NUC, (3) Intel Realsense D455, (4) Pixhawk Cube Flight Controller, (5) Garmin single beam LiDAR, (6) Telemetry module, (7) LED strips, (8) T-motor MN3508 kV700, (9) 12.5in Propellers, (10) Battery, (11) Roll cage, (12) Landing gear.} \label{fig:robots}
    \vspace{-0.6cm}
\end{figure*}

For the experiments, two different types of robots were employed: a legged robot and an aerial robot. For the legged platform, the quadruped robot named Spot, developed by Boston Dynamics, was used, shown on Fig.~\ref{subfig:spot}. The robot was equipped with an autonomy package that consisted of a Velodyne Puck Hi-Res 3D LiDAR and an Intel NUC on-board computer that has a 10th Gen Intel® Core™ i5-10210U @ 1.60GHz CPU and 8GB of RAM as depicted on Fig.~\ref{subfig:spot_drawing}. In addition to the legged robot, the custom-built quadrotor was utilized as an aerial platform and it can be seen on Fig.~\ref{subfig:shafter}. This quadrotor carried an Ouster OS1-32, 3D LiDAR, and had the same on-board computer as the legged robot, as seen on Fig.~\ref{subfig:shafter_drawing}. One of the significant differences between the two LiDARs used in this experiment is the laser beam count and the vertical field of view. While the Velodyne Puck Hi-Res features 16 channels, the Ouster OS1-32 offers 32 channels, doubling the resolution of the produced depth images. As for the field of view, the first LiDAR has a vertical field of view of $20^o$, while the  Ouster LiDAR has a vertical field of view of $45^o$. These discrepancies have a direct impact on the quality and similarity of the depth images that each robot produced, as well as on the descriptors that were extracted in the early stages of the framework's pipeline. To integrate the algorithms, the ROS framework was utilized, on both Ubuntu 18.04 with Melodic version and Ubuntu 20.04 with Noetic version. By using the ROS framework, there is a higher ability to control and communicate with both robots, collect, process and evaluate the data from the sensors. Overall, the use of two different robotic platforms in these experiments allowed the investigation and comparison of the effectiveness of the proposed framework across different robot types and configurations.

\subsection{Metrics}

In order to evaluate the performance of FRAME against existing methods, we introduce several metrics, which are presented in Table~\ref{table:results}. These metrics include the number of points for each set of maps, the traveled distance for each trajectory, the approximate overlap percentage between the two maps, the transform error, and the computational time. Since the experiments were conducted in underground environments, GPS data were not available to provide a ground truth. Therefore, we utilize the tool CloudCompare~\cite{cloud_compare}, an open-source software that provides point cloud alignment and merging. By importing two point cloud maps, roughly aligning them by hand, and then letting the software refine them, we extract the final translation $T_{gt}$ and final rotation $R_{gt}$, which are used as the ground truth. To evaluate the performance of our framework against the existing methods, we define the translation and rotation error $T_e, R_e$ as:
\begin{equation}
    T_e = ||T_{gt} - T||, \:\: R_e = ||R_{gt} R^{-1} - I_3||,
\end{equation}
The final transform ${}^1\mathbf{T}_{2}$ used for the evaluation process consists of a translation part $T$ and a rotation matrix $R$. These errors were calculated by comparing the estimated translations and rotations produced by each method with the ground truth values obtained from CloudCompare. Overall, by introducing these metrics and using CloudCompare to provide ground truth data, we are able to conduct a comprehensive and accurate comparison of our framework against the other available approaches. This allows us to demonstrate the efficacy of the proposed method and make meaningful conclusions about its strengths and weaknesses.

\subsection{Descriptor Extraction Training} \label{subsec:training}

For the description extraction using the 3DEG framework, we leverage data from all presented environments. To enhance generalizability across environments, we initiate training with the mine depicted in Fig.~\ref{fig:tunnels}C, where we have the most data. Employing transfer learning, we progressively extend the training to encompass the remaining environments. For a more in-depth understanding of network parameters and loss functions, readers can refer to the relevant 3DEG article~\cite{stathoulopoulos20223deg}. Cross-sensor generalizability, a common challenge in the existing literature, is also addressed here. To optimize performance for the two distinct sensors, we train two separate models with adjusted input parameters. Since Spot was present in only two experiments, the model for the Velodyne LiDAR is trained on a smaller dataset due to limited availability. Both training and experiments were conducted on an 11th Gen Intel® Core™ i7-1165G7 @ 2.80GHz CPU.


\section{Field Evaluation Results} \label{sec:results}

The experimental field evaluation section presents a comprehensive evaluation of the proposed framework in a variety of subterranean environments and is divided into three subsections, each highlighting a specific aspect of the framework's performance. The first subsection discusses the evaluation of the framework in different subterranean environments and its ability to adapt to varying conditions, highlighting its robustness. The second subsection focuses on the framework's performance in fully autonomous missions, demonstrating the use-cases and real-life applications that could leverage such a framework. Last but not least, in the third subsection, a comparison with other available algorithms is performed, as we aim to highlight that the proposed approach outperforms existing methods in terms of accuracy, speed, and scalability. Together, these subsections provide a thorough analysis of the framework's capabilities and potential applications in real-world scenarios.


\subsection{Evaluation in a Variety of Subterranean Environments} \label{subsec:evaluation}

\subsubsection{\textbf{Underground Tunnel System}} \label{subsubsec:mjolk}

\begin{figure*}[!ht]
    \begin{subfigure}{0.23\textwidth}
    \includegraphics[width=\textwidth]{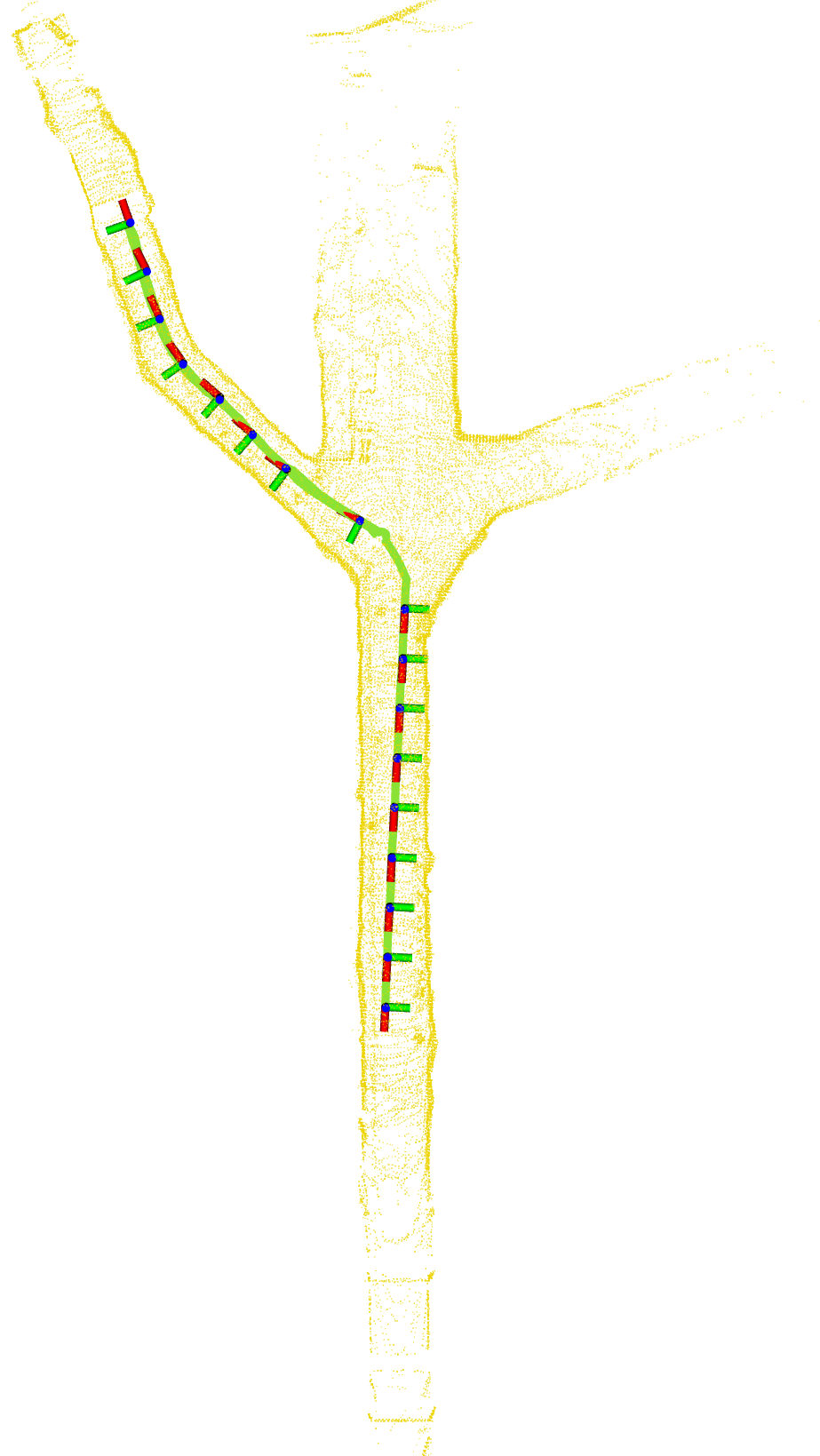}
    \caption{$\mathbf{M}_1$ and $\mathbf{P}_1$}
    \end{subfigure}
    \begin{subfigure}{0.23\textwidth}
    \includegraphics[width=\textwidth]{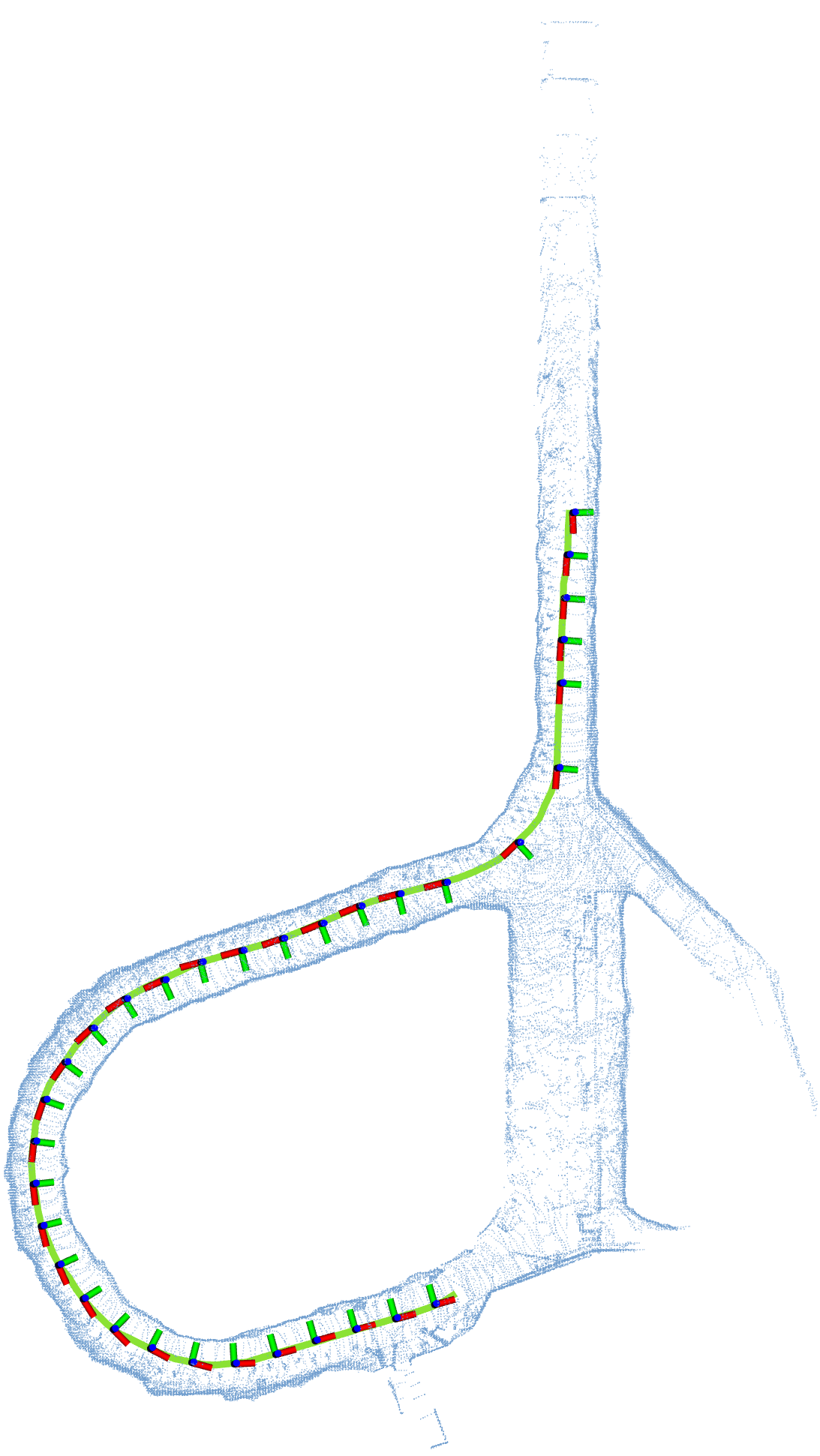}
    \caption{$\mathbf{M}_2$ and $\mathbf{P}_2$}
    \end{subfigure}
    \begin{subfigure}{0.26\textwidth}
    \includegraphics[width=\textwidth]{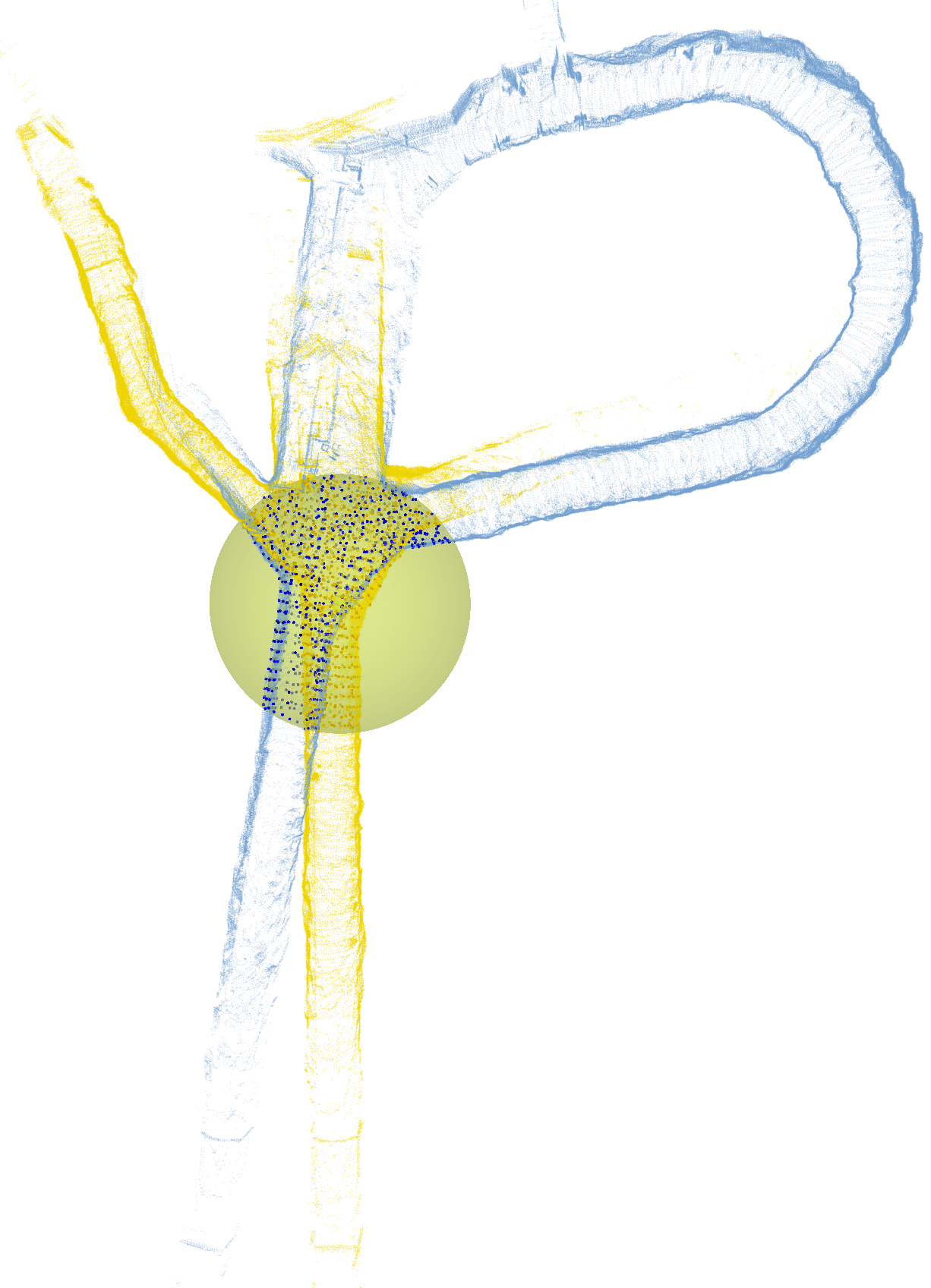} 
    \caption{Initial alignment}
    \end{subfigure}
    \begin{subfigure}{0.26\textwidth}
    \includegraphics[width=\textwidth]{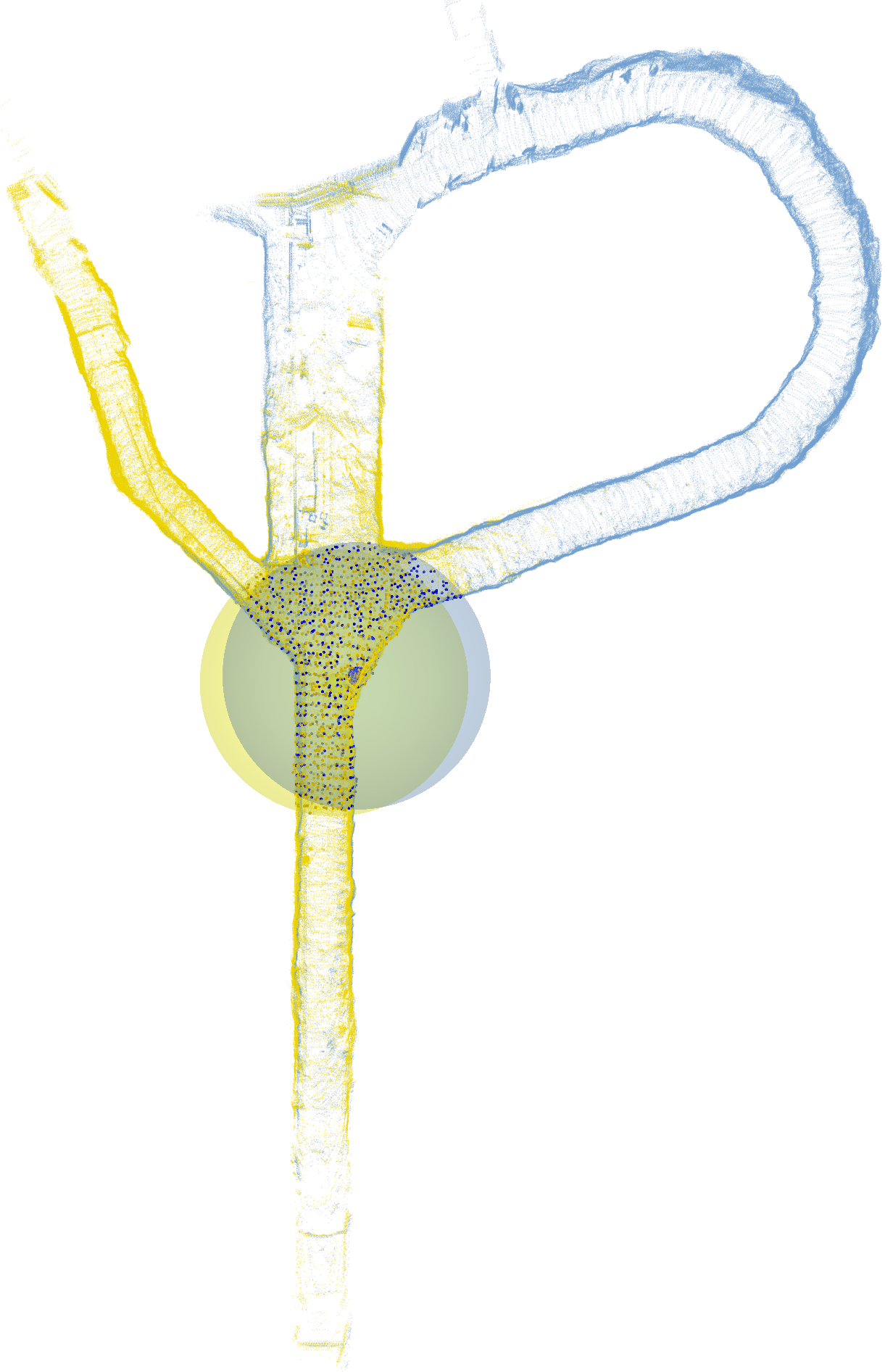}
    \caption{Refined merged map $\mathbf{M}$}
    \end{subfigure}
    \caption{The point cloud maps $\mathbf{M}_1$ and $\mathbf{M}_2$ were generated using different platforms and SLAM algorithms, namely LIO-SAM~\cite{LIO-SAM} on an aerial platform and DLO~\cite{DLO} on a legged platform, respectively. The initial yaw difference between the two maps was $180^o$. Subfigure (c) depicts the initial rough alignment based on the initial transform $\mathbf{T}_0$. The refined merged map (d), denoted as $\mathbf{M}$, is generated based on the final transform ${}^1\mathbf{T}_{2}$.}
    \label{fig:mjolkberget}
\end{figure*}

We start evaluating FRAME with the two maps visualized in Fig.~\ref{fig:mjolkberget}, each consisting of around $1.5 \cdot 10^5$ points, covering a total distance of 136 and 257 \unit{meters}, respectively. 
Map $\mathbf{M}_1$ was created using LIO-SAM\cite{LIO-SAM} on the quadrotor, while map $\mathbf{M}_2$ was created using DLO~\cite{DLO} on the legged robot, resulting in different point densities.
Even though as an initial evaluation experiment it poses many challenges as a multi-modal system as well as the initial yaw difference of approximately $180^o$, FRAME is able to leverage the $33\%$ overlap and detect the junction and consequently regress the angle discrepancy, in roughly $0.25$ seconds. 
The final transform ${}^1\mathbf{T}_{2}$ resulted in a translational error of $T_e = 0.09$ \unit{meters} and a rotational error of $R_e = 3.43$ \unit{degrees}. 
Further discussion on this dataset is presented in subsection~\ref{subsec:comparisons} as we utilize it to test against other methods.

\begin{figure*}[!t]
    \begin{subfigure}{0.54\textwidth}
    \includegraphics[width=\textwidth]{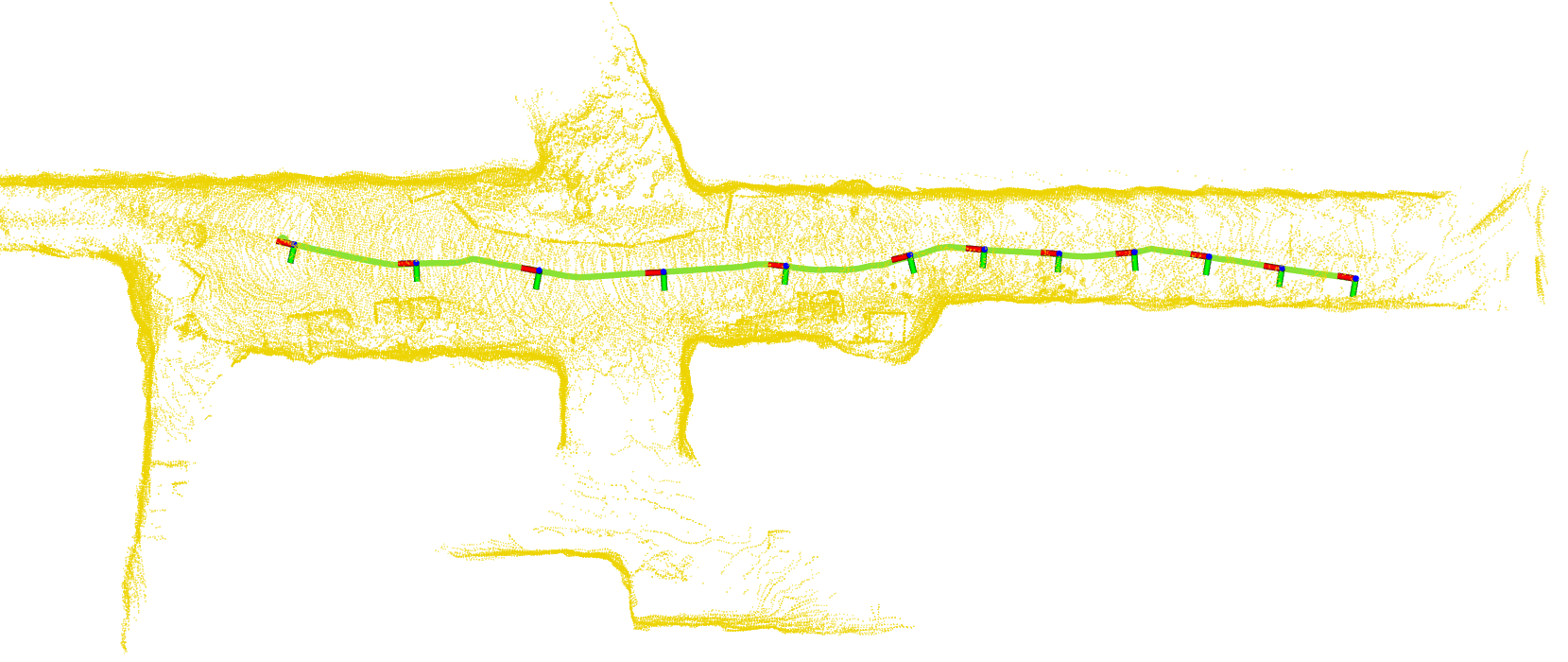}
    \caption{$\mathbf{M}_1$ and $\mathbf{P}_1$}
    \end{subfigure}
    \hfill
    \begin{subfigure}{0.45\textwidth}
    \includegraphics[width=\textwidth]{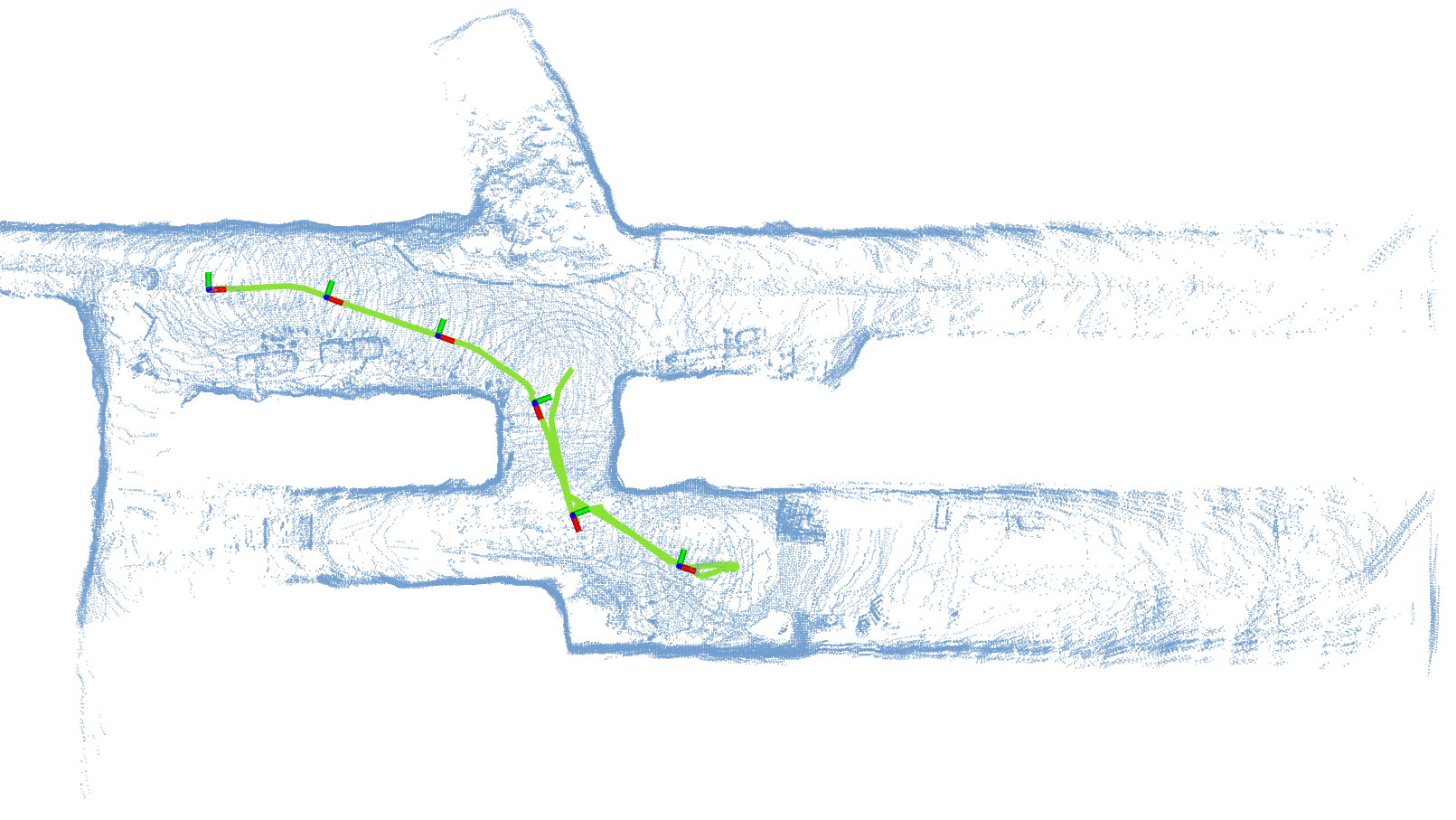}
    \caption{$\mathbf{M}_2$ and $\mathbf{P}_2$}
    \end{subfigure} \\
    \begin{subfigure}{0.4\textwidth}
    \includegraphics[width=\textwidth]{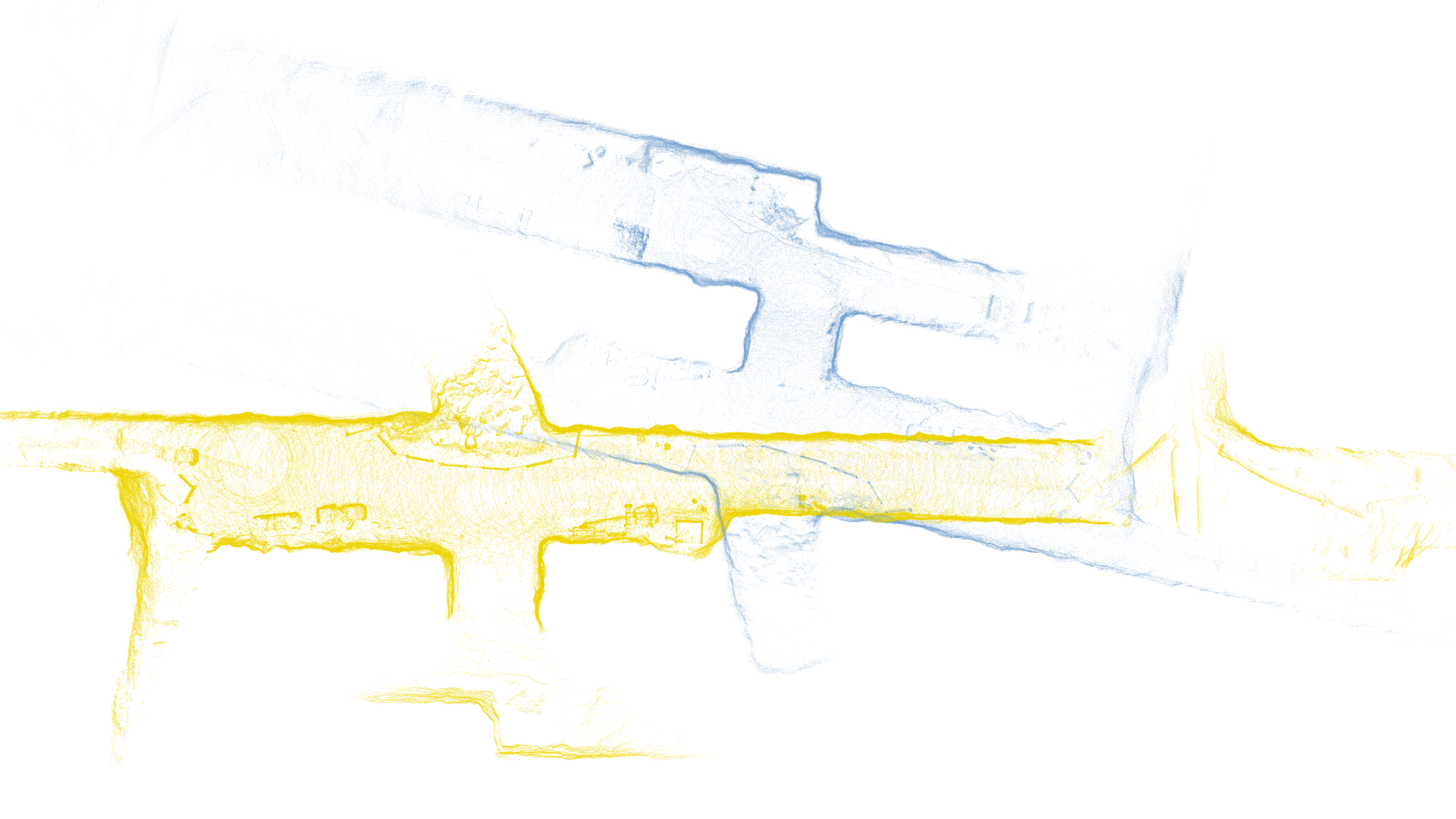} 
    \caption{Before alignment} \label{subfig:ncc_before}
    \end{subfigure}
    \hfill
    \begin{subfigure}{0.6\textwidth}
    \includegraphics[width=\textwidth]{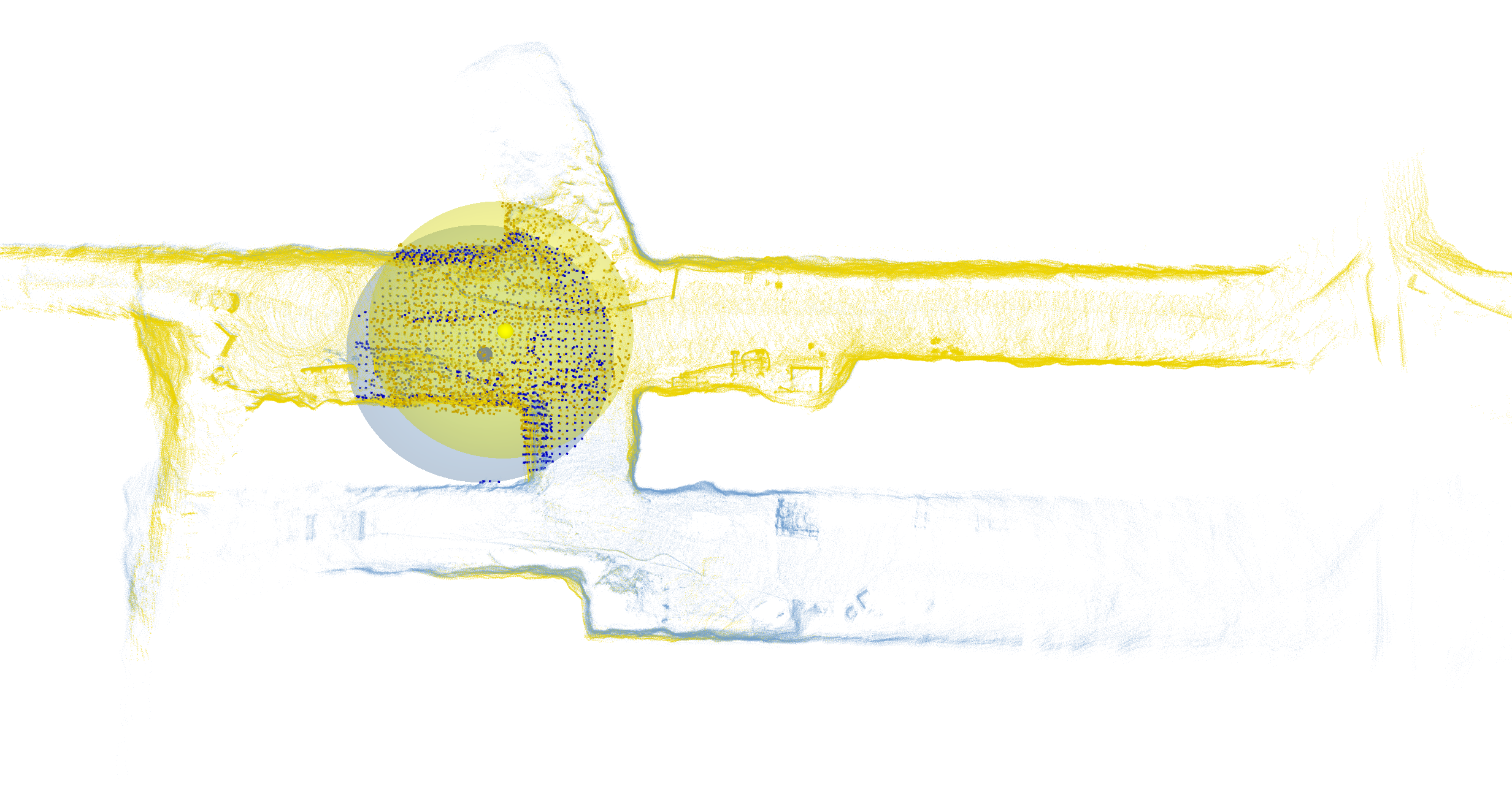}
    \caption{Refined merged map $\mathbf{M}$} \label{subfig:ncc_merged}
    \end{subfigure}
    \caption{The point cloud maps $\mathbf{M}_1$ and $\mathbf{M}_2$ were generated by the quadruped legged robot Spot, from Boston Dynamics and the SLAM algorithm, LIO-SAM~\cite{LIO-SAM}. The initial yaw difference between the two maps was approximately $45^o$. Subfigure (c) depicts the maps before any alignment while the refined merged map, denoted as $\mathbf{M}$, is generated based on the final transform ${}^1\mathbf{T}_{2}$, defined as $\mathbf{M} = {}^1\mathbf{M}_1 \cup {}^1\mathbf{T}_{2} {}^2\mathbf{M}_2$ and depicted in subfigure (d).}
    \label{fig:ncc}
\end{figure*}
\subsubsection{\textbf{Subterranean Construction Site}} \label{subsubsec:NCC}

The second experiment to take place is presented in Fig.~\ref{fig:ncc}, in an underground construction site.
The map $\mathbf{M}_1$ and $\mathbf{M}_2$ are from the top and bottom branch of a metro line, with a common overlapping area being the junction connecting them.
Similar to the first experiment, the concept remains that two different branches are maneuvered through independently, resulting in a common global map where the overlapping junction connects them.
The total traveled distance of the robotic platform is approximately $60$ and $70$ \unit{meters}, respectively, resulting in two maps of $5 \cdot 10^5$ and $1 \cdot 10^6$ points. 
As seen in subfigure~\ref{subfig:ncc_before}, the second map $\mathbf{M}_2$ is initially rotated by $45^o$ with respect to $\mathbf{M}_1$ and has an overlap percentage of $37\%$, providing enough information for the merging process. 
Through the overlap estimation process, FRAME is able to find the overlapping regions and extract the corresponding spheres $\mathbf{S}_1$ and $\mathbf{S}_2$, with a radius of $r=10$ \unit{meters}, yielding an initial transform $T_0$. 
Looking closely in the resulting merged map $M$ and the overlapping spheres $\mathbf{S}_1$ and $\mathbf{S}_2$ in subfigure~\ref{subfig:ncc_merged}, we notice that even though there is a relatively high overlap percentage between the maps $\mathbf{M}_1$ and $\mathbf{M}_2$, it is highly reduced between the two spheres, as the detected trajectory points $\mathbf{p}_{1,k_i}$ and $\mathbf{p}_{2,k_j}$ are far apart. 
Despite this fact, the proposed framework is able to regress the difference and successfully merge the two maps with a computational time of just $0.1$ seconds, providing useful insight about the limitations and the important aspects of the main pipeline, that could lead to a failed registration.

\subsubsection{\textbf{Multi-branch Junction in Subterranean Mine}} \label{subsubsec:lkab_multi}

\begin{figure*}[!ht]
    \begin{subfigure}{0.49\textwidth}
    \includegraphics[width=\textwidth]{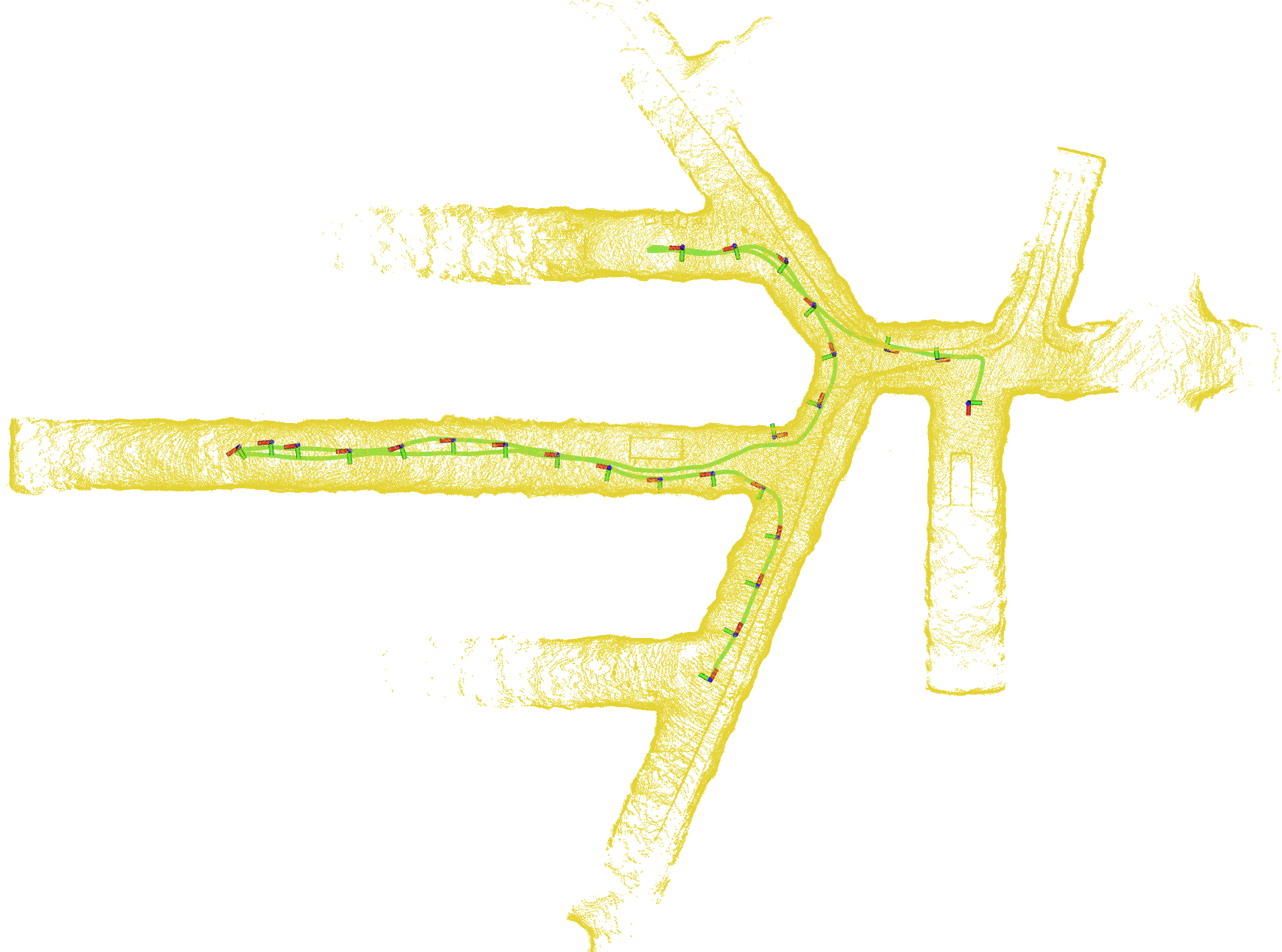}
    \caption{$\mathbf{M}_1$ and $\mathbf{P}_1$}
    \end{subfigure}
    \hfill
    \begin{subfigure}{0.49\textwidth}
    \includegraphics[width=\textwidth]{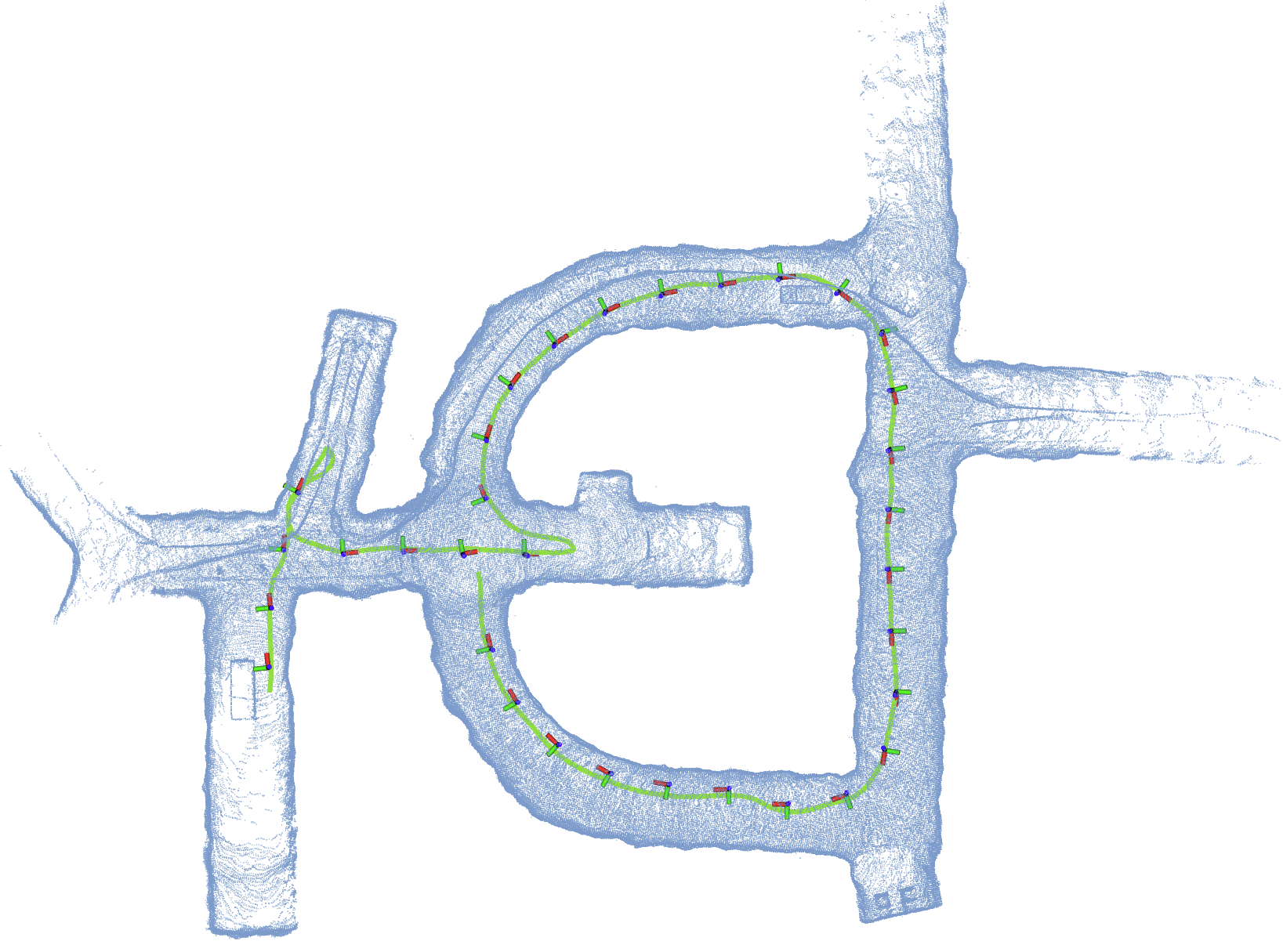}
    \caption{$\mathbf{M}_2$ and $\mathbf{P}_2$}
    \end{subfigure} \\
    \begin{subfigure}{1.0\textwidth}
    \includegraphics[width=\textwidth]{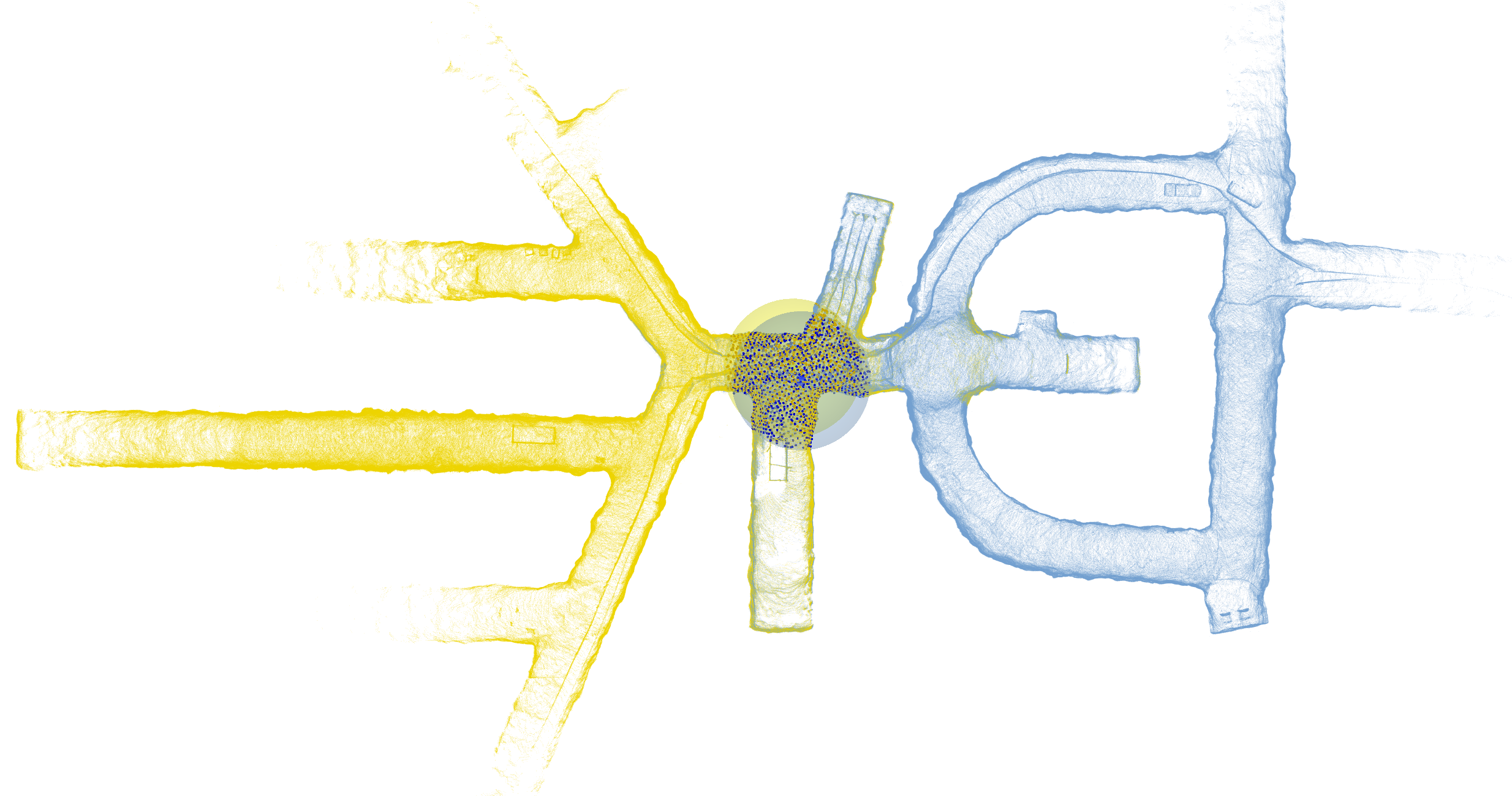}
    \caption{Final merged map $\mathbf{M}$} \label{subfig:lkab_junction_merged}
    \end{subfigure}
    \caption{The point cloud maps, $\mathbf{M}_1$ and $\mathbf{M}_2$, were obtained using the custom-built quadrotor that was fitted with the Ouster OS1-32 3D LiDAR sensor. The maps were created by exploring multiple branches of the area separately.}
    \label{fig:lkab_junction}
    \vspace{-0.6cm}
\end{figure*}

FRAME was also evaluated from datasets in a larger scale subterranean real mining environment with an overlapping junction, shown in Fig.~\ref{fig:lkab_junction}.
Here two maps, $\mathbf{M}_1$ and $\mathbf{M}_2$, were generated by a robotic platform that traversed through two different direction in a multi-branch junction. 
Once again, the robotic platform was equipped with the Ouster OS1-32 3D LiDAR, capturing $2.8 \cdot 10^6$ points per map and covering a total distance of $145$ \unit{meters} per map.
The initial rotation difference between the two maps was approximately $45^o$ and the overlap percentage was roughly $31\%$, making it sufficient for the overlap estimation module to handle this scenario as well.
As the corresponding spheres $\mathbf{S}_1$ and $\mathbf{S}_2$ are close together and the $r=10$ \unit{meters} radius include important features from the junction area, the registration and final alignment of the two maps is successful, providing a low translation error $T_e=0.082$ \unit{meters} in just above $0.2$ seconds. With this experiment, we demonstrate that despite the complexity of the environment, FRAME still manages to deliver a correct global transform. 

\subsubsection{\textbf{Large-scale Operational Mining Area}} \label{subsubsec:kps}

\begin{figure*}[!ht]
    \centering
    \begin{subfigure}{0.36\textwidth}
    \includegraphics[width=\textwidth]{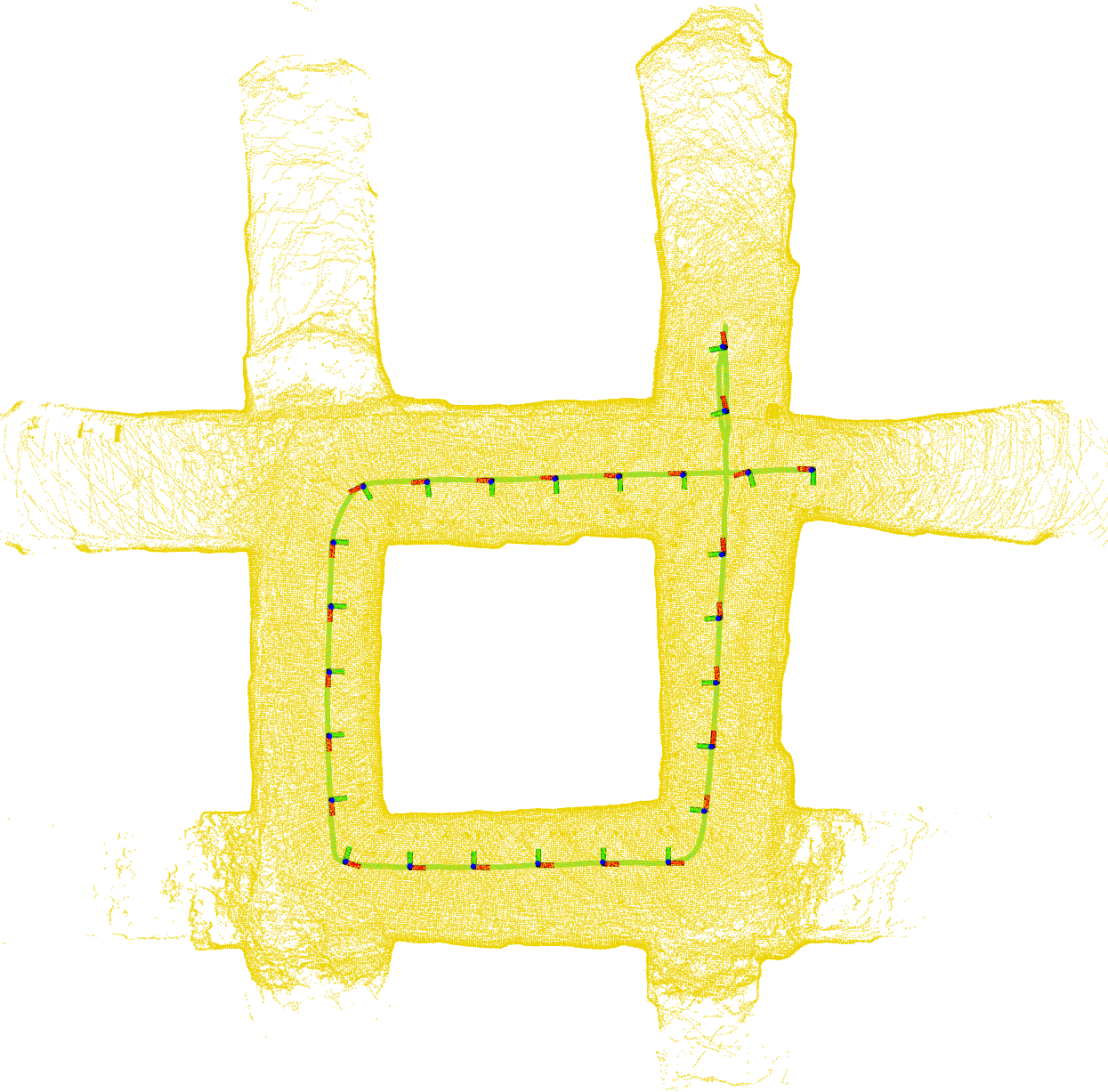}
    \caption{$\mathbf{M}_1$ and $\mathbf{P}_1$}
    \end{subfigure}
    \hfill
    \begin{subfigure}{0.62\textwidth}
    \includegraphics[width=\textwidth]{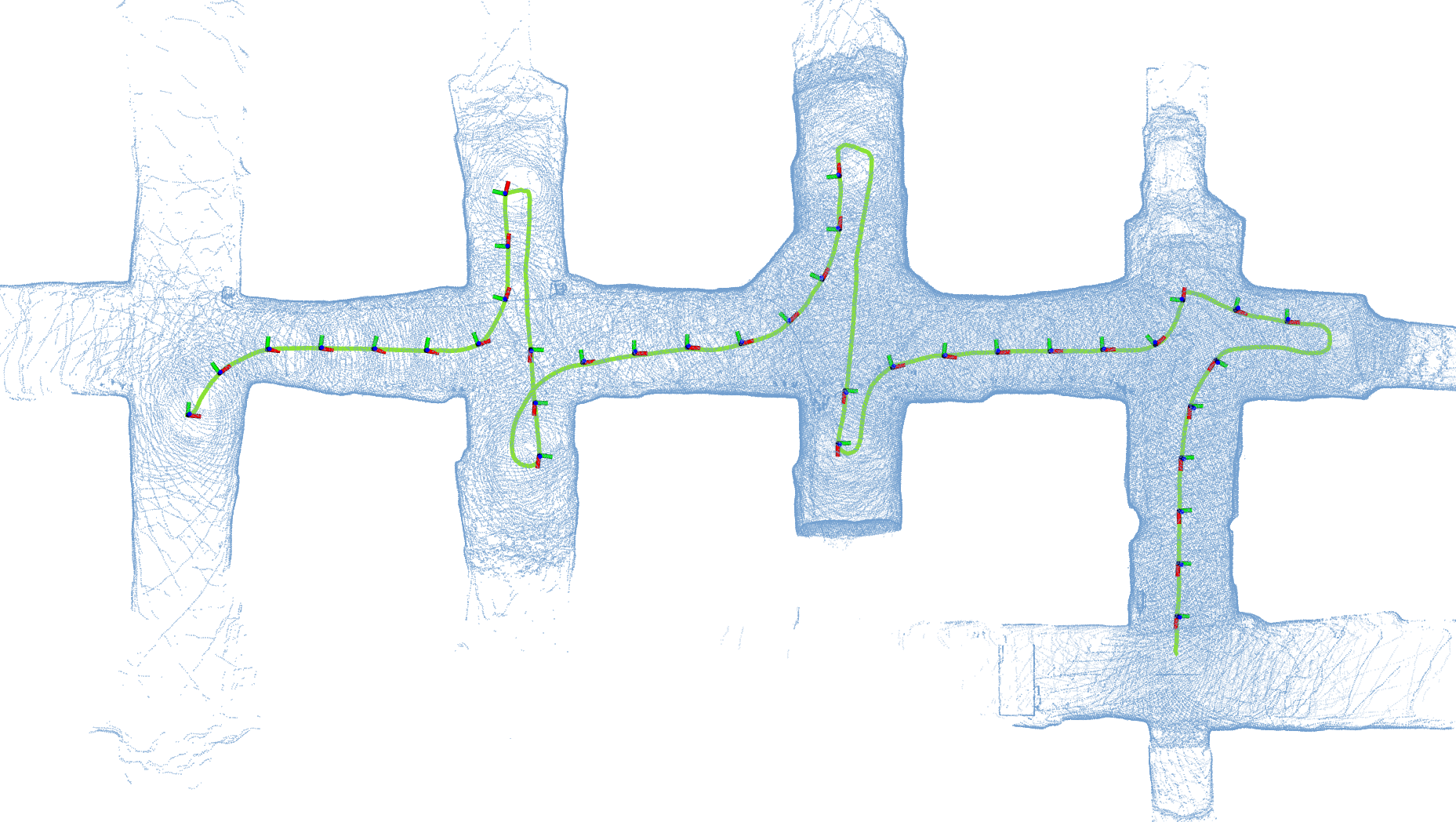}
    \caption{$\mathbf{M}_2$ and $\mathbf{P}_2$}
    \end{subfigure}
    \begin{subfigure}{1.0\textwidth} 
    \includegraphics[width=\textwidth]{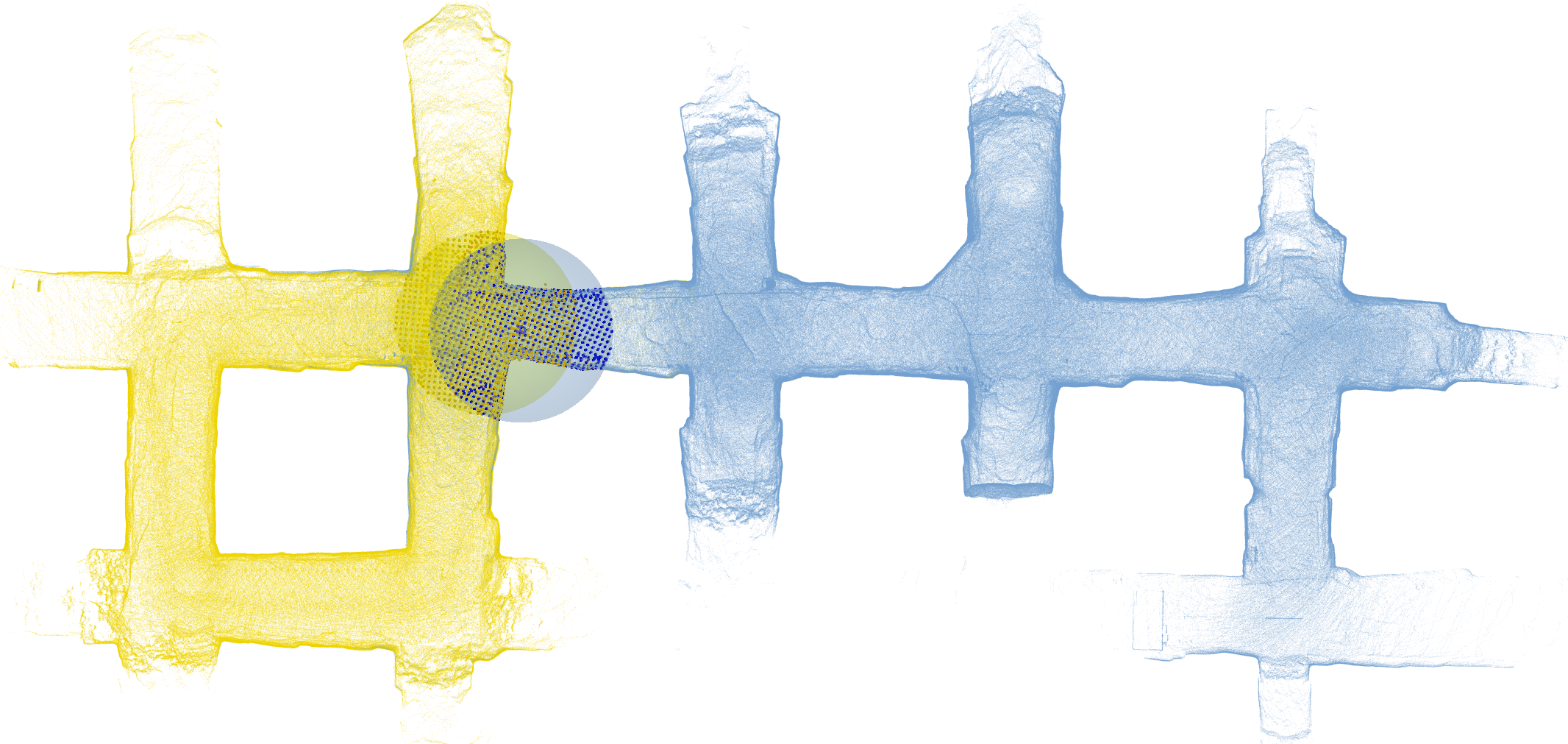}
    \caption{Final merged map $\mathbf{M}$} \label{subfig:kps_merged}
    \end{subfigure}
    \caption{The custom-built quadrotor equipped with the Ouster OS1-32 3D LiDAR was used to generate the point cloud maps $\mathbf{M}_1$ and $\mathbf{M}_2$. The two maps were created by exploring in a loop for one map and inspecting the drifts for the other.} \label{fig:kps}
    \vspace{-0.6cm}
\end{figure*}

This dataset covers a larger area of an underground mining facility, as seen on Fig.~\ref{fig:kps}, with a traversed distance of $115$ and $157$ \unit{meters} respectively for each trajectory $\mathbf{P}_1$ and $\mathbf{P}_2$.
The merging process involves the two maps $\mathbf{M}_1$ and $\mathbf{M}_2$ consisting of $2 \cdot 10^6$ and $4.5 \cdot 10^6$ points each and an estimated overlap of roughly $22\%$.
Similarly to previous encounters, the center points $\mathbf{p}_{1,k_i}$ and $\mathbf{p}_{2,k_j}$ are not very close to each other, but due to the automated sphere radius selection, the radius is chosen at $r=15$ \unit{meters}, including more feature points and accommodating for the wider tunnel environment.
All in all, the computed final transform ${}^1\mathbf{T}_{2}$ is able to align the point cloud maps with an accuracy of just under $T_e=0.1$ \unit{meters} and $R_e=1$ degree with a computational time of $0.35$ seconds, demonstrating the scalability of the proposed approach. 


\subsection{FRAME in Fully Autonomous Missions} \label{subsec:autonomous}

\subsubsection{\textbf{Deployment Framework}}

In this section, we highlight the combination of FRAME with a reactive exploration framework for subterranean tunnel environments, denoted as the COMPRA~\cite{compra} mission. COMPRA was designed to efficiently and quickly navigate through tunnel environments using a reactive 3D-LiDAR based Artificial Potential Field combined with a depth-camera based heading regulation technique to orient the UAV towards open areas, but does not consider complete exploration coverage. As such we can appropriately demonstrate the FRAME use-case of map merging multi-session point cloud maps, e.g. scenarios where the environment is too large, or too complex, for the robot exploration algorithm to handle it through one mission. Instead, we can deploy the robot multiple times towards different areas and merge the resulting maps into one global map with greater area coverage. We showcase this through a series of fully autonomous exploration missions in real field mining environments, where FRAME was used to merge the resulting point cloud mine maps from exploration runs with overlapping areas.

\subsubsection{\textbf{Opposite Parallel Exploration in a Subterranean Mining Tunnel Environment}} \label{subsubsec:lkab_drift}

\begin{figure*}[!t]
    \begin{subfigure}{0.49\textwidth}
    \includegraphics[width=\textwidth]{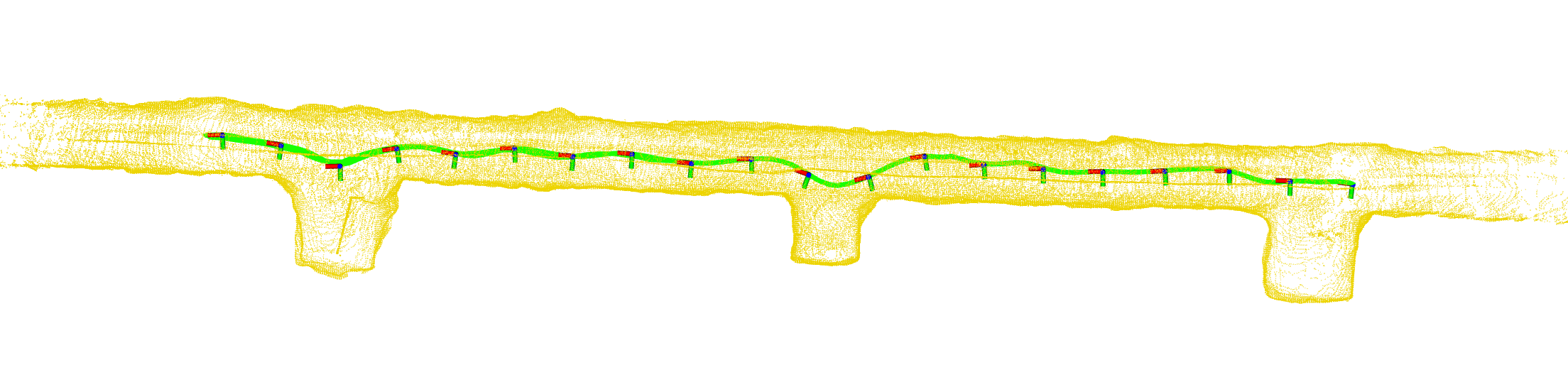}
    \caption{$\mathbf{M}_1$ and $\mathbf{P}_1$}
    \end{subfigure}
    \hfill
    \begin{subfigure}{0.49\textwidth}
    \includegraphics[width=\textwidth]{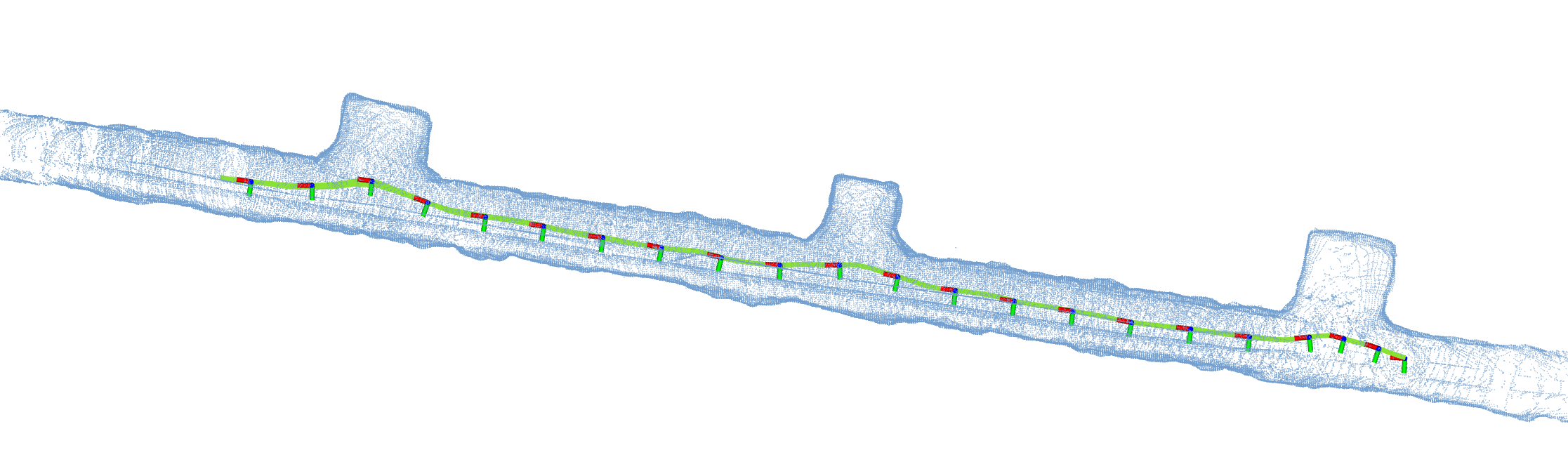}
    \caption{$\mathbf{M}_2$ and $\mathbf{P}_2$}
    \end{subfigure}
    \\
    \begin{subfigure}{1.\textwidth}
    \includegraphics[width=\textwidth]{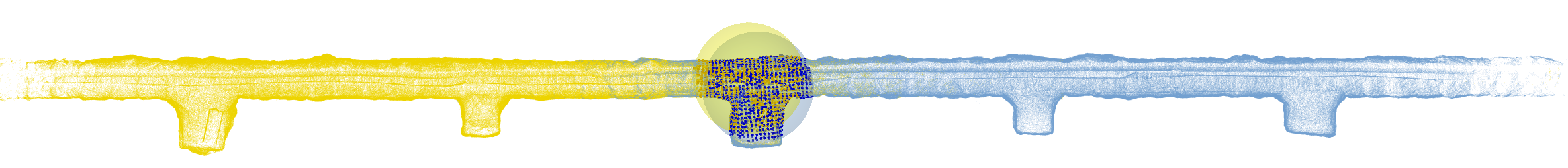}
    \caption{Final merged map $\mathbf{M}$} \label{subfig:lkab_left_right_merged}
    \end{subfigure}
    \caption{The point cloud maps $\mathbf{M}_1$ and $\mathbf{M}_2$ were generated by a custom-built quadrotor equipped with the Ouster OS1-32 3D LiDAR. The initial yaw difference between the two maps was approximately $180^o$, as the two directions of the tunnel were explored independently.}
    \label{fig:lkab_left_right}
    \vspace{-0.6cm}
\end{figure*}
\begin{figure*}[!b]
    \centering
    \begin{subfigure}{0.65\textwidth}
    \includegraphics[width=\textwidth]{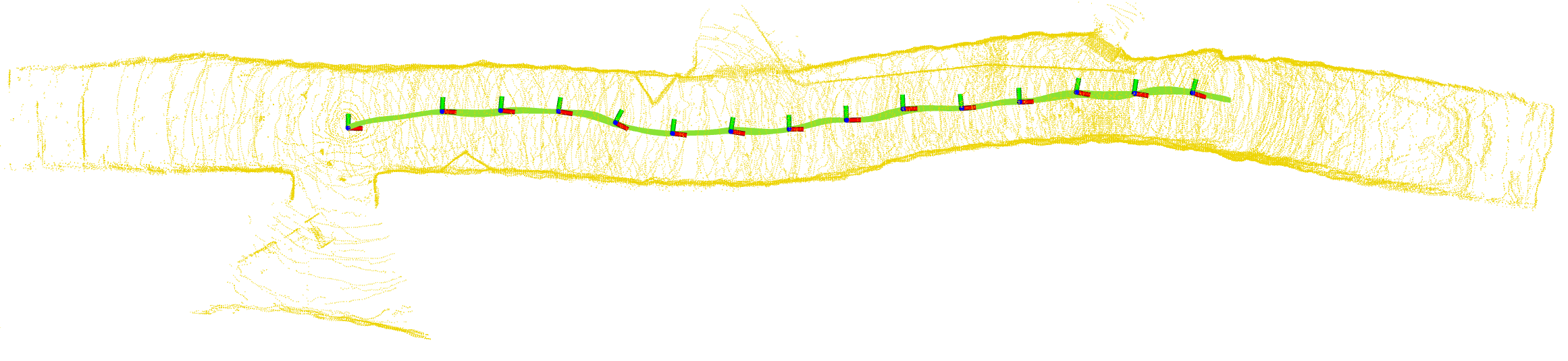}
    \caption{$\mathbf{M}_1$ and $\mathbf{P}_1$}
    \end{subfigure}
    \hfill
    \begin{subfigure}{0.34\textwidth}
    \includegraphics[width=\textwidth]{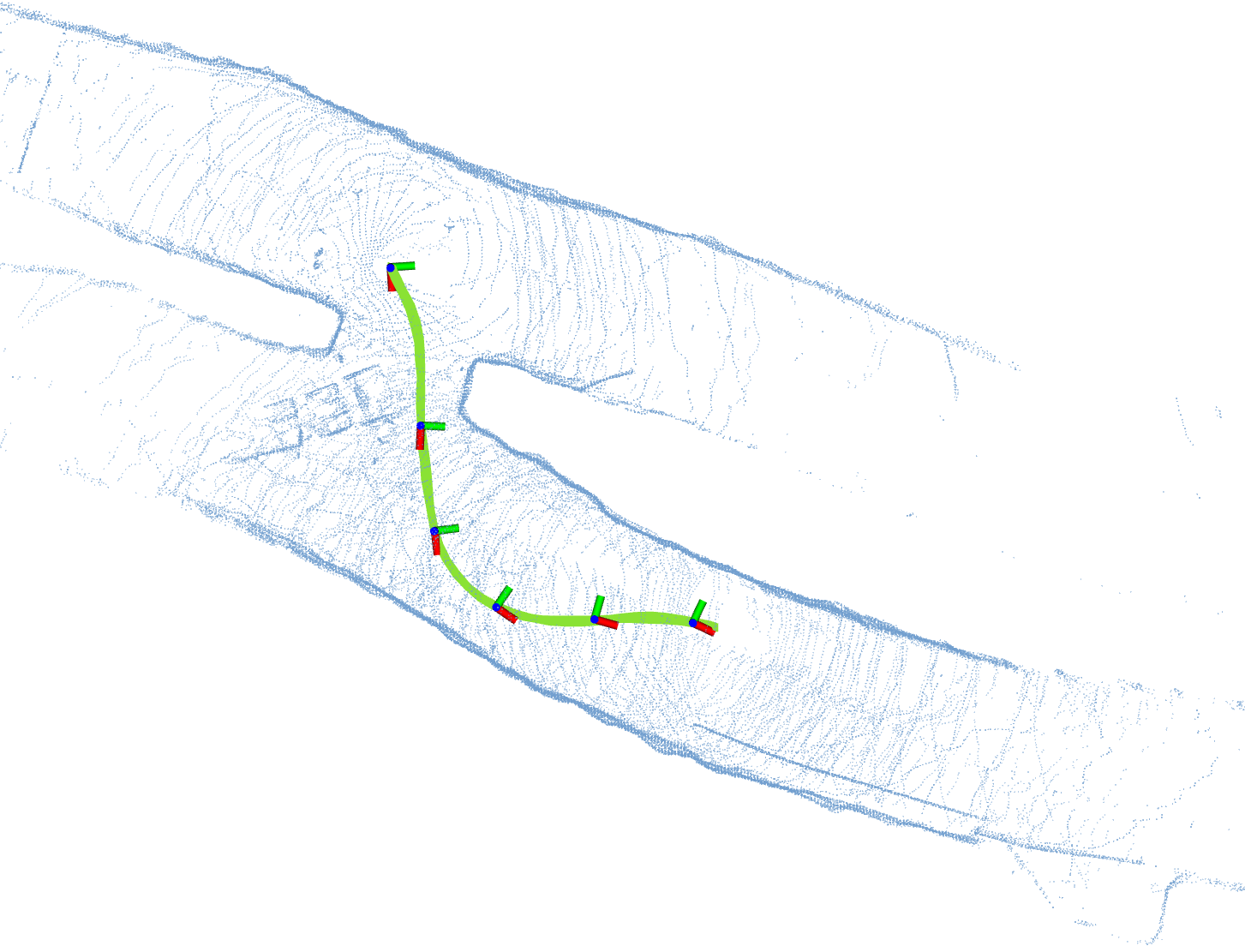}
    \caption{$\mathbf{M}_2$ and $\mathbf{P}_2$}
    \end{subfigure}
    \begin{subfigure}{1.0\textwidth} 
    \centering
    \includegraphics[width=\textwidth]{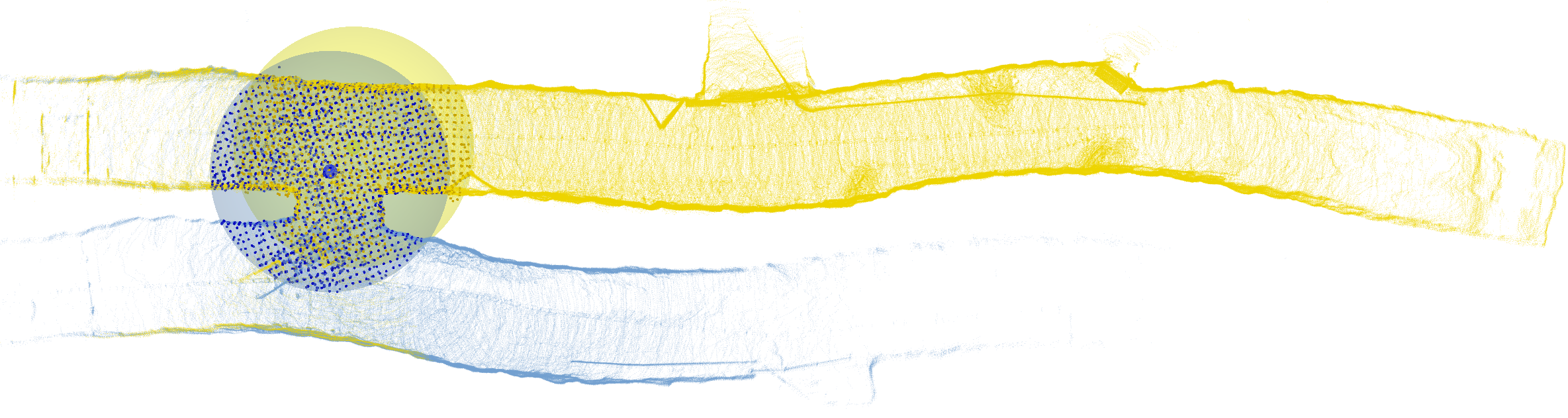}
    \caption{Final merged map $\mathbf{M}$} \label{subfig:epiroc_machine_merged}
    \end{subfigure}
    \caption{The point cloud maps $\mathbf{M}_1$ and $\mathbf{M}_2$ were generated by the custom-built quadrotor and the SLAM algorithm, LIO-SAM. The starting position on both maps is the same, as they take off and explore two parallel tunnels.}
    \label{fig:epiroc_machine}
\end{figure*}
In this scenario the COMPRA mission was deployed twice, shown in Fig.~\ref{fig:lkab_left_right}, where two maps, $\mathbf{M}_1$ and $\mathbf{M}_2$ were generated by the UAV platform exploring through a common underground mining tunnel in opposite directions with an overlapping drift. 
The distance traveled by the platform for $\mathbf{M}_1$ and $\mathbf{M}_2$ was approximately $140$ \unit{meters} for each map, resulting in maps with $2.4 \cdot 10^5$ points respectively.
The initial rotation between the two maps was $180^o$, and the overlap percentage was $18\%$, making it a difficult scenario, along with the self-similar, featureless tunnel walls.
FRAME is yet again able to extract the overlapping regions and corresponding spheres, $\mathbf{S}_1$ and $\mathbf{S}_2$, with a radius of $r=10$ \unit{meters}, yielding an initial transform $T_0$. 
Contrary to the second experiment, the maps $\mathbf{M}_1$ and $\mathbf{M}_2$ share a small overlapping percentage, but since the sampled spheres $\mathbf{S}_1$ and $\mathbf{S}_2$ have their center points $\mathbf{p}_{1,k_i}$ and $\mathbf{p}_{2,k_j}$ very close to each other, the registration and therefore the alignment of the maps is executed seamlessly. As presented in the Table~\ref{table:results} the translation error is kept very low at $T_e=0.1$ \unit{meters} while the rotation error is under $R_e=0.8$ \unit{degrees} and the computational time is at $0.22$ seconds.

\subsubsection{\textbf{Exploring Two Branches of a Test Mine Environment Junction}} \label{subsubsec:epiroc}

The figure labeled as Fig.~\ref{fig:epiroc_machine} shows two shorter exploration runs in two different branches of a subterranean junction area and the resulting point cloud maps.  
In this experiment, the two parallel drifts of the mining tunnel are explored independently and are connected through a common junction highlighted in subfigure~\ref{subfig:epiroc_machine_merged}, with the aim of creating a common global map.
The robotic platform traveled approximately $106$ and $61$ \unit{meters} to cover $\mathbf{M}_1$ and $\mathbf{M}_2$, respectively, resulting in two maps with $1.4 \cdot 10^6$ and $6 \cdot 10^5$ points.
Initially, $\mathbf{M}_2$ was rotated by $30^o$ with respect to $\mathbf{M}_1$, and the overlap percentage was estimated to be $33\%$. 
Using FRAME, overlapping regions were detected, and corresponding spheres $\mathbf{S}_1$ and $\mathbf{S}_2$ with a radius of $r=15$ \unit{meters} were extracted to obtain an initial transform $T_0$.
During this experiment, it was crucial to increase the sphere radius $r$ in order to accommodate for the larger tunnel width.
By increasing the radius we ensure that the spheres $\mathbf{S}_1$ and $\mathbf{S}_2$ will include enough points and more specifically feature points, e.g. edges, to make the registration possible.
As presented in Table~\ref{table:results} the computational time is kept low at $0.1$ seconds.

\subsubsection{\textbf{Multiple Short Missions in a Large-scale Multi-branch Mining Area}} \label{subsubsec:epiroc_multi}
\begin{figure*}[!b] 
    \begin{subfigure}{0.305\textwidth}
    \includegraphics[width=\textwidth]{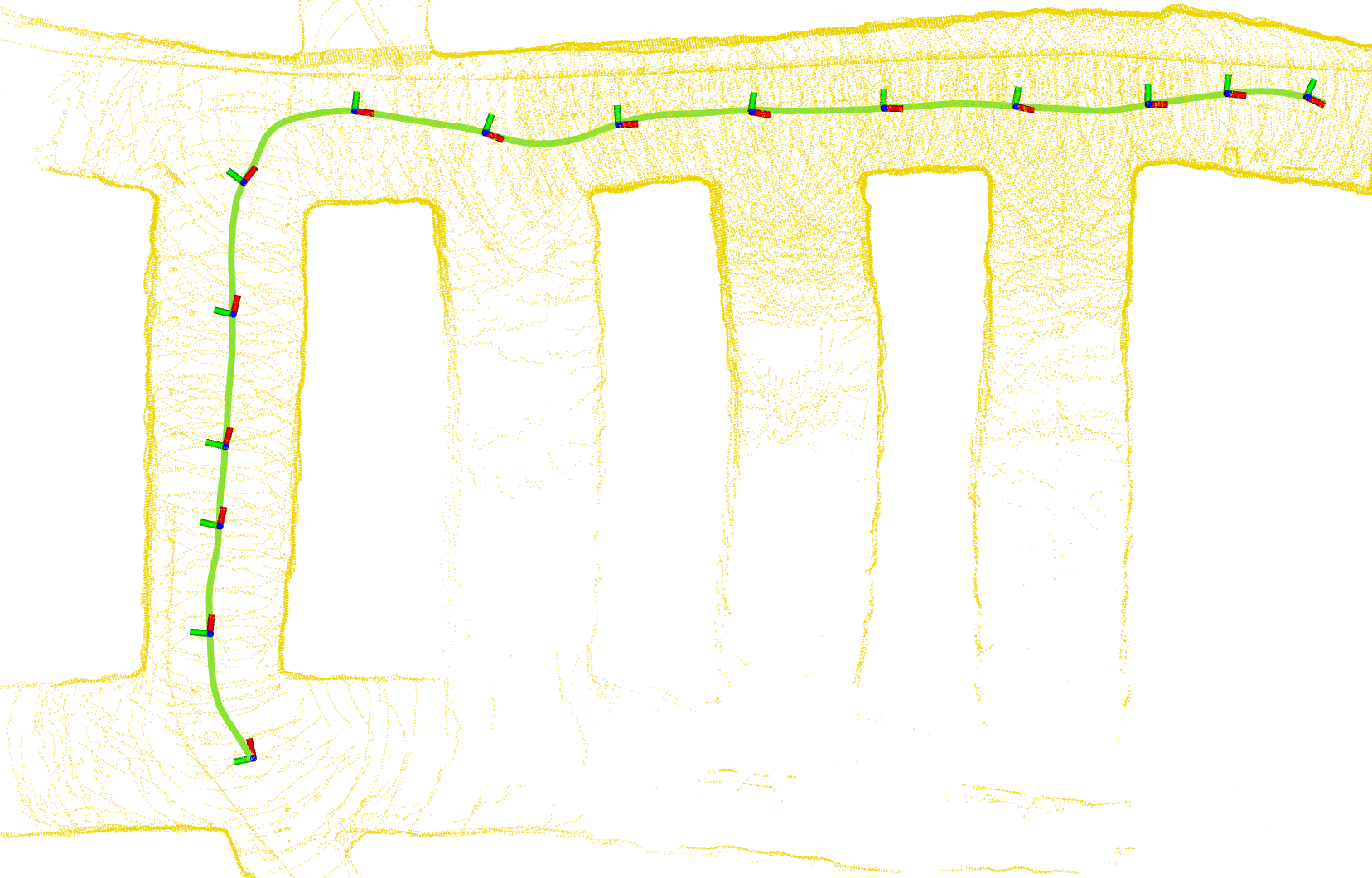} \caption{$\mathbf{M}_1$ and $\mathbf{P}_1$}
    \end{subfigure}
    \begin{subfigure}{0.30\textwidth}
    \includegraphics[width=\textwidth]{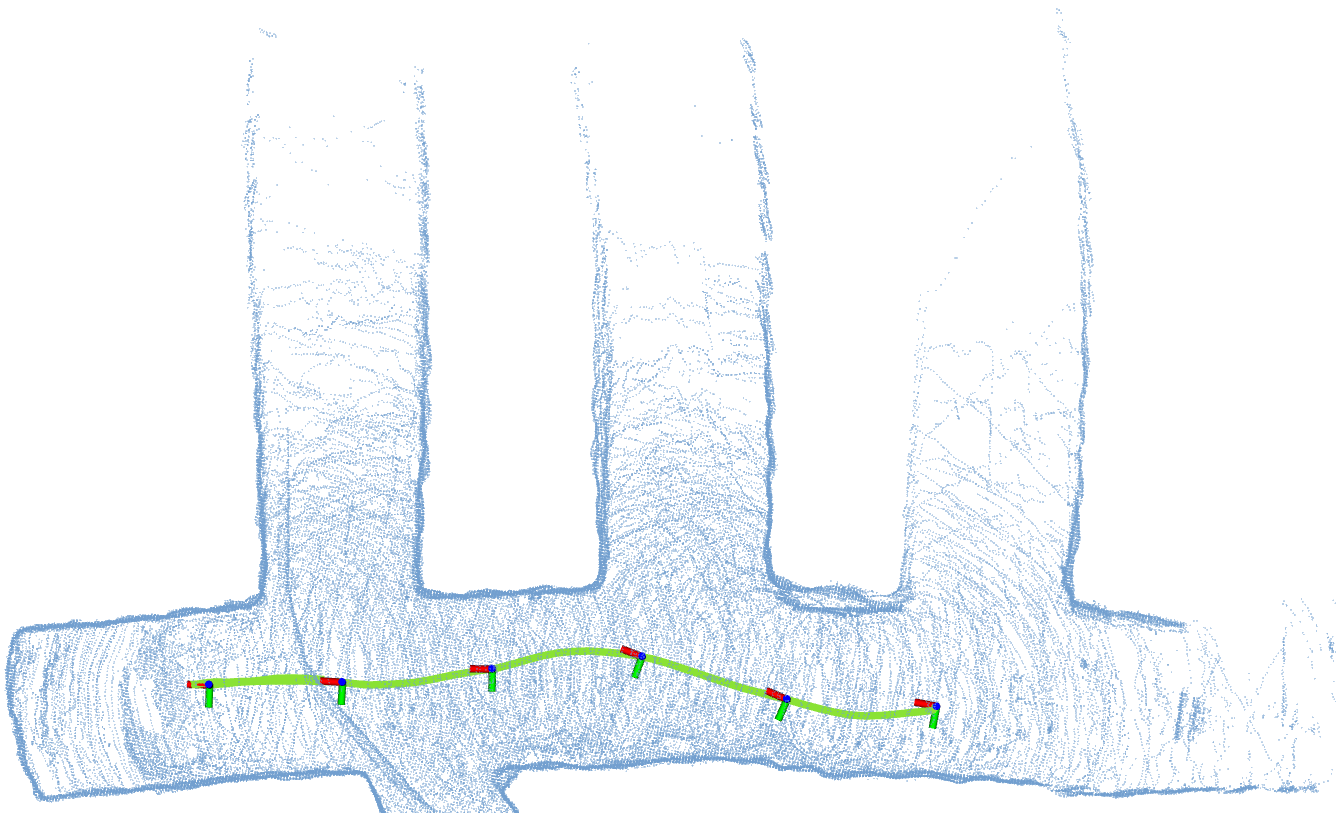} \caption{$\mathbf{M}_2$ and $\mathbf{P}_2$}
    \end{subfigure}
    \begin{subfigure}{0.20\textwidth}
    \includegraphics[width=\textwidth]{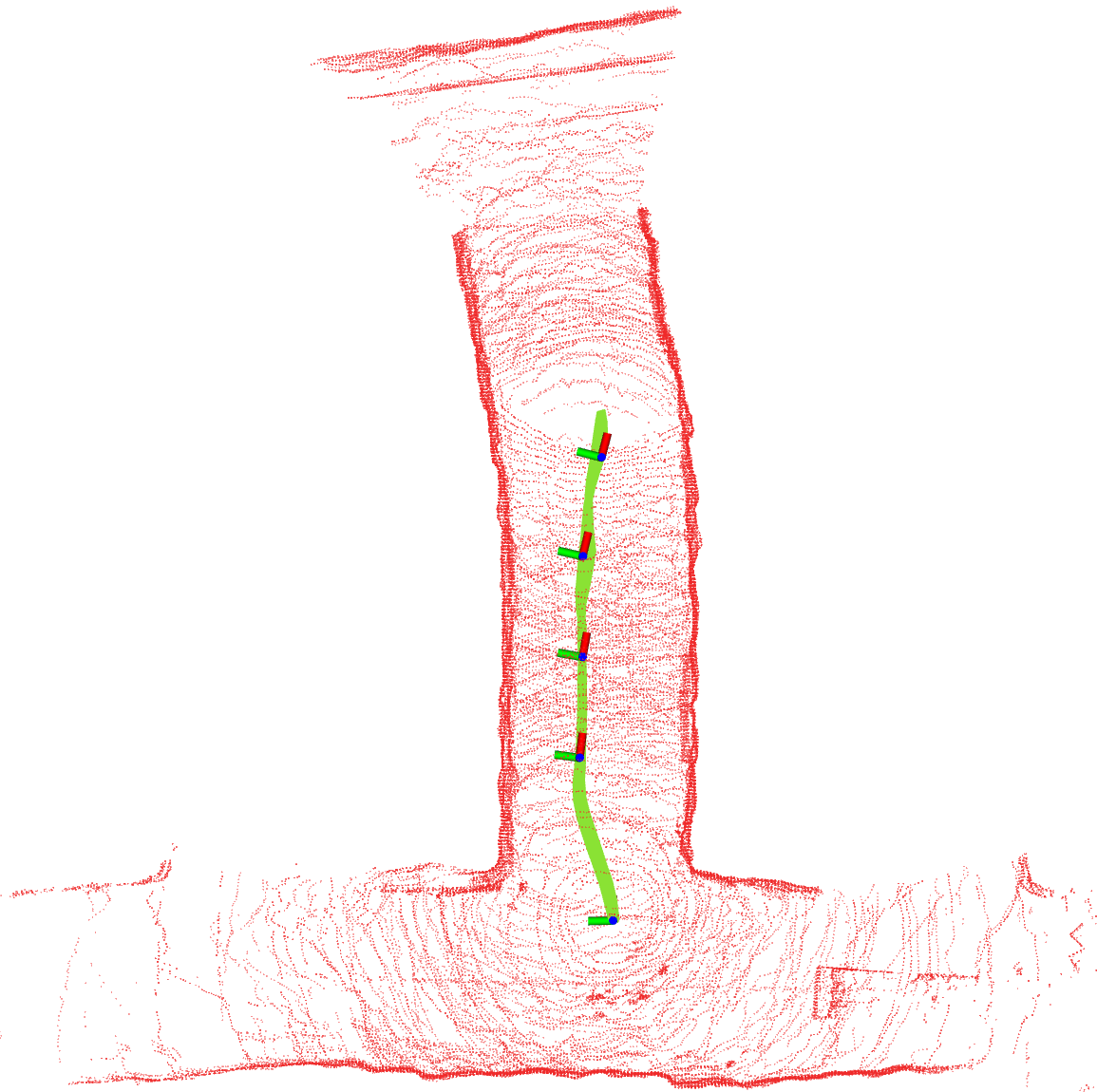} \caption{$\mathbf{M}_3$ and $\mathbf{P}_3$}
    \end{subfigure} 
    \begin{subfigure}{0.18\textwidth}
    \includegraphics[width=\textwidth]{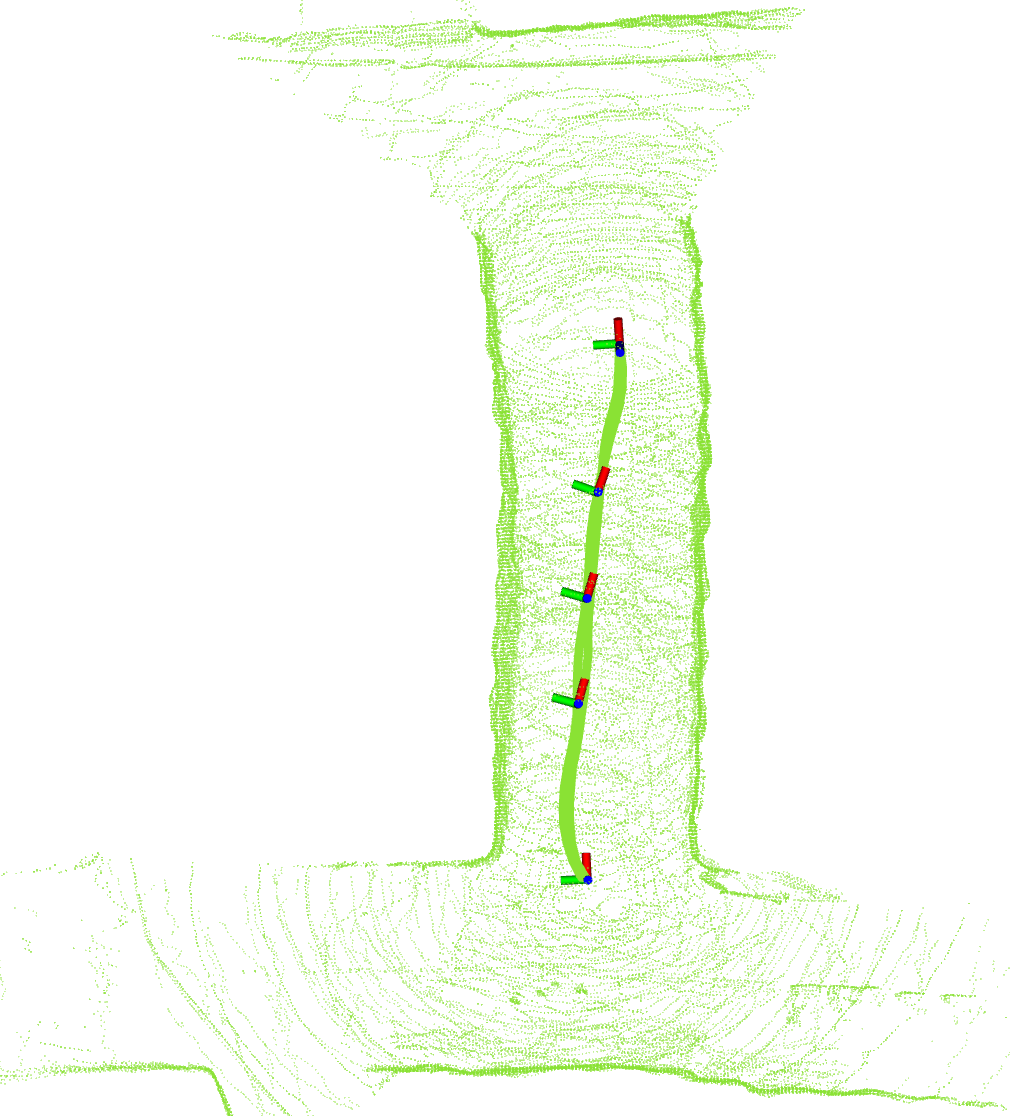} \caption{$\mathbf{M}_4$ and $\mathbf{P}_4$}
    \end{subfigure} \\
    \begin{subfigure}{0.32\textwidth}
    \includegraphics[width=\textwidth]{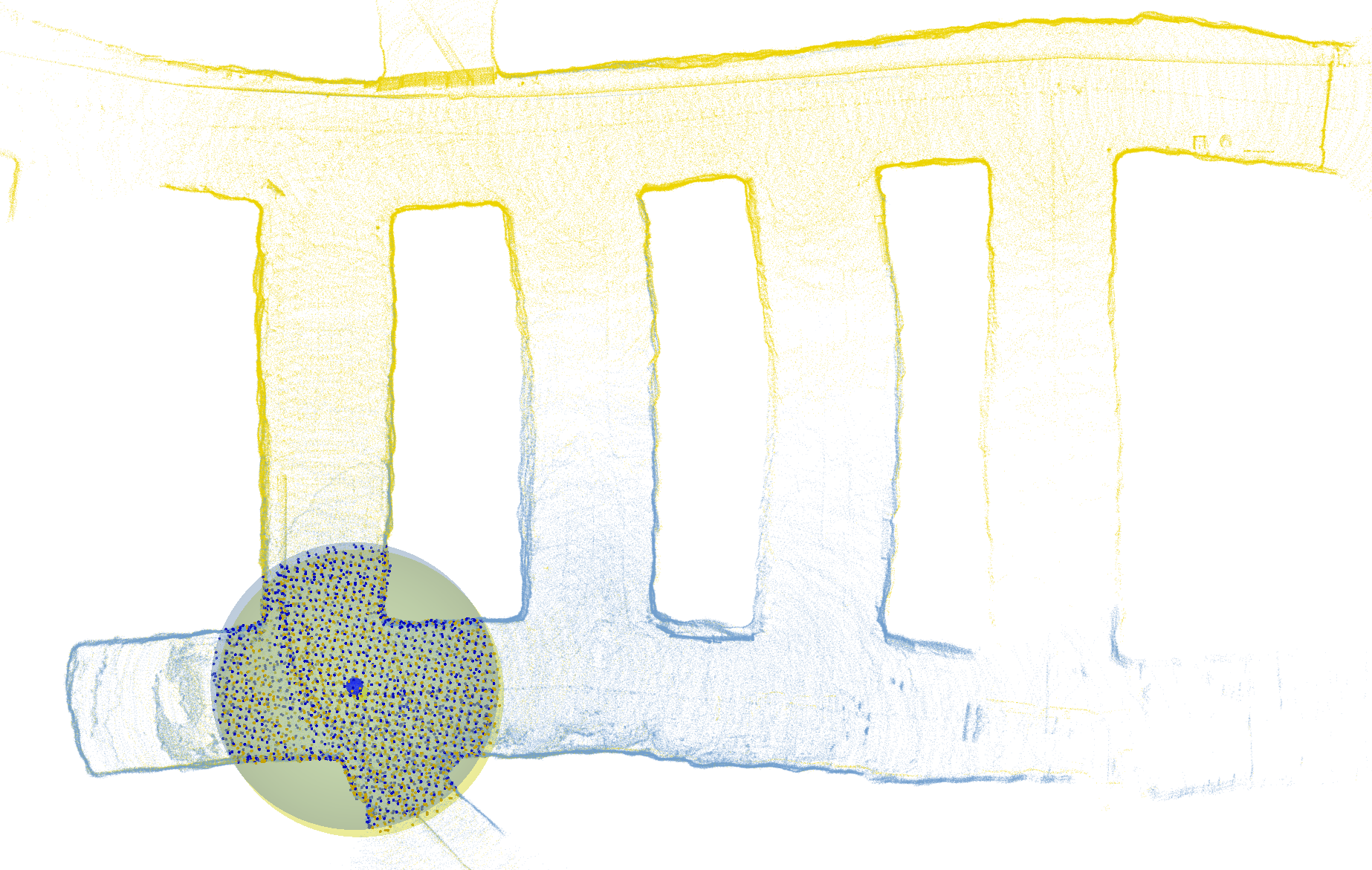} \caption{$\mathbf{S}_1$, $\mathbf{S}_2$ and $\mathbf{M_{12}}$}
    \end{subfigure} \hfill
    \begin{subfigure}{0.32\textwidth}
    \includegraphics[width=\textwidth]{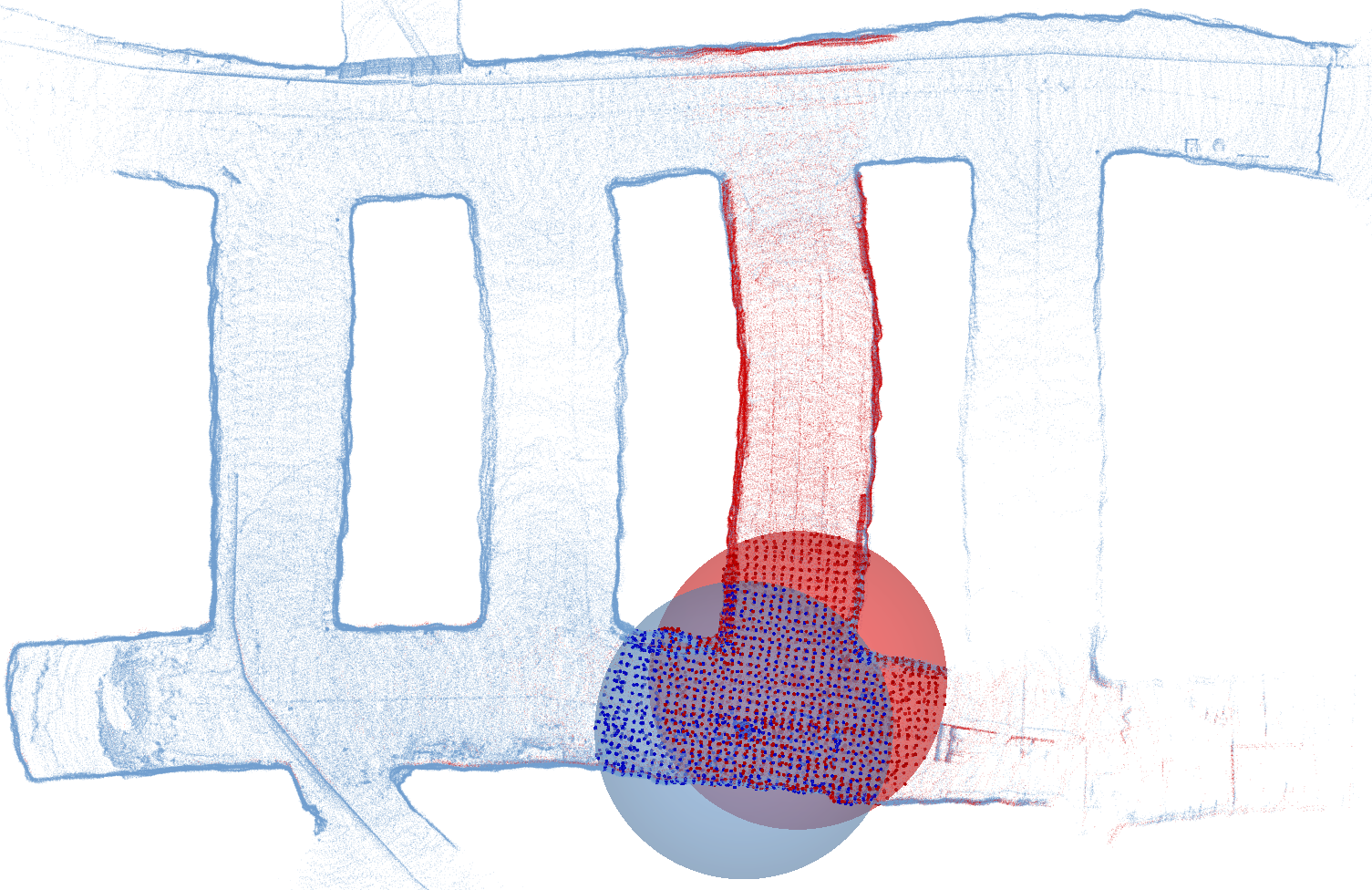} \caption{$\mathbf{S}_{12}$, $\mathbf{S}_3$ and $\mathbf{M}_{13}$}
    \end{subfigure} \hfill
    \begin{subfigure}{0.32\textwidth}
    \includegraphics[width=\textwidth]{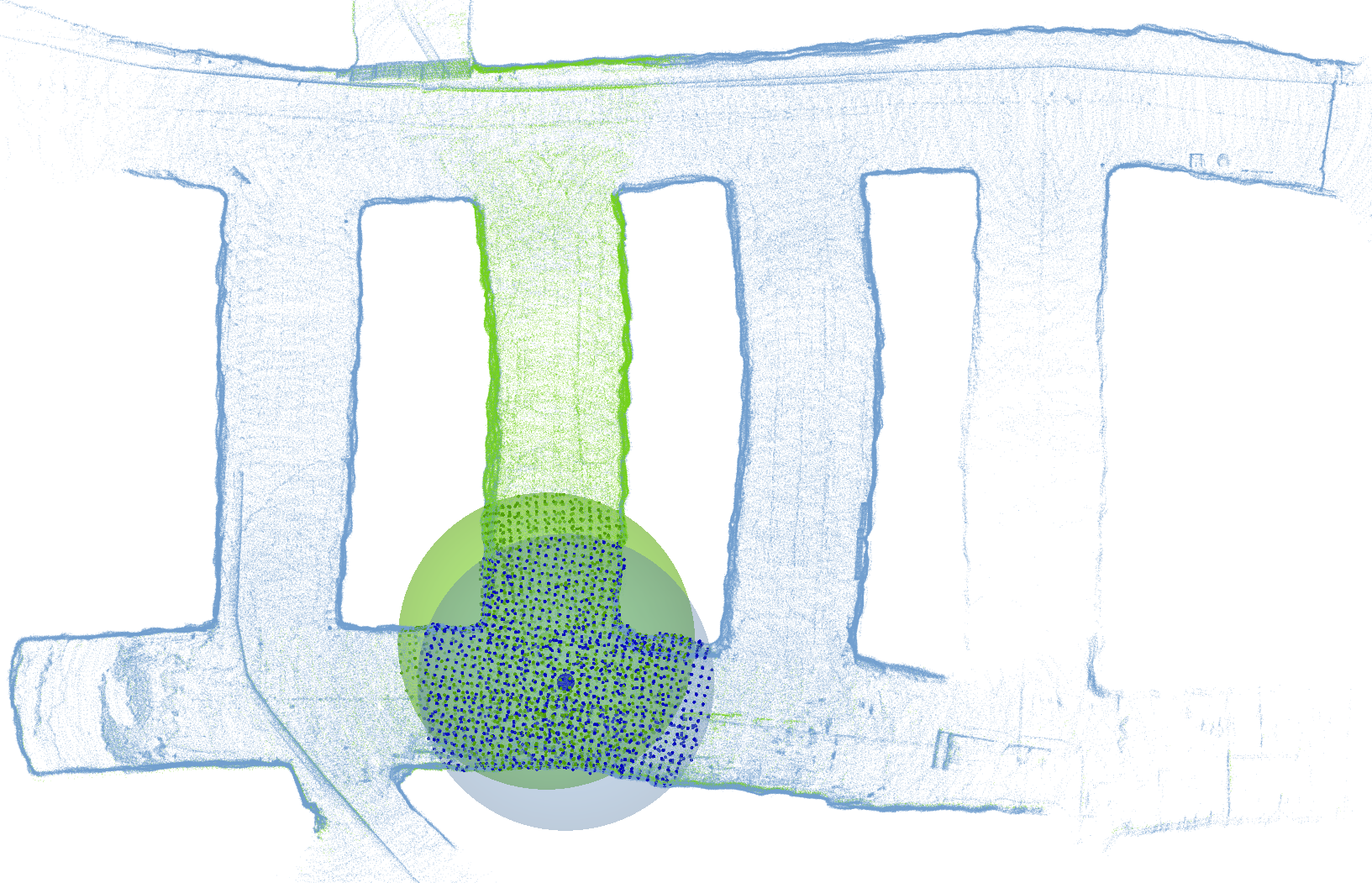} \caption{$\mathbf{S}_{13}$, $\mathbf{S}_4$ and $\mathbf{M}_{14}$}
    \end{subfigure}
    \caption{A set of four individual maps, namely $\mathbf{M}_1$, $\mathbf{M}_2$, $\mathbf{M}_3$, and $\mathbf{M}_4$, from a real-world test mine that we aim to merge sequentially. The merging process is based on identifying the overlapping regions and using them to obtain the merged map ${}^1\mathbf{M}_{12} = {}^1\mathbf{M}_1 \cup {}^1\mathbf{T}_{2} {}^2\mathbf{M}_2$. Next, we use the overlapping regions $\mathbf{S}_{12}$, $\mathbf{S}_3$ to merge $\mathbf{M}_3$ with ${}^1\mathbf{M}_{12}$ to obtain ${}^1\mathbf{M}_{13}$. Finally, we use the overlapping regions $\mathbf{S}_{13}$, $\mathbf{S}_4$ to merge $\mathbf{M}_4$ with ${}^1\mathbf{M}_{13}$ to obtain ${}^1\mathbf{M}_{14}$.}
    \label{fig:epiroc}
\end{figure*}

In this scenario, the COMPRA exploration mission was deployed in one long mission through the environment, and three short ones in order to complete the map. Here, we target the use-case of specifically sending the autonomous UAV to map areas that were yet not explored from the initial exploration run.
The subterranean scenario depicted in Fig.~\ref{fig:epiroc} consists of four instances, namely $\mathbf{M}_1$, $\mathbf{M}_2$, $\mathbf{M}_3$, and $\mathbf{M}_4$. Only the aerial platform was used to generate maps, with $\mathbf{M}_1$ and $\mathbf{M}_2$ consisting of $5-6 \cdot 10^5$ points and $\mathbf{M}_3$ and $\mathbf{M}_4$ consisting of $2.8 \cdot 10^5$ points. 
The merging process began with $\mathbf{M}_1$ and $\mathbf{M}_2$, resulting in a merged map ${}^1\mathbf{M}_{12} = {}^1\mathbf{M}_1 \cup {}^1\mathbf{T}_{2} {}^2\mathbf{M}_2$, where ${}^1\mathbf{T}_{2}$ is the transform obtained by merging $\mathbf{M}_1$ and $\mathbf{M}_2$. 
This process was repeated with ${}^1\mathbf{M}_{12}$ taking the place of $\mathbf{M}_1$ until all four maps were merged. 
The process of merging maps $\mathbf{M}_3$ and $\mathbf{M}_4$ introduce the challenge that the trajectories of the robot do not overlap, and therefore the initial guess $\mathbf{T}_0$ will contain a minimum error of multiple \unit{meters}.
Despite this fact, by estimating a good enough initial guess $\mathbf{T}_0$, automatically adjusting the parameters such as the sphere radius to $r=15$ \unit{meters}, and the correspondence threshold radius, FRAME was able to align the point clouds while keeping the translation and rotation error low, in an average computational time of $0.36$ seconds.
We will further discuss on this dataset as we utilize it to compare with other methods in subsection~\ref{subsec:comparisons}.

\begin{figure*}[!b] 
    \centering
    \includegraphics[width=\textwidth]{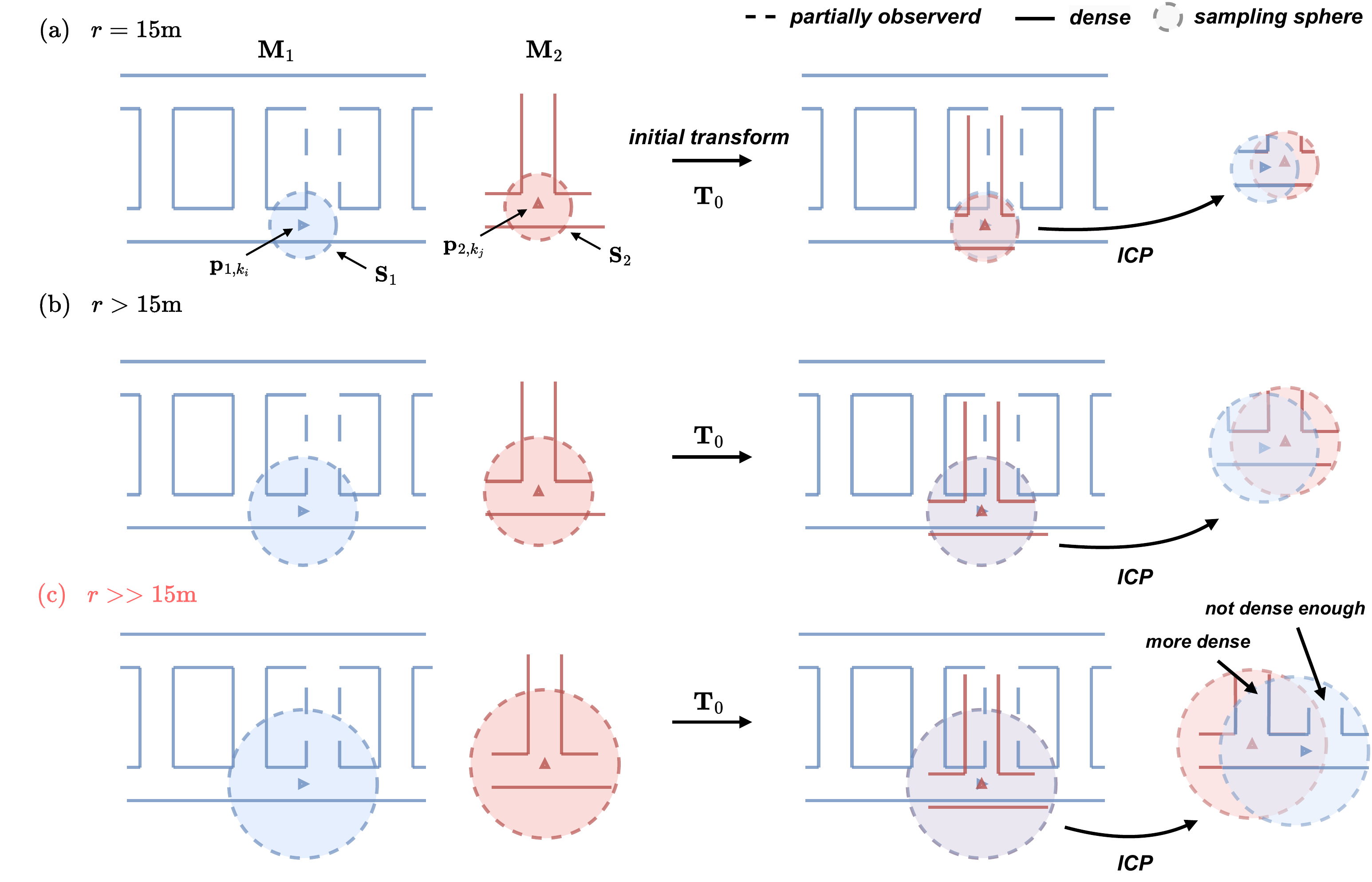}
    \caption{An illustrative example demonstrating the impact of increasing the radius of the sampling sphere on introducing false positive correspondences and the subsequent failure of the ICP algorithm. The solid lines depict the observed point cloud map, while the dotted lines represent areas that are partially observed, containing significantly fewer points.}
    \label{fig:sphere_example}
\end{figure*}

Before we move on to the comparisons with other frameworks, we would like to pose this discussion on the order of merging multiple maps and its effect on the result. This analysis relies on Eq.~(\ref{eq:f_m-general}) and our assumption of an egocentric approach. In the case of merging multiple maps sequentially, each map undergoes an independent transformation. Considering that maps are defined as sets of points and the global map represents the union of correctly transformed maps, the map merging process adheres to the commutative and associative properties of the union operation. Emphasizing the constraint introduced in the problem formulation, which requires an overlap between two maps, we delve into the specifics using this experiment as an example. For $N=4$, Eq.~(\ref{eq:f_m-general}) unfolds as follows:
\begin{equation}
\begin{split}
    {}^{1}\mathbf{M}_G &= f_m \big( {}^{1}\mathbf{M}_1, {}^{2}\mathbf{M}_2, 
    {}^{3}\mathbf{M}_3, {}^{4}\mathbf{M}_4 \big) \\
    &= {}^{1}\mathbf{M}_1 \cup {}^{1}\mathbf{M}_2 \cup {}^{1}\mathbf{M}_3 \cup {}^{1}\mathbf{M}_4 \\
    &= {}^{1}\mathbf{M}_1 \cup {}^{1}\mathbf{T}_{2} {}^{2}\mathbf{M}_2 \cup {}^{1}\mathbf{T}_{3} {}^{3}\mathbf{M}_3 \cup {}^{1}\mathbf{T}_{4} {}^{4}\mathbf{M}_4,
\end{split}
\end{equation}
subject to the constraints:
\begin{equation}
\begin{split}
    {}^{1}\mathbf{V}_1 \cap &{}^{1}\mathbf{V}_2 \neq \O, \:\: 
    ({}^{1}\mathbf{V}_1 \cup {}^{1}\mathbf{V}_2) \cap {}^{1}\mathbf{V}_3 \neq \O,\:\: \\
    & ({}^{1}\mathbf{V}_1 \cup {}^{1}\mathbf{V}_2 \cup {}^{1}\mathbf{V}_3) \cap {}^{1}\mathbf{V}_4 \neq \O.
\end{split}
\end{equation}
The commutative and associative properties of the union operation allow for a change in the order of merging, provided that these constraints are consistently met. Consequently, the conclusion drawn is that while the order of the map merging process can be altered, it is crucial to ensure that overlapping is maintained at each step. To clarify the constraints, it is helpful to express them in simpler terms. The fundamental requirement is that the map to be merged must overlap with at least one of the previous maps. This condition becomes more apparent when we unfold the second and third constraints, leveraging the distributed laws of the union and intersection operations defined as $A \cup (B \cap C) = (A \cup B) \cap (A \cup C)$ and $A \cap (B \cup C) = (A \cap B) \cup (A \cap C)$, we get the following:
\begin{equation}
\begin{split}
    ({}^{1}\mathbf{V}_1 \cup &{}^{1}\mathbf{V}_2) \cap {}^{1}\mathbf{V}_3 \neq \O \Rightarrow {}^{1}\mathbf{V}_3 \cap ({}^{1}\mathbf{V}_1 \cup {}^{1}\mathbf{V}_2) \neq \O \\
    &\Rightarrow ({}^{1}\mathbf{V}_3 \cap {}^{1}\mathbf{V}_1) \cup ({}^{1}\mathbf{V}_3 \cap {}^{1}\mathbf{V}_2) \neq \O.
\end{split}
\end{equation}
This derivation essentially mandates that ${}^{1}\mathbf{M}_3$ should overlap with at least one of the other maps, namely ${}^{1}\mathbf{M}_2$ and ${}^{1}\mathbf{M}_1$. A similar derivation can be applied to the third constraint and the four maps in general. This clarification reinforces the significance of maintaining overlapping relationships between maps during the merging process, even when the order of merging is altered.

Last but not least, we would like to discuss the automated sphere radius selection, as this experiment is a good example of rare instances where increasing the radius introduces points from the map that may confuse the registration algorithm, leading to incorrect correspondences, as briefly discussed previously in subsection~\ref{subsec:sphere_radius}. In the map merging process, the two spheres are sampled from the point cloud maps without using LiDAR scans. This approach captures points that would typically be out of the sensor's field of view due to obstructions, potentially causing the registration algorithm to fail. To illustrate this point more comprehensively, we present an example in Fig.~\ref{fig:sphere_example}, inspired by the aforementioned experiment and its mine structure. The example demonstrates the merging process between the larger map $\mathbf{M}_1$ and one of the missing tunnels $\mathbf{M}_3$. Starting with (a) and a radius $r=15\text{m}$ in Fig.~\ref{fig:sphere_example}, the sphere sampling includes sufficient information, such as the shared corner, facilitating correct alignment through the ICP. However, examples (b) and (c) show that increasing the radius $r$ may lead to failure. More specifically, in (c), a significantly increased radius causes the sphere to sample the adjacent tunnel, leading to the registration algorithm attempting to fit $\mathbf{M}_2$ to the previous tunnel due to the partially observed area (dotted lines) containing fewer points. To mitigate such rare cases, we introduce the automatic selection of the sampling sphere radius based on the spaciousness. This enhancement aims to eliminate instances where an overly large radius introduces points that could adversely affect the registration algorithm.

\begin{table*}[b!]
\centering \caption{The experimental metric results, where $\mathbf{M}_1$ and $\mathbf{M}_2$ represent the amount of points in each point cloud and $\mathbf{P}_1$ and $\mathbf{P}_2$ represent the trajectory travelled.} \label{table:results}
\resizebox{0.9\linewidth}{!}{%
\begin{tblr}{
  row{1} = {c},
  cell{2}{2} = {c},
  cell{2}{3} = {c},
  cell{2}{4} = {c},
  cell{2}{5} = {c},
  cell{2}{6} = {c},
  cell{2}{7} = {c},
  cell{2}{8} = {c},
  cell{2}{9} = {c},
  cell{3}{2} = {c},
  cell{3}{3} = {c},
  cell{3}{4} = {c},
  cell{3}{5} = {c},
  cell{3}{6} = {c},
  cell{3}{7} = {c},
  cell{3}{8} = {c},
  cell{3}{9} = {c},
  cell{4}{2} = {c},
  cell{4}{3} = {c},
  cell{4}{4} = {c},
  cell{4}{5} = {c},
  cell{4}{6} = {c},
  cell{4}{7} = {c},
  cell{4}{8} = {c},
  cell{4}{9} = {c},
  cell{5}{2} = {c},
  cell{5}{3} = {c},
  cell{5}{4} = {c},
  cell{5}{5} = {c},
  cell{5}{6} = {c},
  cell{5}{7} = {c},
  cell{5}{8} = {c},
  cell{5}{9} = {c},
  cell{6}{2} = {c},
  cell{6}{3} = {c},
  cell{6}{4} = {c},
  cell{6}{5} = {c},
  cell{6}{6} = {c},
  cell{6}{7} = {c},
  cell{6}{8} = {c},
  cell{6}{9} = {c},
  cell{7}{2} = {c},
  cell{7}{3} = {c},
  cell{7}{4} = {c},
  cell{7}{5} = {c},
  cell{7}{6} = {c},
  cell{7}{7} = {c},
  cell{7}{8} = {c},
  cell{7}{9} = {c},
  cell{8}{2} = {c},
  cell{8}{3} = {c},
  cell{8}{4} = {c},
  cell{8}{5} = {c},
  cell{8}{6} = {c},
  cell{8}{7} = {c},
  cell{8}{8} = {c},
  cell{8}{9} = {c},
  cell{9}{2} = {c},
  cell{9}{3} = {c},
  cell{9}{4} = {c},
  cell{9}{5} = {c},
  cell{9}{6} = {c},
  cell{9}{7} = {c},
  cell{9}{8} = {c},
  cell{9}{9} = {c},
  cell{10}{2} = {c},
  cell{10}{3} = {c},
  cell{10}{4} = {c},
  cell{10}{5} = {c},
  cell{10}{6} = {c},
  cell{10}{7} = {c},
  cell{10}{8} = {c},
  cell{10}{9} = {c},
  hline{1-2,11} = {-}{0.08em},
}
                      & $\mathbf{M}_1$             & $\mathbf{M}_2$             & $\mathbf{P}_1$ [m] & $\mathbf{P}_2$ [m] & Overlap [\%] & $T_e$ [m] & $R_e$ [deg] & Time [s] \\
Fig.~\ref{fig:mjolkberget} (a)-(b)         & $1.59 \cdot 10^5$ & $1.61\cdot 10^5$  & 136        & 257        & 33           & 0.092     & 3.430       & 0.229    \\
Fig.~\ref{fig:ncc} (a)-(b)         & $5.23\cdot 10^5$  & $1.06 \cdot 10^6$ & 60         & 70         & 37           & 0.220     & 1.492       & 0.108    \\
Fig.~\ref{fig:lkab_junction} (a)-(b)         & $2.85 \cdot 10^6$ & $2.80 \cdot 10^6$ & 145        & 147        & 31           & 0.082     & 0.551       & 0.224    \\
Fig.~\ref{fig:kps} (a)-(b)         & $2.08\cdot 10^6$  & $4.50\cdot 10^6$  & 115        & 157        & 22           & 0.098     & 0.914       & 0.352 \\
Fig.~\ref{fig:lkab_left_right} (a)-(b)         & $2.83 \cdot 10^6$ & $2.49 \cdot 10^6$ & 140        & 140        & 18           & 0.108     & 0.744       & 0.221    \\
Fig.~\ref{fig:epiroc_machine} (a)-(b)         & $1.44 \cdot 10^6$ & $6.17 \cdot 10^5$ & 106        & 61         & 33           & 0.120     & 2.988       & 0.109    \\
Fig.~\ref{fig:epiroc} (a)-(b)         & $5.95\cdot 10^5$  & $5.14\cdot 10^5$  & 139        & 113        & 35           & 0.161     & 1.031       & 0.325    \\
Fig.~\ref{fig:epiroc} (a)-(b)-(c)     & $1.11\cdot 10^6$  & $2.82\cdot 10^5$  & 252        & 86         & 82           & 0.119     & 0.286       & 0.360    \\
Fig.~\ref{fig:epiroc} (a)-(b)-(c)-(d) & $1.39\cdot 10^6$  & $2.73\cdot 10^5$  & 338        & 61         & 87           & 0.079     & 0.229       & 0.391   
\end{tblr}
} 
\end{table*}



\subsection{Comparison with Other Frameworks} \label{subsec:comparisons}

To compare the performance of the proposed framework, FRAME, with existing methods, two other publicly available algorithms were evaluated, namely map-merge-3D~\footnote{\url{https://github.com/hrnr/map-merge.git}}~\cite{map-merge-3d} and 3D map server~\footnote{\url{https://github.com/mdrwiega/3d\_map\_server.git}}~\cite{3d_map_server}. 
In the remainder of the article, these methods will be referred to as MM3D and 3DMS, respectively. 
The map-merge-3D algorithm first pre-processes the point cloud maps to remove any outliers, then extracts 3D features using SIFT points or Harris corners, and finally compares the features to find correspondences and align the two point clouds. 
On the other hand, the 3D map server method, which was previously discussed in Section~\ref{sec:related_work}, relies on the SHOT descriptors for overlap estimation and uses the SAC-IA for alignment. 

\begin{figure*}[!t]
    \begin{subfigure}{0.25\textwidth}
    \includegraphics[width=\textwidth]{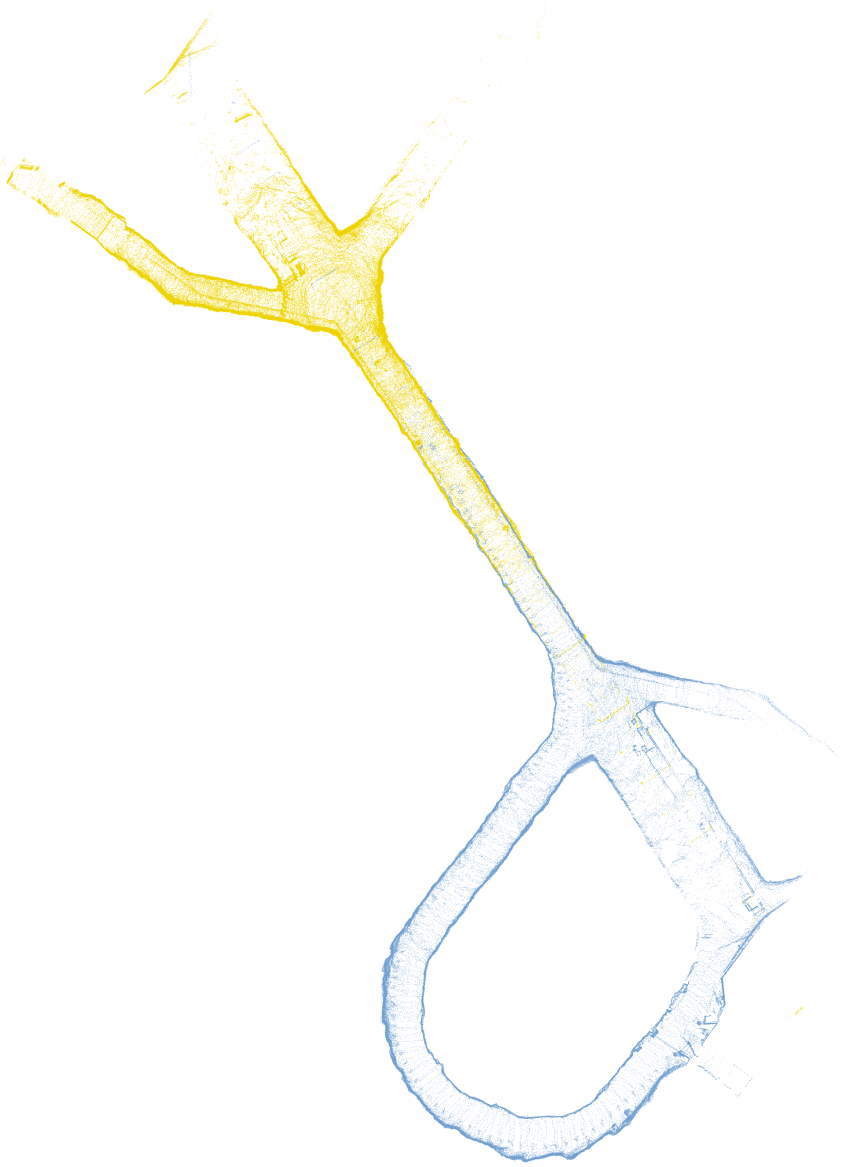}
    \caption{Original $|$ MM3D}
    \end{subfigure}
    \begin{subfigure}{0.27\textwidth}
    \includegraphics[width=\textwidth]{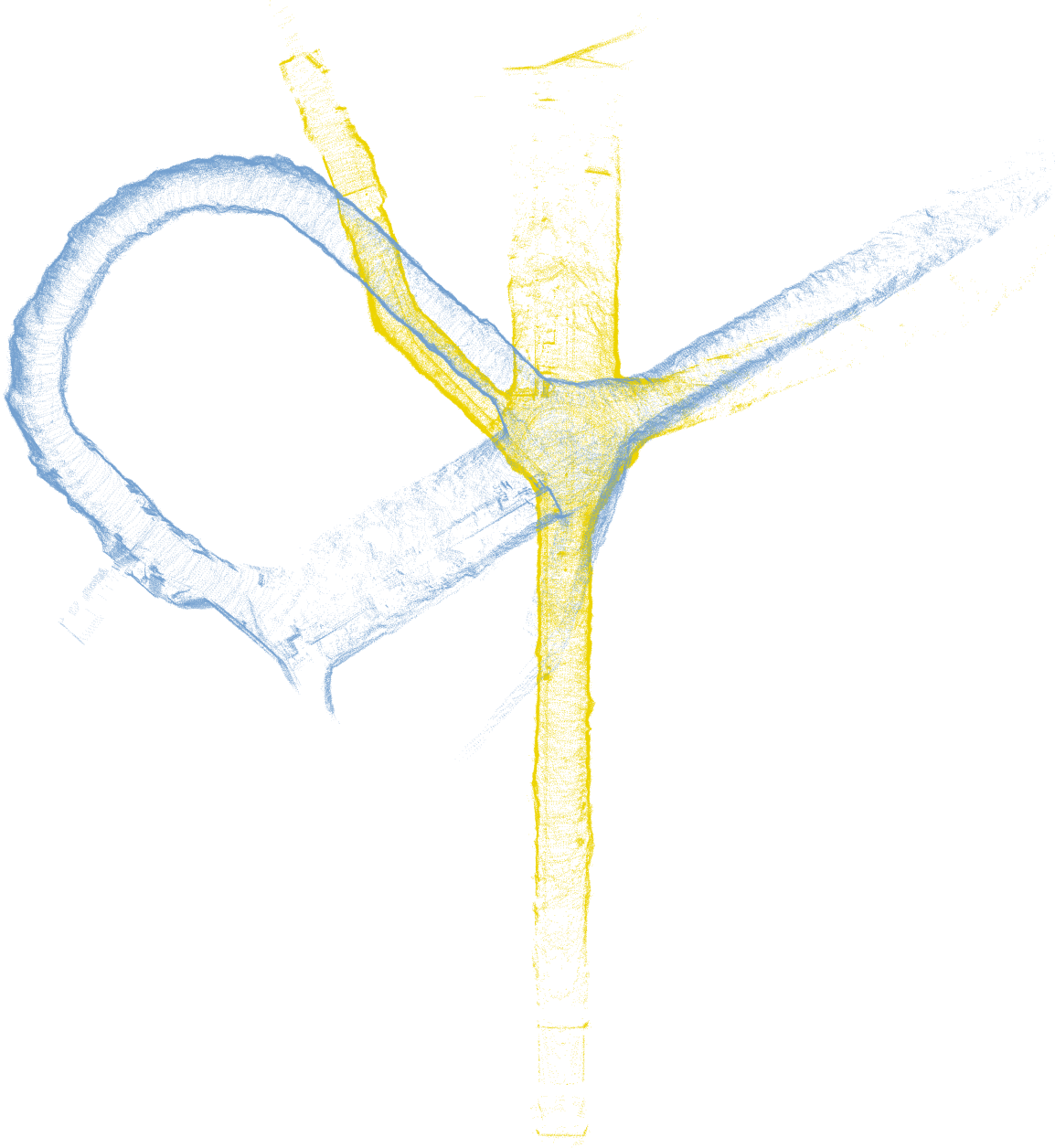}
    \caption{Original $|$ 3DMS}
    \end{subfigure}
    \begin{subfigure}{0.22\textwidth}
    \includegraphics[width=\textwidth]{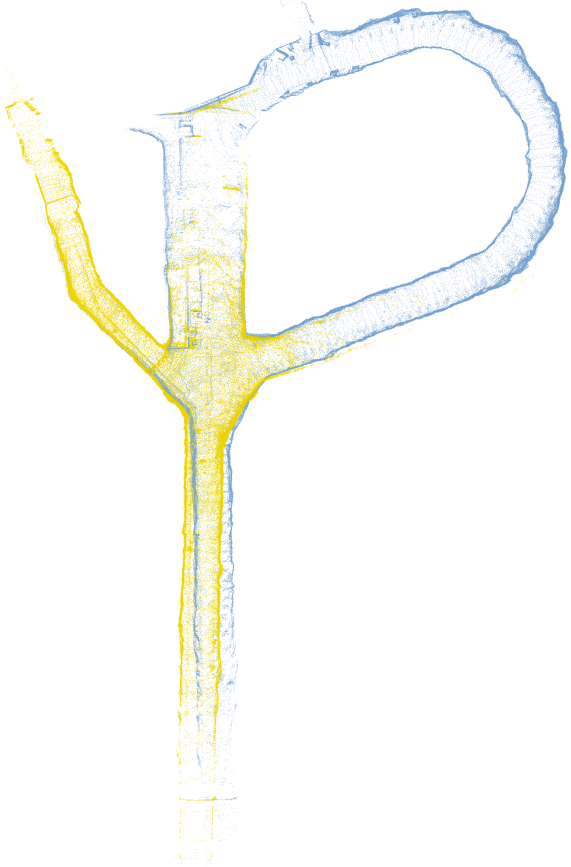} 
    \caption{Simplified $|$ MM3D}
    \end{subfigure}
    \begin{subfigure}{0.23\textwidth}
    \includegraphics[width=\textwidth]{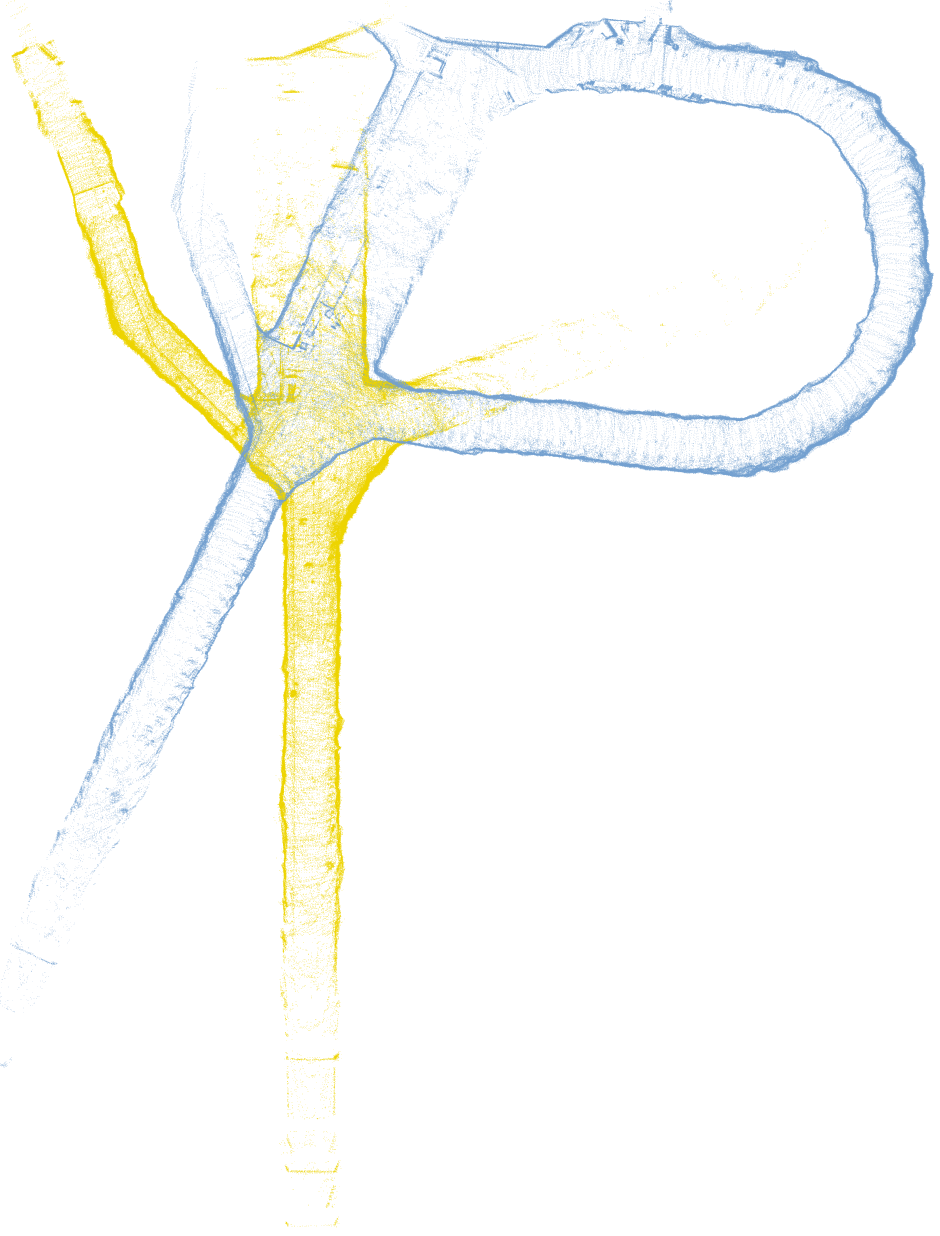}
    \caption{Simplified $|$ 3DMS}
    \end{subfigure}
    \caption{The resulted maps derived from the compared methods. Subfigure (a) and (b) depict the results for the initial experiment with the high yaw difference between map $\mathbf{M}_1$ and $\mathbf{M}_2$, while subfigure (c) and (d) depict the results from the simplified experiment where the initial yaw difference is significantly reduced.}
    \label{fig:mjolkberget_comparison}
\end{figure*}

For evaluating FRAME against the two state-of-the-art algorithms, we start by using the two maps previously discussed in Figure \ref{fig:mjolkberget}.
Despite extensive tuning of the parameters and descriptors offered by the aforementioned comparison map merging packages, the authors were unable to achieve satisfactory results in any of the resulting maps presented in Figure \ref{fig:mjolkberget_comparison}. 
The first package used, MM3D, produced a high rotational error $R_e=159.52^o$ due to its inability to regress the 180$^o$ yaw difference between the two maps, as shown in Table~\ref{table:simplified}.
Although the second package, 3DMS, was able to estimate the yaw difference $\delta \psi$ up to some degree, it failed to correctly match the two side corridors, leading to a high roll angle error. 
In order to further evaluate these packages, the first pair of maps was simplified. 
For that, the yaw discrepancy was intentionally reduced to approximately to 15$^o$, giving an advantage to and assisting the comparison methods.
As indicated in Table~\ref{table:simplified}, after the simplifications, MM3D produced a sufficient result this time, but with a longer computational time of around 9 seconds, compared to FRAME, which achieved higher translation accuracy in just under 0.3 seconds. 
The 3DMS package still could not produce a satisfactory result after 260 seconds of computational time.
\begin{figure*}[b!]
    \begin{subfigure}{0.3\textwidth}
    \includegraphics[width=\textwidth]{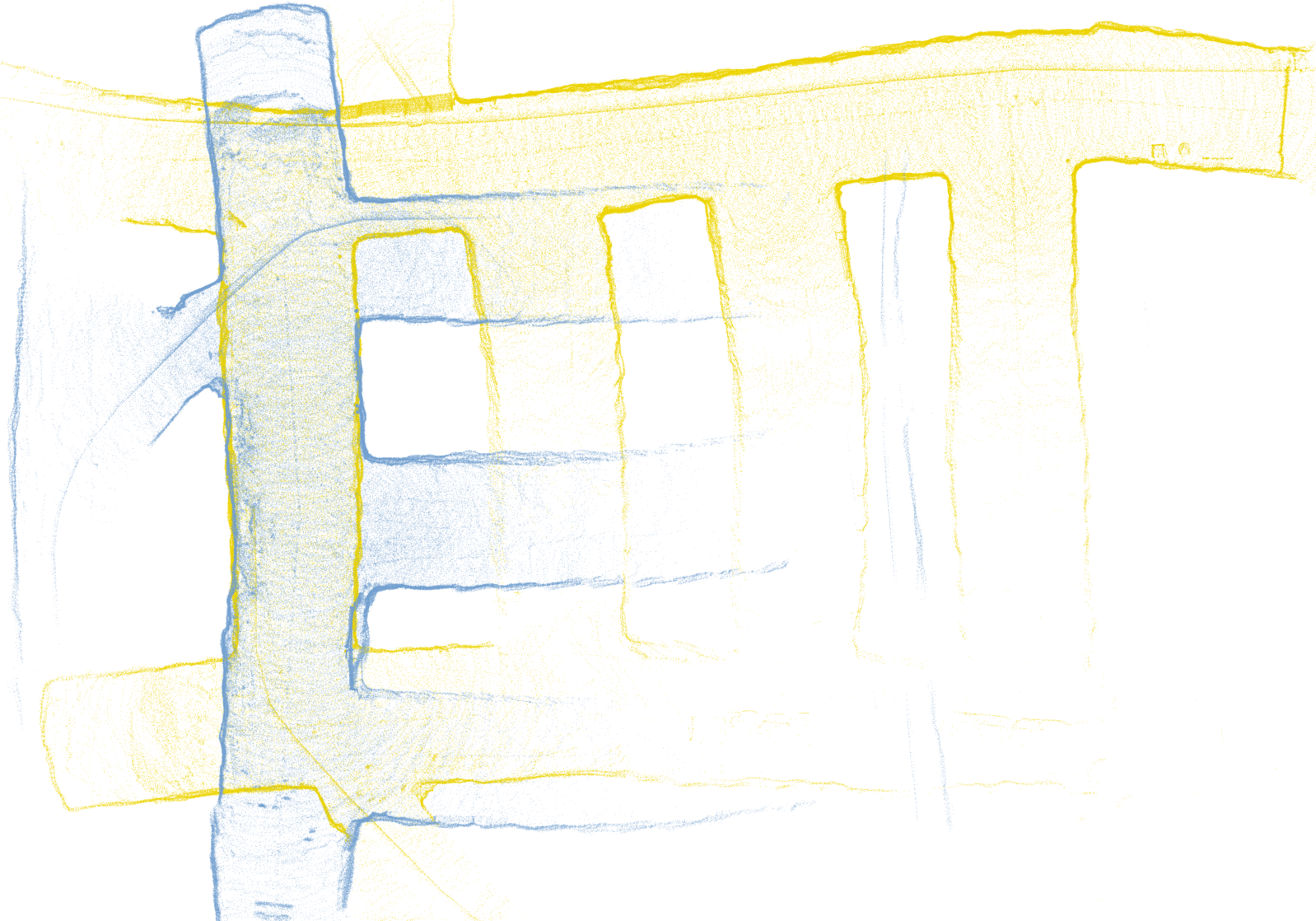}
    \caption{MM3D}
    \end{subfigure}
    \begin{subfigure}{0.33\textwidth}
    \includegraphics[width=\textwidth]{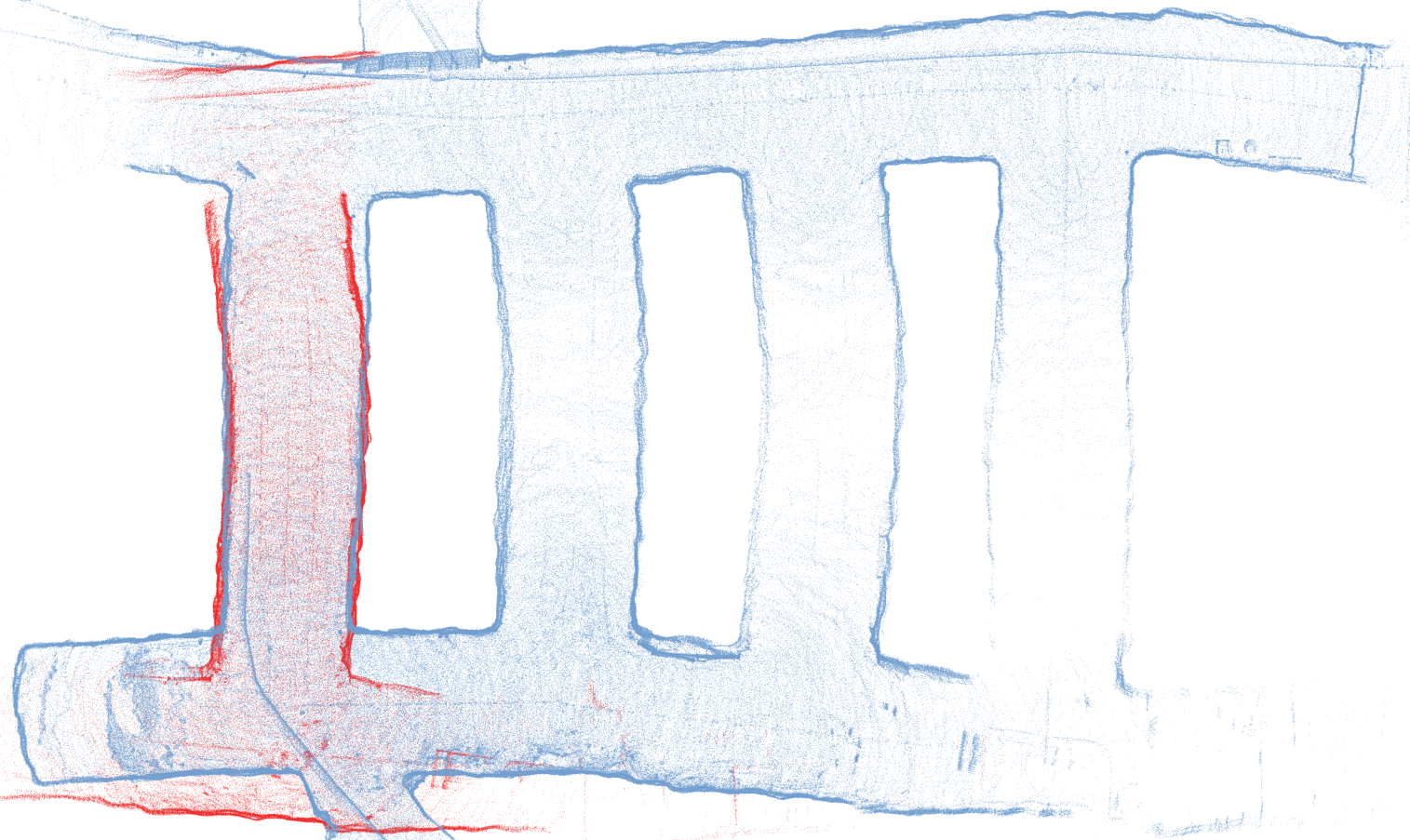}
    \caption{3DMS}
    \end{subfigure}
    \begin{subfigure}{0.34\textwidth}
    \includegraphics[width=\textwidth]{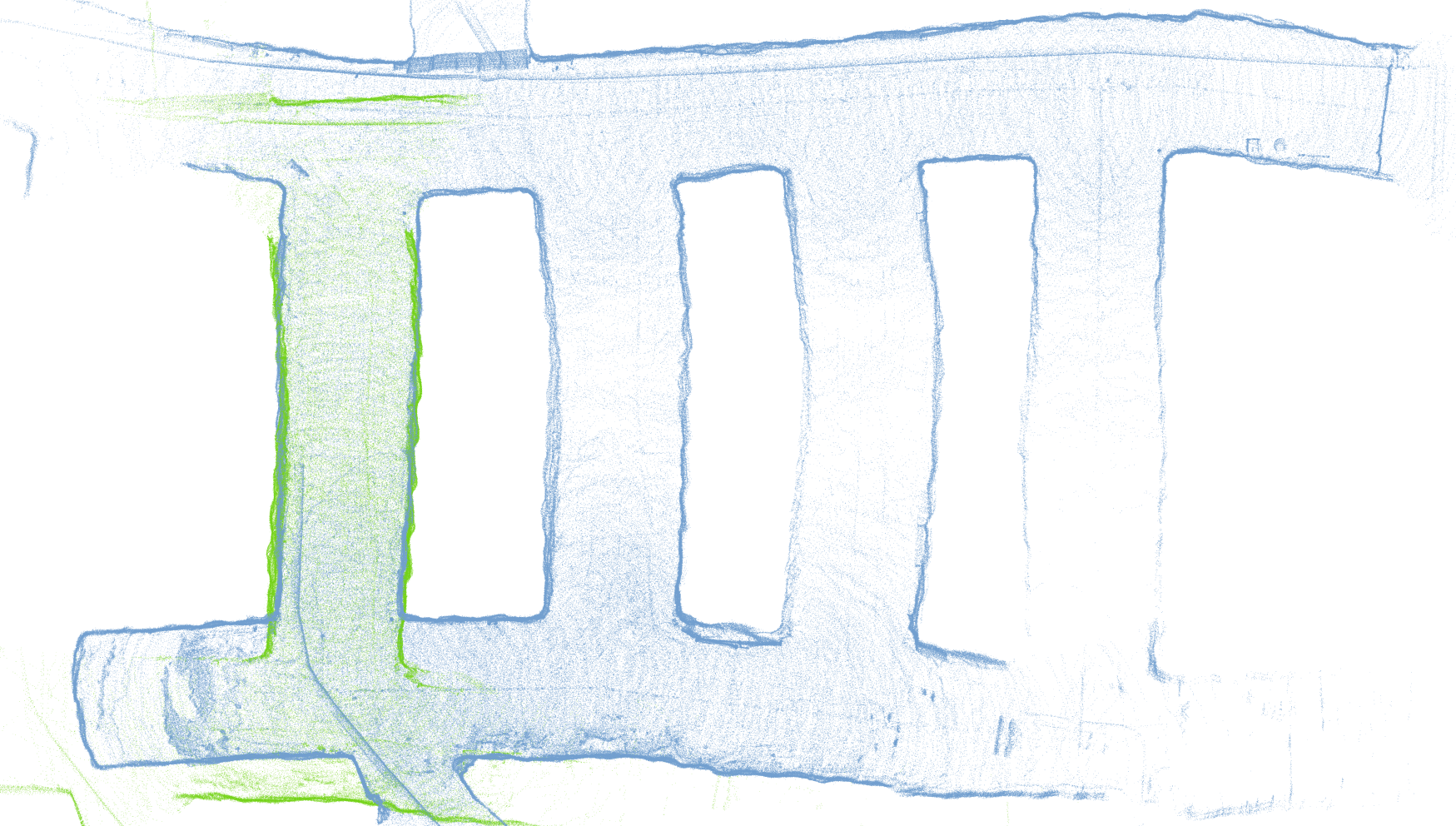} 
    \caption{MM3D}
    \end{subfigure}
    \caption{The resulted maps derived from the compared methods from the experiments presented in subsection~\ref{subsubsec:epiroc_multi}. Subfigure (a) depicts the result of MM3D for maps $\mathbf{M}_1$ and $\mathbf{M}_2$, subfigure (b) depicts the result of 3DMS for maps $\mathbf{M_{12}}$ and $\mathbf{M}_3$ and subfigure (c) depicts the result of MM3D for maps $\mathbf{M_{12}}$ and $\mathbf{M}_4$.}
    \label{fig:epiroc_comparison}
\end{figure*}

Another comparison test was performed in the dataset presented in~\ref{subsubsec:epiroc_multi} on the same set of maps as in Figure \ref{fig:epiroc}.
As mentioned before, this environment is challenging as it contains highly self-similar corridors. 
However, unlike FRAME, both packages used for comparison failed to compute a correct transform for the initial maps $\mathbf{M}_1$ and $\mathbf{M}_2$, due to the initial yaw difference of approximately 90$^o$.
Moving on, to maps $\mathbf{M}_1$ and $\mathbf{M}_4$, they presented particular challenges as they were not only similar to each other but also to the first corridor connecting $\mathbf{M}_1$ and $\mathbf{M}_2$. 
Due to the lack of distinct features, the algorithms used for comparison, matched both $\mathbf{M}_1$ and $\mathbf{M}_4$ to the first corridor as depicted in Fig.~\ref{fig:epiroc_comparison}. 
Nevertheless, FRAME was able to detect overlapping regions, namely $\mathbf{S}_{12}$, $\mathbf{S}_3$, and $\mathbf{S}_4$, even though the trajectories from each robot did not overlap (as previously demonstrated in Figure \ref{fig:epiroc}). 
By providing a good enough initial guess $\mathbf{T}_0$, automatically adjusting the sphere radius to $r=15$ \unit{meters} due to the tunnel spaciousness, and keeping the correspondence threshold radius low, FRAME was able to align the point clouds and at the same time maintaining the translation and rotation error low.
\begin{table*}[!t]
\centering
\caption{The experimental metric results for the original set of maps from Fig.~\ref{fig:mjolkberget} and for the simplified version.}
\label{table:simplified}
\begin{tblr}{
  row{1} = {c},
  row{2} = {c},
  cell{1}{2} = {c=3}{},
  cell{1}{5} = {c=3}{},
  cell{3}{2} = {c},
  cell{3}{3} = {c},
  cell{3}{4} = {c},
  cell{3}{5} = {c},
  cell{3}{6} = {c},
  cell{3}{7} = {c},
  cell{4}{2} = {c},
  cell{4}{3} = {c},
  cell{4}{4} = {c},
  cell{4}{5} = {c},
  cell{4}{6} = {c},
  cell{4}{7} = {c},
  cell{5}{2} = {c},
  cell{5}{3} = {c},
  cell{5}{4} = {c},
  cell{5}{5} = {c},
  cell{5}{6} = {c},
  cell{5}{7} = {c},
  hline{1-3,6} = {-}{},
}
               & Original       &               &               & Simplified    &               &               \\
               & $T_e$ [m]      & $R_e$ [deg]   & Time [s]      & $T_e$ [m]     & $R_e$ [deg]   & Time [s]      \\
Map-merge 3D~\cite{map-merge-3d}  & 70.78          & 159.52        & 30.78         & 1.13          & 3.56          & 9.21          \\
3D Map Server~\cite{3d_map_server} & 5.43           & 110.42        & 191.39        & 9.70          & 16.52         & 260.47        \\
FRAME~\cite{stathoulopoulos2023frame}          & \textbf{ 0.09} & \textbf{3.43} & \textbf{0.25} & \textbf{0.09} & \textbf{3.43} & \textbf{0.27} 
\end{tblr}
\end{table*}


\section{Limitations and Future Work} \label{sec:limitations}

The proposed approach, while promising, it is not without its limitations, each of which provides valuable insights for future research directions. A significant limitation stems from our reliance on learned descriptors, which can be vulnerable to sensor noise, dust, and other obstructions affecting range images. While we recognize the challenges posed by these factors, it's important to highlight that the generalizability of descriptors remains an active research area outside the scope of this paper. We acknowledge the potential alternative of feature extraction from direct LiDAR scans; however, this introduces high computational loads, particularly challenging for small mobile robots like UAVs. FRAME, designed to be adaptable to various descriptors based on deployment environments, is actively exploring solutions to reinforce robustness in the face of sensor-related challenges. Another limitation, closely tied to descriptors, is the assumption of yaw-invariance with minimal actuation along the other two axes. Although this assumption aligns with common practices in existing literature, we acknowledge its current limitation and anticipate further exploration of descriptor robustness across all axes. Drift in localization poses a significant challenge, particularly in large-scale scenarios. The current solution assumes a single overlap point and may encounter difficulties with distant branches affected by drift that will not be adjusted according to their accumulated error.

Future iterations of FRAME aim to extend its applicability to scenarios featuring multiple overlapping regions, addressing potential drift over time. Comparative assessments with established solutions like pose graph optimization or bundle adjustment will provide valuable insights. Then, an often-overlooked challenge in map merging solutions pertains to communication aspects. Our exploration of both centralized and decentralized communication scenarios has led to the proposal of a communication-aware control function in recent work~\cite{damigos2023com_aware}, regulating bandwidth and accommodating temporal loss for centralized communication. For decentralized approaches, where agents can only exchange information during rendezvous instances with each other or through a communication link, ensuring fast data exchange is crucial. The time available for these exchanges is limited, and it's essential to minimize their impact on exploration times and the overall mission efficiency. This becomes particularly challenging considering the substantial size of point cloud maps, often in the hundreds of megabytes or more, that need to be transmitted. To address this, we are currently investigating descriptors that are capable of efficiently compressing and decoding information~\cite{dube2020segmap, stathoulopoulos2024recnet}, facilitating faster and more streamlined data transfer during rendezvous instances. The identified limitations and proposed future directions collectively contribute to the ongoing evolution of FRAME, refining its capabilities and extending its applicability to more complex real-world scenarios.


\section{Conclusions} \label{sec:conclusions}

In conclusion, the proposed framework offers a significant improvement over traditional map merging approaches for 3D point cloud map merging in complex subterranean environments. The event-triggered, egocentric approach introduced by FRAME, which relies on learned descriptors, allows for fast and efficient querying of place recognition and yaw discrepancy regression descriptors, resulting in reduced computational time. The comprehensive evaluation presented, shows that FRAME is robust and adaptable to varying conditions in different subterranean environments, and offers faster convergence and lower translational and rotational errors. The comparison with state-of-the-art methods demonstrates that FRAME outperforms existing approaches in terms of computational time and accuracy, making it an ideal solution for multi-robot and multi-session exploration scenarios. These results highlight the potential of FRAME for real-life applications, such as the use of map merging in the mining industry that is crucial for better monitoring, maintenance, and decision-making, leading to improved safety and operational efficiency. The proposed framework offers a reliable and efficient solution for map merging in subterranean environments, which can help in reducing costs and increasing efficiency in mapping and inspection missions. As future work, its real-time integration into multi-agent exploration missions can further evaluate and quantify the benefits of a shared global frame between agents and enhance its robustness, enabling more efficient and effective exploration of complex environments.

\section*{Acknowledgments}
The authors would like to acknowledge and thank our collaborators in the mining and construction industry for enabling the field evaluation results: Epiroc Rock Drills AB, NCC AB, LKAB, and K+S Group. 

This work has been partially funded by the European Unions Horizon 2020 Research and Innovation Programme under the Grant Agreement No. 869379 illuMINEation and No.101003591 NEXGEN-SIMS, and partially funded by the Swedish Energy Agency through the Sustainable Underground Mining Academy Programme SP14.

\bibliographystyle{IEEEtranBST/IEEEtran}
\bibliography{IEEEtranBST/IEEEabrv,root}

\end{document}